\documentclass{article}
\usepackage{amsmath,epsfig,fancyhdr,amssymb,setspace,fullpage, amsthm}
\usepackage{tcolorbox}
\usepackage{comment}
\usepackage{algorithm}
\usepackage{algpseudocode}
\usepackage{graphicx}
\usepackage{subcaption}
\usepackage{hyperref}
\usepackage{aliascnt}
\usepackage{cleveref}
\usepackage{authblk}

\def\L{{\cal L}}

\newcommand{\R}{\mathbb{R}}
\def\S{\mathbb{S}}

\newcommand{\mL}{\mathcal{L}_{\drift,\Diff}}
\newcommand{\mLhat}{\mathcal{L}_{\hat{\drift},\hat{\Diff}}}

\newcommand{\A}{\mathcal{A}}

\newtheorem{theorem}{Theorem}[section]

\newaliascnt{lemma}{theorem}
\newtheorem{lemma}[lemma]{Lemma}
\aliascntresetthe{lemma}

\newaliascnt{proposition}{theorem}
\newtheorem{proposition}[proposition]{Proposition}
\aliascntresetthe{proposition}

\newaliascnt{definition}{theorem}
\newtheorem{definition}[definition]{Definition}
\aliascntresetthe{definition}

\newaliascnt{remark}{theorem}
\newtheorem{remark}[remark]{Remark}
\aliascntresetthe{remark}

\newaliascnt{example}{theorem}

\aliascntresetthe{example}

\newaliascnt{examples}{theorem}

\aliascntresetthe{examples}

\newaliascnt{problem}{theorem}

\aliascntresetthe{problem}

\newaliascnt{corollary}{theorem}

\aliascntresetthe{corollary}

\newaliascnt{conjecture}{theorem}

\aliascntresetthe{conjecture}

\newaliascnt{question}{theorem}

\aliascntresetthe{question}

\newtheorem{assumption}{Assumption}
\crefname{assumption}{Assumption}{Assumptions}
\Crefname{assumption}{Assumption}{Assumptions}

\newcommand{\beq}{\begin{equation}}
\newcommand{\eeq}{\end{equation}}

\newcommand{\eps}{\varepsilon}

\def\l{\left}
\def\r{\right}

\def\nb{\nabla}

\def\drift{b}

\def\Diff{\Sigma}

\def\E{\mathbb{E}}

\def\mP{\mathcal{P}}

\DeclareMathOperator*{\argmin}{arg\,min}

\def\dt{{\Delta t}}
\def\dta{{\Delta \tau}}
\def\th{\theta}
\def\lam{{\lambda}}

\def\nb{\nabla}
\def\o{\omega}

\newcommand{\rd}{\mathrm{d}}

\newcommand{\ns}{\nabla_s}
\newcommand{\nt}{\nabla_\theta}
\newcommand{\ve}{\varepsilon}

\newcommand{\veb}{\varepsilon_{\mathrm{bias}}}

\newcommand{\pta}{\partial_\tau}

\newcommand{\parentheses}[1]{\left(#1\right)}

\newcommand{\inner}[2]{\left\langle#1,\,#2\right\rangle}

\newcommand{\KL}[2]{\textrm{KL}\left(#1 || #2\right)}

\title{Continuous-Time Policy Gradient from Discrete Data: Convergence Analysis and Model-Free Algorithms}

\author[1]{Martin Hernandez}
\author[2]{Xinyan Qing}
\author[3]{Mo Zhou}
\author[1]{Yuhua Zhu}

\affil[1]{Department of Statistics and Data Science,
University of California, Los Angeles}
\affil[2]{Department of Statistics, University of Michigan}
\affil[3]{School of Mathematical Sciences, Peking University}

\date{}

\begin{document}

\maketitle
\begin{abstract}
Continuous-time reinforcement learning seeks an optimal policy under unknown dynamics driven by stochastic differential equations using observations collected at discrete times. We develop a policy gradient framework that combines rigorous convergence analysis with learning from discrete data. We first establish a performance difference lemma and derive policy gradient formulas for randomized and parametrized policies. We then prove exponential convergence of an entropy-regularized policy flow to the regularized optimum. For discrete parametrized updates, we derive a convergence bound that explicitly separates the contributions of critic error, policy-regression error, and the learning step size. To enable learning under unknown dynamics, we integrate this framework with PhiBE, which constructs a continuous-time surrogate generator from discrete transition data. The resulting model-free actor--critic algorithms, including an online algorithm, preserve the continuous-time structure of the control problem. Numerical experiments on an entropy-regularized linear-quadratic regulator and a nonlinear wheeled robot demonstrate the performance and stability of the proposed methods with both known and data-driven dynamics.
\end{abstract}

\section{Introduction}\label{sec:introduction}


Reinforcement learning (RL) in continuous time provides a natural framework for sequential decision making in systems governed by differential equations, with applications in robotics \cite{lin1998optimal}, autonomous systems \cite{song2023reaching}, manufacturing \cite{sethi2002optimal}, and finance \cite{pham2009continuous}. In many of these settings, the system evolves continuously while data are recorded only at discrete times through sensors or simulations. We consider controlled stochastic systems modeled by stochastic differential equations (SDEs), with decisions represented by randomized feedback policies. Thus, the dynamics and objective remain continuous in time, while learning relies on discrete observations.

Value-based methods face several difficulties in continuous-time RL. Under a fixed policy, the value function is characterized by a PDE involving the infinitesimal generator of the controlled diffusion rather than by a Bellman recursion \cite{wang2020reinforcement,flemingSoner2006}. Moreover, the usual discrete-time state-action value function loses its action-dependent information in the continuous-time limit, as the $Q$-function converges to the state value function when the action duration tends to zero \cite{jia2023q}. Even after policy evaluation, extracting a policy requires additional computation: a greedy construction requires optimization over the action space at every state \cite{lillicrap2016continuous}, while a softmax construction requires approximating its normalization constant \cite{vieillard2022implicit}. Thus, continuous-time value-based methods require suitable formulations of policy evaluation and advantage estimation, followed by a separate policy-extraction step.

Policy gradient methods avoid this separate policy-extraction step by optimizing a parametrized policy directly \cite{sutton1999policygradient}. Direct policy parametrization also makes it possible to control the size of each update: trust-region and proximal methods restrict the change between consecutive policies through a KL constraint or a clipped surrogate objective \cite{schulman2015trpo,schulman2017ppo}. Policy gradient methods still require policy evaluation, since the gradient depends on an advantage function, but the resulting estimate is used to update the policy directly rather than to recover it through an additional maximization problem.

For continuous-time stochastic control with known dynamics, policy gradient formulas and convergence results have been established under suitable assumptions \cite{zhou2025policy,giegrich2024convergence,KERIMKULOV2025104765,sethi2025entropy}. In this setting, the advantage can be constructed from the drift, the diffusion, and the policy value function \cite{jia2023q}. These results provide a rigorous foundation for continuous-time policy optimization, but their convergence analyses generally assume access to the system coefficients or to exact policy-evaluation quantities.

When the dynamics are unknown, the value and advantage must instead be learned from observed transitions. Existing approaches use actor--critic or adaptive dynamic programming methods \cite{yang2014reinforcement,jiang2012computational,zhao2022data}, while other work focuses specifically on continuous-time policy evaluation \cite{jia2022policy}. Another approach is to formulate a discrete-time MDP on the observation grid, but this discards the continuous-time structure and introduces a discretization error even when the discrete-time transition kernel is known exactly \cite{zhu2025optimal}. A remaining challenge is to develop data-driven continuous-time policy gradient methods with convergence guarantees that explicitly account for errors in both policy evaluation and policy parametrization.

\subsection*{Our contribution}

We establish rigorous convergence guarantees for policy gradient algorithms and develop model-free actor--critic methods based on this analysis. Our contributions are threefold.

First, we establish a policy gradient theorem for randomized and parametrized policies from a performance difference lemma, expressed in terms of the continuous-time advantage and discounted occupancy measure.

Second, we establish convergence guarantees for continuous-time policy optimization and finite-step parametrized updates. The entropy regularization yields exponential convergence of the policy gradient flow, while the finite-step analysis explicitly separates critic error, policy-regression error, and the learning step size.

Third, we develop model-free actor--critic algorithms, including an online algorithm, based on the PhiBE framework directly from discrete transition data, without identifying the dynamics or reducing the problem to a discrete-time MDP. The methods are demonstrated on an entropy-regularized LQ problem and a nonlinear differential wheeled robot.

\subsection*{Related work}

\paragraph{Exploratory continuous-time RL.} The entropy-regularized exploratory formulation was introduced in \cite{wang2020reinforcement}, connecting randomized policies with relaxed stochastic control. Subsequent work studied the exploratory HJB equation and the vanishing-regularization limit \cite{tang2022exploratory,reisinger2021regularity}. We adopt this formulation but focus on policy gradient optimization and learning from discrete data.

\paragraph{Policy gradient and actor--critic in continuous time.} Advantage updating was introduced in \cite{Baird1994ReinforcementLI} to learn state values and action-dependent advantage rates separately. The collapse of the conventional discrete-time $Q$-function towards the value function in the continuous-time limit was subsequently formalized in \cite{tallec2019}. For controlled diffusions, \cite{jia2022policyeval} characterized policy evaluation through martingale identities and developed methods based on martingale losses or orthogonality conditions. This approach was extended to parametrized policy gradients and actor--critic algorithms in \cite{jia2022policy}, and to the little-$q$ function in \cite{jia2023q}, which represents an instantaneous action-dependent return rate and is jointly characterized with the value function through  martingale conditions. Closest to our theoretical analysis, \cite{zhao2023policy} uses discounted occupation measures to derive performance-difference and local-approximation formulas, together with a parametric policy-gradient representation. We first characterize the directional derivative of the performance functional along general admissible policy perturbations, controlling the perturbed occupation measure through SDE stability estimates, and then establish the corresponding parametric score formula. At the algorithmic level, we develop model-free actor--critic methods based on PhiBE \cite{zhuphibe_evaluation,zhu2025optimal}. In contrast to the neural actor--critic solver for elliptic HJB equations in \cite{zhou2021actor}, our methods require neither known drift nor known diffusion coefficients. While the online actor updates in \cite{jia2022policy}, the sampled advantage estimator in \cite{zhao2023policy}, and the control-randomization actor updates in \cite{denkert2025control} use differences of value-function estimates, our construction preserves the differential structure of both policy evaluation and advantage estimation: it retains the first- and second-order spatial derivatives of the critic and replaces the unknown generator by a PhiBE operator that uses discrete-time data.

\paragraph{Relation to discrete-time policy gradient.} Our analysis shares two central ingredients with policy gradient theory for discounted MDPs. First, the parametric policy gradient is expressed as an occupation-weighted expectation of the advantage multiplied by the policy score \cite{sutton1999policygradient}. Second, the convergence argument relies on a performance difference lemma, as in discrete-time policy-improvement analyses \cite{kakade2002approximately}. The approximation of the advantage-driven direction by a linear combination of policy scores is also related to compatible function approximation and natural policy gradient \cite{kakade2001natural,agarwal2021theory}. The main difference is that the continuous-time advantage is an instantaneous rate determined by the generator of the controlled diffusion, and the occupation measure accumulates rewards over continuous time. Accordingly, our finite-step update discretizes the policy-optimization dynamics rather than the controlled system.

\paragraph{Convergence of policy gradient in continuous control.} 
Existing convergence results can be distinguished by whether they concern policy flows, discrete updates in function space, or in finite-dimensional parametrized policies.
At the flow level, convergence guarantees have been established for regularized measure-valued controls \cite{siska2024gradient}, policy mirror descent for exit-time control \cite{sethi2025entropy}, feedback policy gradients \cite{zhou2025policy}, and coupled actor--critic dynamics \cite{zhou2024solving}. These results describe the zero-step-size limit of policy optimization. We likewise establish exponential convergence of an entropy-regularized policy flow, but consider infinite-horizon stationary randomized policies and use this result as the foundation for analyzing discrete parametrized updates.

For nonzero step sizes, \cite{reisinger2023linear} proves linear
convergence of a proximal policy gradient method for finite-horizon
feedback control, including stability under approximate gradients,
while \cite{KERIMKULOV2025104765} establishes convergence rates for
discrete mirror descent with measure-valued controls. These analyses
optimize directly over suitable function spaces of feedback controls
or adapted control processes, rather than over a prescribed
finite-dimensional policy class. Consequently, they do not explicitly
quantify the additional approximation error introduced by restricting
policy updates to such a parametrized class.

The existing convergence guarantees for parametrized updates among these works rely on LQ or closely related problem-specific structure \cite{giegrich2024convergence,zhang2025infinite,wang2026reinforcement}.
To the best of our knowledge, our work provides the first convergence bound for discrete, finite-dimensional parametrized policy-gradient updates in a general nonlinear diffusion control setting that explicitly accounts for critic error, compatible score-regression error, and the actor step size. Under suitable regularity and approximation-error assumptions, the bound establishes geometric decay, at a rate set by the actor step size, to an error neighborhood whose size is controlled by the critic and score-regression errors.

\paragraph{Data-driven and PDE-based continuous-time RL.} 
PhiBE \cite{zhuphibe_evaluation} embeds discrete transition information into a continuous-time PDE for policy evaluation, while Optimal-PhiBE \cite{zhu2025optimal} extends this construction to control and establishes exact recovery of the continuous-time optimal policy in the undiscounted LQ setting.  Earlier actor--critic and adaptive-dynamic-programming methods \cite{yang2014reinforcement,zhao2022data,jiang2012computational,kim2020hamilton} focus mainly on structured systems, with theoretical guarantees also available for learning unknown LQ dynamics from discrete observations \cite{basei2022logarithmic}. We use PhiBE as the critic within policy gradient and treat the discrepancy between the estimated and true advantages as the critic error appearing in the discrete convergence bound.

\section{Continuous-time reinforcement learning}\label{sec:background}
Let $\S=\R^d$ and $\A\subseteq\R^m$ be the action space, endowed with its Borel $\sigma$-algebra. We fix a reference measure $\nu$ on $\A$ and denote by $\mP(\A)$ the set of Borel probability measures on $\A$. A stationary randomized Markov control is a Markov kernel $\pi(\rd a|s)$ from $\S$ to $\A$. We denote by $\Pi$ the class of admissible stationary randomized Markov controls, specified in \Cref{ass:wellposed_policy_evaluation}. Each $\pi\in\Pi$ admits a strictly positive density, still denoted by $\pi(a|s)$, with respect to $\nu$, so that $\pi(\rd a|s)=\pi(a|s)\nu(\rd a).$
\smallbreak

The goal of this work is to solve the optimization problem
\begin{equation}\label{eq:J}
\sup_{\pi\in\Pi}\left\{J[\pi]:=\int_\S V^\pi(s)\rho_0(\rd s)\right\},
\end{equation}
where $\rho_0\in\mP(\S)$ is the prescribed distribution of the initial state, and the value function, under a fixed policy $\pi$, is defined as the infinite-horizon expected discounted reward
\begin{equation}\label{eq:value_function}
V^\pi(s):=\E\left[\int_0^\infty e^{-\beta t}r_\lambda^\pi(s_t)\,\rd t\,\Big|\,s_0=s\right],
\end{equation}
where $\beta>0$ is the discount factor, and the
entropy-regularized state reward is
\begin{equation}\label{eq:regularized_reward}
r_\lambda^\pi(s) := \int_\A \bigl(r(s,a)-\lambda\log\pi(a|s)\bigr) \pi(a|s)\nu(\rd a).
\end{equation}
Here $r:\S\times\A\to\R$ is the instantaneous reward and $\lambda\geq0$
is the entropy-regularization parameter. Thus, $r_\lambda^\pi(s)$ is the
policy average of the regularized state-action reward
$r(s,a)-\lambda\log\pi(a| s)$. The state process  $(s_t)_{t\geq0}$ in \eqref{eq:value_function} is the exploratory diffusion associated with $\pi$, that is
\begin{equation}\label{eq:average_SDE}
\rd s_t
=
b^\pi(s_t)\,\rd t
+
\sigma^\pi(s_t)\,\rd W_t,
\qquad
s_0=s,
\end{equation}
where $W$ is a $d$-dimensional Brownian motion, and for $b:\S\times \A\to \R^d$ and $\sigma:\S\times\A\to \R^{d\times q}$, the averaged drift and covariance in \eqref{eq:average_SDE} are
\begin{equation}\label{eq:averaged_coefficients}
b^\pi(s)
:=
\int_\A b(s,a)\pi(a|s)\nu(\rd a),
\qquad
\Sigma^\pi(s)
:=
\int_\A \Sigma(s,a)\pi(a|s)\nu(\rd a),
\end{equation}
with $\Sigma(s,a):=\sigma(s,a)\sigma(s,a)^\top$, and $\sigma^\pi(s):=(\Sigma^\pi(s))^{1/2}$ denotes the symmetric nonnegative square root of $\Sigma^\pi(s)$. The averaged SDE \eqref{eq:average_SDE} is the so-called exploratory formulation; see \Cref{remark:formulations_sde} for its connection with related formulations. Under \Cref{ass:wellposed_policy_evaluation}, stated below, \eqref{eq:average_SDE} admits a unique global strong solution for every initial condition; see \Cref{rem:wellposed_average_sde}.

\paragraph{Overview.} The rest of the paper is organized as follows. In \Cref{sec:policy_gradient_and_pdl}, we introduce the continuous-time advantage function and the performance difference lemma, which is used to characterize the first-order variation of $J$ and to derive the policy gradient theorem for parametrized policies $\pi_\theta$. \Cref{sec:convergence} establishes convergence guarantees for both the continuous-time policy gradient flow and its discrete parametrized counterpart. \Cref{sec:algorithm} develops the resulting data-driven algorithms, building on the PhiBE framework \cite{zhu2025optimal} to exploit the continuous-time structure using only discrete-time transition data. Finally, \Cref{sec:numerical_results} presents numerical experiments for a linear--quadratic regulator and a differential wheeled robot.

\section{Policy gradient}\label{sec:policy_gradient_and_pdl}
We now turn to the first-order variation of the objective $J$ with respect to the policy. The analysis relies on the value function and the infinitesimal generator associated with a fixed policy. In particular, under the regularity conditions introduced below, the value function $V^\pi$ defined in \eqref{eq:value_function} is characterized as the classical solution of the linear policy-evaluation equation
\begin{equation}\label{eq:Bellman_discounted}
\L_{b^\pi,\Sigma^\pi}V^\pi(s)- \beta V^\pi(s) + r_\lambda^\pi(s) =0,
\end{equation}
where the infinitesimal generator is
\begin{equation*}
\begin{aligned}
\L_{b^\pi,\Sigma^\pi}V(s) &:= b^\pi(s)\cdot\ns V(s) + \frac12\Sigma^\pi(s):\ns^2V(s).
\end{aligned}
\end{equation*}

For $k\geq 1$, let $\mathcal P_k(\A)$ denote the set of probability measures
on $\A$ with finite $k$th moment, and define $\mathcal P_k(\S)$ analogously.
We impose the following conditions on the model coefficients, the admissible
policies, and the corresponding policy-averaged quantities.

\begin{assumption}\label{ass:wellposed_policy_evaluation}
The following conditions hold.
\begin{enumerate}
    \item[(i)] \emph{Model coefficients.} The maps $b$, $\sigma$ and $r$ are continuous, and there are constants $C,L,\underline\theta>0$ such that $\Sigma(s,a)\succeq\underline\theta I_d$ and, for every $s,s'\in\S$, $a\in\A$ and $\varphi\in\{b,\sigma,r\}$,
\begin{align}
    |b(s,a)|+\|\sigma(s,a)\|&\leq C(1+|s|+|a|),
    \qquad
    |r(s,a)|\leq C(1+|s|+|a|)^2,
    \label{eq:coeff_growth}\\
    |\varphi(s,a)-&\varphi(s',a)|\leq L(1+|s|+|s'|+|a|)\,|s-s'| .
    \label{eq:coeff_lipschitz}
\end{align}
\item[(ii)]  \emph{Admissible policies.} Every $\pi\in\Pi$ has a strictly positive continuous density $\pi(a|s)$ with respect to $\nu$. Writing $\pi_s:=\pi(\cdot|s)$ and $\|f\|_k:=\int_\A(1+|a|^k)|f(a)|\nu(\rd a)$, there exist constants $C_\pi,M_\pi,L_\pi>0$ such that, for all $s,s'\in\S$ and $a\in\A$,
\begin{align}
    |\log\pi(a|s)|\leq C_\pi(1+|s|&+|a|)^2,
    \qquad
    \|\pi_s\|_8\leq M_\pi(1+|s|)^8,
    \label{eq:policy_moment}\\
    \|\pi_s-\pi_{s'}\|_8&\leq L_\pi(1+|s|+|s'|)^8|s-s'|,
    \label{eq:policy_weighted_lipschitz}\\
    \|\pi_s\log\pi_s-\pi_{s'}\log&\pi_{s'}\|_2\leq L_\pi(1+|s|+|s'|)^8|s-s'|.
    \label{eq:policy_entropy_lipschitz}
\end{align}
In particular, $\|\pi_s\log\pi_s\|_2\leq C(1+|s|)^4$, with $C$ depending only on $C_\pi$ and $M_\pi$.
\item[(iii)]  \emph{Averaged coefficients.} For every $\pi\in\Pi$, the averaged coefficients $b^\pi$ and $\sigma^\pi$ are twice continuously differentiable with bounded first and second derivatives, and the averaged reward $r_\lambda^\pi$ is twice continuously differentiable with
\begin{equation}\label{eq:closed_loop_reward_regularity}
    |\ns r_\lambda^\pi(s)|\leq C_\pi(1+|s|),
    \qquad
    \|\ns^2r_\lambda^\pi(s)\|\leq C_\pi .
\end{equation}
Moreover $\rho_0\in\mathcal P_8(\S)$ and the discount factor satisfies
\begin{equation}\label{eq:discount_condition}
    2\nu_\pi+\kappa_\pi<\beta,
\end{equation}
where $\kappa_\pi$ and $\nu_\pi$ are the exponential rates of \Cref{lem:closed_loop_rates}.
\end{enumerate}
\end{assumption}

Under \Cref{ass:wellposed_policy_evaluation}, for every $\pi\in\Pi$, equation \eqref{eq:Bellman_discounted} admits a unique classical solution $V^\pi$ that grows at most quadratically and whose first two derivatives grow at most linearly; see \Cref{prop:value_function_regularity}.

\subsection{Advantage function and policy gradient theorem}
We first introduce the continuous-time advantage function, which quantifies
the instantaneous gain of deviating from a policy $\pi$ at a given
state-action pair. The definition is motivated by the advantage function
in discrete-time reinforcement learning and by the continuous-time
advantage-rate formulation in \cite{jia2023q,zhao2023policy}.

Fix $s\in\S$ and $a\in\A$. Let $s_t^a$ be the state process satisfying
\[
    \rd s_t^a
    =
    b(s_t^a,a)\,\rd t
    +
    \sigma(s_t^a,a)\,\rd B_t,
    \qquad
    s_0^a=s,
\]
where $B$ is a $q$-dimensional Brownian motion. By \Cref{ass:wellposed_policy_evaluation}(i), this SDE admits a unique strong solution. We define
\begin{equation}\label{eq:advantage_definition}
\begin{aligned}
A^\pi(s,a)
:={}&
\lim_{\dt\to0^+}
\frac{1}{\dt}
\E\Bigg[
    \int_0^\dt
    e^{-\beta t}
    \parentheses{
        r(s_t^a,a)
        -
        \lambda\log\pi(a|s_t^a)
    }
    \,\rd t
    +e^{-\beta\dt}V^\pi(s_\dt^a)
    -V^\pi(s)
    \,\Big|\,
    s_0^a=s
\Bigg].
\end{aligned}
\end{equation}
Thus, $A^\pi(s,a)$ measures the first-order change in value obtained by
holding the action $a$ fixed over a short time interval before reverting
to the policy $\pi$. It is therefore an instantaneous advantage rate. Denote by $\mL $ the operator
\[
[\mL V](s,a)
:=
b(s,a)\cdot\ns V(s)
+
\frac12\Sigma(s,a):\ns^2V(s),
\]
for every $(s,a)\in\S \times\A$. The following proposition gives a pointwise representation of the advantage.

\begin{proposition}[Pointwise representation of the advantage]\label{prop:adv}
Assume \Cref{ass:wellposed_policy_evaluation}. Then the limit in \eqref{eq:advantage_definition} exists and
\begin{equation}\label{eq:advantage_representation}
A^\pi(s,a)
=
r(s,a)
-
\lambda\log\pi(a|s)
+
[\mL V^\pi](s,a)
-
\beta V^\pi(s).
\end{equation}
Moreover,
\begin{equation}\label{eq:centered_advantage}
    \int_\A A^\pi(s,a)\pi(a|s)\nu(\rd a)=0,
    \qquad s\in\S.
\end{equation}
\end{proposition}
Combining \eqref{eq:advantage_representation} with the growth conditions
on the model coefficients and the regularity of $V^\pi$, the advantage
has at most cubic growth in $(s,a)$, and its policy average in absolute
value has at most cubic growth in $s$; see
\eqref{eq:advantage_growth}--\eqref{eq:advantage_integrability}.
Together with the centering property \eqref{eq:centered_advantage},
these bounds provide the integrability needed for the variational
analysis below.

We now characterize how the global objective $J$
changes under a perturbation of the policy. Since $J$ aggregates rewards
along the entire discounted state trajectory, the local quantity
$A^\pi(s,a)$ is naturally weighted by the discounted frequency with
which the state $s$ is visited under $\pi$. This weighting is described
by the discounted occupancy measure.
\begin{definition}[Discounted occupancy measure]
    For each $\pi\in\Pi$, define the discounted occupancy measure $\rho^\pi$ on $\S$ by
\begin{equation}\label{eq:discounted_occupancy_measure}
\rho^\pi(B):=\beta\int_\S
\E\left[
    \int_0^\infty
    e^{-\beta t}\mathbf 1_B(s_t)
    \,\rd t
    \,\Big|\,
    s_0=s
\right]
\rho_0(\rd s),
\qquad
B\in\mathcal B(\S),
\end{equation}
where $s_t$ solves \eqref{eq:average_SDE}. Note that if the law of $s_t$ has a density $\rho^\pi(t,s)$, then the measure in \eqref{eq:discounted_occupancy_measure} has density $\rho^\pi(s) =\beta\int_0^\infty e^{-\beta t}\rho^\pi(t,s)\,\rd t$.
\end{definition}

\begin{lemma}[Performance difference]\label{lem:performance_difference}
Assume \Cref{ass:wellposed_policy_evaluation}, and let
$\pi,\widehat\pi\in\Pi$. Then
\begin{equation}\label{eq:performance_difference}
J[\widehat\pi]-J[\pi] = \frac{1}{\beta}\int_\S \left[\int_\A A^\pi(s,a)\widehat\pi(a|s)\nu(\rd a)-\lambda \KL{\widehat\pi_s}{\pi_s}\right]\rho^{\widehat\pi}(\rd s).
\end{equation}
\end{lemma}

\begin{remark}
     The performance-difference lemma measures the performance gap between two policies. A related identity was previously derived in \cite{zhao2023policy}. Our result provides a rigorous extension of this identity to a class of unbounded state--action models under explicit growth and regularity conditions. In particular, we allow the drift and diffusion coefficients to have linear growth and the reward to have quadratic growth.
\end{remark}

\begin{proof}[Proof sketch]
By the definition of the discounted occupancy measure,
\[
J[\widehat\pi] = \frac{1}{\beta} \int_\S r_\lambda^{\widehat\pi}(s) \rho^{\widehat\pi}(\rd s).
\]
On the other hand, applying the discounted resolvent identity to
$V^\pi$ along the dynamics induced by $\widehat\pi$ gives
\[
J[\pi]
=
\frac{1}{\beta}
\int_\S
\left(
    \beta V^\pi(s)
    -
    \L_{b^{\widehat\pi},\Sigma^{\widehat\pi}}V^\pi(s)
\right)
\rho^{\widehat\pi}(\rd s).
\]
Subtracting these two identities and expanding the
$\widehat\pi$-averaged reward and generator gives
\[
J[\widehat\pi]-J[\pi]
=
\frac{1}{\beta}
\int_\S\int_\A
\left(
    r(s,a)-\lambda\log\widehat\pi(a|s)
    +[\mL V^\pi](s,a)-\beta V^\pi(s)
\right)
\widehat\pi(a|s)\nu(\rd a)\rho^{\widehat\pi}(\rd s).
\]
Adding and subtracting $\lambda\log\pi(a|s)$ inside the integrand
and using the representation of $A^\pi$ in
\eqref{eq:advantage_representation} yields
\eqref{eq:performance_difference}.
\end{proof}

To formulate variations directly in the space of randomized policies, rather than through a particular finite-dimensional parametrization, we consider additive perturbations of the policy density.
\begin{definition}[Admissible policy direction]
\label{def:admissible_policy_direction}
Fix $\pi\in\Pi$. A continuous function
$\phi:\S\times\A\to\R$ is called an admissible policy direction at $\pi$
if $\int_\A \phi(a|s)\nu(\rd a)=0,$ for every $s\in\S,$ and there exists $\ve_0>0$ such that
\begin{equation}\label{eq:additive_policy_perturbation}
    \pi^\ve(a|s)
    :=
    \pi(a|s)+\ve\phi(a|s)
\end{equation}
belongs to $\Pi$ for every $|\ve|\leq\ve_0$, with all the estimates in \Cref{ass:wellposed_policy_evaluation} holding uniformly in $\ve$. In particular, the rates in \eqref{eq:discount_condition} can be chosen so that $\sup_{|\ve|\leq\ve_0}
    \left(
        2\nu_{\pi^\ve}+\kappa_{\pi^\ve}
    \right)
    <\beta.$ We denote the set of admissible policy directions at $\pi$ by $T_\pi\Pi$.
\end{definition}
Some useful consequences of this definition, together with a sufficient condition for admissibility in terms of the Radon--Nikodym derivative of the signed kernel $\phi(a|s)\nu(\rd a)$ with respect to $\pi(a|s)\nu(\rd a)$, are given in
\Cref{rem:admissible_policy_direction}.

\smallbreak
The following theorem characterizes the directional derivative of the objective functional along any admissible policy direction.
\begin{theorem}[Policy gradient theorem]\label{thm:PG}
Assume \Cref{ass:wellposed_policy_evaluation}. Let $\pi\in\Pi$ and
$\phi\in T_\pi\Pi$. Then the directional derivative of $J$ at $\pi$ in
the direction $\phi$ exists and satisfies
\begin{equation}\label{eq:PG_pi_discounted}
\left.
\frac{\rd}{\rd\ve}
J[\pi+\ve\phi]
\right|_{\ve=0}
=
\frac{1}{\beta}\int_\S\int_\A
A^\pi(s,a)\phi(a|s)
\nu(\rd a)
\rho^\pi(\rd s).
\end{equation}
\end{theorem}
\begin{remark}[Relation to existing policy-gradient formulas]
Continuous-time policy-gradient representations for parametrized policies have previously been established in \cite{jia2022policy,zhao2023policy}. The distinguishing feature of \Cref{thm:PG} is its formulation for arbitrary admissible additive policy directions, which provides a general representation of the Gâteaux derivative of $J$.
\end{remark}

\begin{proof}[Proof sketch.]
Let $\pi^\ve=\pi+\ve\phi$. By \Cref{lem:performance_difference}
\[
J[\pi^\ve]-J[\pi]
=
\frac{1}{\beta}
\int_\S
\left[
\int_\A A^\pi(s,a)\pi^\ve(a|s)\nu(\rd a)
-
\lambda\,\operatorname{KL}\!\left(\pi^\ve(\cdot|s)\,\middle\|\,\pi(\cdot|s)\right)
\right]
\rho^{\pi^\ve}(\rd s).
\]
Since $\int_\A A^\pi(s,a)\pi(a|s)\nu(\rd a)=0$, the first term is exactly
\[
\int_\A A^\pi(s,a)\pi^\ve(a|s)\nu(\rd a)
=
\ve\int_\A A^\pi(s,a)\phi(a|s)\nu(\rd a).
\]

Moreover, admissibility implies that $\phi=\pi g$ for a bounded
function $g$ satisfying $\int_\A g(s,a)\pi(a|s)\nu(\rd a)=0$. Hence $\pi^\ve=\pi(1+\ve g)$ and
\begin{align*}
\KL{\pi^\ve_s}{\pi_s}
&=
\int_\A
\pi(a|s)(1+\ve g(s,a))
\log(1+\ve g(s,a))\,\nu(\rd a)= \mathcal O(\ve^2),
\end{align*}
uniformly in $s$, where the first order term vanishes by the zero-mean property. Therefore,
\[
\frac{J[\pi^\ve]-J[\pi]}{\ve}
=
\frac{1}{\beta}
\int_\S
\left[
\int_\A A^\pi(s,a)\phi(a|s)\nu(\rd a)
+\mathcal O(\ve)
\right]
\rho^{\pi^\ve}(\rd s).
\]
Passing to the limit yields \eqref{eq:PG_pi_discounted}.
\end{proof}


\subsection{Parametric policy gradient}

We now extend the policy gradient analysis to a parametrized family. Let $\Theta\subseteq\R^p$ be a parameter domain and let $(\pi^\theta)_{\theta\in\Theta}\subseteq\Pi$ be a family of admissible policies. We impose the following regularity conditions on the parametrization.

\begin{assumption}[Parametric policies]\label{ass:assumption_pi_theta}
The map $\theta\mapsto\pi^\theta(a|s)$ is continuously differentiable for every $(s,a)\in\S\times\A$. Moreover, for every compact set $K\subset\Theta$, there exist $C_K>0$ and a nondecreasing modulus of continuity $\o_K$, with $\o_K(\varrho)\to0$ as $\varrho\to0$, such that, for all $\theta,\theta'\in K$, $s,s'\in\S$, and $a\in\A$,
\begin{align}
    |\nt\log\pi^\theta(a|s)|
    &\leq
    C_K(1+|s|^2+|a|^2),
    \label{eq:score_polynomial_growth}\\
    |\nt\log\pi^\theta(a|s)-\nt\log\pi^\theta(a|s')|
    &\leq
    C_K(1+|s|^2+|s'|^2+|a|^2)|s-s'|,
    \label{eq:score_state_lipschitz}\\
    |\nt\log\pi^\theta(a|s)-\nt\log\pi^{\theta'}(a|s)|
    &\leq
    \o_K(|\theta-\theta'|)(1+|s|^2+|a|^2).
    \label{eq:score_parameter_continuity}
\end{align}
In addition, all constants in \Cref{ass:wellposed_policy_evaluation} can be chosen uniformly over $\theta\in K$.
\end{assumption}

\begin{remark}[Gaussian policies]\label{rem:gaussian_policies}
The conditions above are designed to include the Gaussian class $\pi^\theta(\cdot|s)=\mathcal N(m_\theta(s),\Gamma_\theta)$, provided that on compact sets of parameters the covariances are uniformly positive definite and bounded, with bounded parameter derivatives, and the mean and its parameter derivatives grow at most linearly and are Lipschitz in the state. In particular, affine means $m_\theta(s)=K_\theta s+k_\theta$ are covered. The same remark applies to the score, which for a Gaussian policy is unbounded in both variables.
\end{remark}

Under these conditions, the finite-dimensional gradient of $\theta\mapsto J[\pi^\theta]$ admits the following representation.
\begin{theorem}[Parametric policy gradient]
\label{cor:parametric_policy_gradient}
Assume \Cref{ass:wellposed_policy_evaluation} and \Cref{ass:assumption_pi_theta}.
Then, for every $\theta\in\Theta$,
\begin{equation}\label{eq:PG_omega_discounted}
\nt J[\pi^\theta]
=
\frac{1}{\beta}\int_\S\int_\A
A^{\pi^\theta}(s,a)
\nt\log\pi^\theta(a|s)
\pi^\theta(a|s)
\nu(\rd a)
\rho^{\pi^\theta}(\rd s).
\end{equation}
Equivalently,
\begin{equation}\label{eq:PG_score_discounted}
\nabla_\theta J[\pi^\theta]
=
\frac{1}{\beta}
\E_{
\substack{
s\sim \rho^{\pi^\theta}\\
a\sim \pi^\theta(\cdot|s)
}}
\left[
A^{\pi^\theta}(s,a)
\nabla_\theta\log\pi^\theta(a|s)
\right].
\end{equation}
\end{theorem}

\begin{remark}
Characterization \eqref{eq:PG_omega_discounted} does not follow directly from \Cref{thm:PG} by the chain rule, since the parametric directions $\nt\pi^\theta\cdot z$ need not be admissible in the sense of \Cref{def:admissible_policy_direction}. The proof follows a similar strategy but additionally requires a first-order expansion of the parametric family, with a remainder that is $o(|z|)$ after integration over the action variable, up to a polynomially growing factor in the state.

Similar characterizations of the parametric policy gradient have been established in \cite{jia2022policy,zhao2023policy} under different assumptions. In \Cref{sec:algorithm}, we develop data-driven algorithms that preserve the differential structure of $A^\pi$ in \eqref{eq:PG_omega_discounted}. While the online actor--critic updates in \cite{jia2022policy} and the sampled advantage estimator in \cite{zhao2023policy} use increments of the value function along the time-discretized MDP, our construction retains its first- and second-order spatial derivatives, replacing the unknown generator by a PhiBE operator constructed from observed state increments.
\end{remark}

\section{A convergence analysis}\label{sec:convergence}
In this section, we present the convergence analysis for policy improvement in both the continuous-time policy gradient flow and the discrete parametrized policy update. We denote by $\tau$ the \emph{learning time}, which should be distinguished from the physical time variable $t$.

\subsection{Convergence analysis for the continuous-time policy gradient flow}

We define the continuous-time policy gradient flow by
\begin{equation}\label{eq:PG_flow}
\pta\pi^\tau(a|s)
=
A^\tau(s,a)\pi^\tau(a|s),
\end{equation}
where $A^\tau:=A^{\pi^\tau}$. A similar construction was proposed in \cite{sethi2025entropy,zhou2025learning}. We denote by $\rho^\tau:=\rho^{\pi^\tau}$ the discounted occupancy measure associated with $\pi^\tau$. The flow is compatible with the probability constraint defining the policy space. Indeed, by \eqref{eq:centered_advantage}, the direction $A^\tau\pi^\tau$ has zero total mass, while the multiplicative form of \eqref{eq:PG_flow} preserves strict positivity.  

 Here and below, $\KL{\widehat\pi_s}{\pi_s}$ denotes the KL divergence of $\widehat\pi(\cdot|s)$ relative to $\pi(\cdot|s)$, defined as:
\[
\KL{\widehat\pi_s}{\pi_s}
:=
\int_\A
\log\left(
\frac{\widehat\pi(a|s)}{\pi(a|s)}
\right)
\widehat\pi(a|s)
\nu(\rd a).
\]

\begin{remark}[Finiteness of the averaged relative entropy]\label{rem:relative_entropy_integrability}
Let $\pi,\widehat\pi\in\Pi$. Since $\pi(\cdot|s)>0$, we have $\widehat\pi_s\ll\pi_s$, so $\KL{\widehat\pi_s}{\pi_s}\in[0,\infty]$ is well defined. Bounding $|\log(\widehat\pi/\pi)|\leq|\log\widehat\pi|+|\log\pi|$ and using \Cref{ass:wellposed_policy_evaluation}(ii) and \eqref{eq:polynomial_integration}, we get $\KL{\widehat\pi_s}{\pi_s}\leq C_{\pi,\widehat\pi}(1+|s|)^4$. Hence, by \eqref{eq:occupancy_moment}, $\int_\S\KL{\widehat\pi_s}{\pi_s}\rho^{\pi'}(\rd s)<\infty$ for every $\pi'\in\Pi$.
\end{remark}

\begin{assumption}[Regularity of the flow]\label{ass:flow_regularity}
For every $T>0$, the family $(\pi^\tau)_{\tau\in[0,T]}$ satisfies \eqref{eq:score_polynomial_growth}--\eqref{eq:score_parameter_continuity} with $\theta$ replaced by $\tau$, $\nt$ by $\pta$, and $K$ by $[0,T]$. Moreover, the constants in \Cref{ass:wellposed_policy_evaluation} for $\pi^\tau$ can be chosen uniformly for $\tau\in[0,T]$.
\end{assumption}

\begin{theorem}[Exponential convergence of the policy gradient flow]
\label{thm:PG_flow_exponential_convergence}
Assume \Cref{ass:wellposed_policy_evaluation} and let $\lam>0$. Let $(\pi^\tau)_{\tau\geq0}\subseteq\Pi$ be a global classical solution of \eqref{eq:PG_flow} satisfying \Cref{ass:flow_regularity}. Suppose that there exists an optimal policy $\pi^*\in\Pi$, and let $\rho^*:=\rho^{\pi^*}$. Then, for every $\tau\geq0$,
\begin{equation}\label{eq:exponential_decay}
\begin{aligned}
&J[\pi^*]-J[\pi^\tau]+\frac{\lam}{\beta} \int_\S\KL{\pi_s^*}{\pi_s^\tau} \rho^*(\rd s) \leq e^{-\lam\tau}\left(J[\pi^*] -J[\pi^0] + \frac{\lam}{\beta}\int_\S \KL{\pi_s^*}{\pi_s^0} \rho^*(\rd s) \right).
\end{aligned}
\end{equation}
\end{theorem}

\begin{proof}[Proof sketch]
Define the Lyapunov functional
\[
\mathcal E(\tau):=J[\pi^*]-J[\pi^\tau]+\frac{\lambda}{\beta}\int_\S\KL{\pi_s^*}{\pi_s^\tau}\rho^*(\rd s).
\]
By the performance difference lemma,
\begin{equation*}
\mathcal E(\tau)
=
\frac{1}{\beta}
\int_\S\int_\A
A^\tau(s,a)\pi^*(a|s)
\nu(\rd a)\rho^*(\rd s).
\end{equation*}
The policy-gradient theorem also shows that the objective is nondecreasing along the flow:
\[
\partial_\tau J[\pi^\tau]
=
\frac{1}{\beta}
\int_\S\int_\A
|A^\tau(s,a)|^2
\pi^\tau(a|s)\nu(\rd a)\rho^\tau(\rd s)
\geq 0.
\]
On the other hand, the policy gradient flow
$\partial_\tau\pi^\tau=A^\tau\pi^\tau$ gives $\partial_\tau\log\pi^\tau=A^\tau$. Consequently,
\[
\partial_\tau
\int_\S
\KL{\pi_s^*}{\pi_s^\tau}\rho^*(\rd s)
=
-\int_\S\int_\A
A^\tau(s,a)\pi^*(a|s)
\nu(\rd a)\rho^*(\rd s)
=
-\beta\mathcal E(\tau).
\]
Therefore,
\[
\partial_\tau\mathcal E(\tau)
=
-\partial_\tau J[\pi^\tau]
-\lambda\mathcal E(\tau)
\leq
-\lambda\mathcal E(\tau).
\]
Gronwall's inequality yields
$\mathcal E(\tau)\leq e^{-\lambda\tau}\mathcal E(0)$,
which is \eqref{eq:exponential_decay}.
\end{proof}

This theorem shows both the optimality gap $J[\pi^*]-J[\pi^\tau]$ and the $\rho^*$-weighted KL divergence decay exponentially at rate $\lam$, which implies stronger regularization yields faster decay but toward the regularized optimum. The result parallels the exponential convergence obtained for continuous-time mirror descent in \cite{sethi2025entropy} and the gradient-flow analysis of \cite{zhou2025policy}, but follows directly from the performance difference lemma and the advantage, without a Bregman or mirror-map construction. The same Lyapunov functional underlies Theorem~\ref{thm:discrete_parametrized_policy_optimization}, with additional discretization and approximation errors.

\subsection{Convergence of discrete parametrized policy optimization}

We now discretize the policy-improvement dynamics and allow both policy-evaluation and policy-regression errors. The goal is to obtain an inexact discrete counterpart of the Lyapunov estimate in \Cref{thm:PG_flow_exponential_convergence}. We consider a discrete policy update with a parametrized policy
$(\pi^\theta)_{\theta\in\Theta}\subseteq\Pi$ and a parametric value
approximation $V^\xi$. At iteration $n$, let $\pi^{(n)}:=\pi^{\theta^{(n)}},$ $A^{(n)}:=A^{\pi^{(n)}},$ and $\rho^{(n)}:=\rho^{\pi^{(n)}}$. Given a critic parameter $\xi^{(n)}$, define
\[
\widehat A^{(n)}(s,a):=r(s,a)-\lam\log\pi^{(n)}(a|s)+[\mL V^{\xi^{(n)}}](s,a)-\beta V^{\xi^{(n)}}(s),
\]
which approximates $A^{(n)}$ by replacing $V^{\pi^{(n)}}$ with
$V^{\xi^{(n)}}$. The policy parameter is updated through
\begin{equation}\label{eq:theta_update}
\theta^{(n+1)} = \theta^{(n)} + \dta \, w^{(n)}.
\end{equation}
To approximate the policy-space direction $A^{(n)}\pi^{(n)}$ within the
parametric class, we fit the score representation
$w^\top\nt\log\pi^{(n)}$ to the estimated advantage $\widehat A^{(n)}$.
For example, we may take
\[
w^{(n)}\in\argmin_{w\in\R^p}\int_\S\int_\A\left|w^\top\nt\log\pi^{(n)}(a|s)-\widehat A^{(n)}(s,a)\right|^2\pi^{(n)}(a|s)\nu(\rd a)\rho^{(n)}(\rd s).
\]
We emphasize that the convergence analysis below applies to any $w^{(n)}\in\R^p$. In general, both the critic estimate $\hat A^{(n)}$ and its approximation by $w^{(n)\top}\nt\log\pi^{(n)}$ are inexact, which gives rise to two errors,
\[
\ve_A^{(n)}(s,a):=\widehat A^{(n)}(s,a)-A^{(n)}(s,a),\quad\text{and}\quad \veb^{(n)}(s,a):=
w^{(n)\top}\nt\log\pi^{(n)}(a|s)-\widehat A^{(n)}(s,a),\]
corresponding respectively to critic and score-regression errors. Hence $w^{(n)\top}\nt\log\pi^{(n)} =A^{(n)}+\ve_A^{(n)}+\veb^{(n)}$. We impose the following uniform regularity and error bounds. The assumptions below control both the first-order effect of these errors under the optimal policy and their mean-square magnitude along the iterates.

\begin{assumption}[Regularity and approximation errors] \label{assump:discrete_policy_optimization} There exist a compact set $\Theta_0\subset\Theta$ and $K,\ve_1,\ve_2>0$ such that:
\begin{enumerate}
\item For every $n\geq0$, the line segment joining $\theta^{(n)}$ and $\theta^{(n+1)}$ is contained in $\Theta_0$, and the second derivative in $\theta$ of $\log\pi^\theta(a|s)$ grows at most quadratically in $(s,a)$, uniformly for $\theta\in\Theta_0$.
\item For every $n\geq0$, $|w^{(n)}|\leq K$, and the critic and regression errors $\ve_A^{(n)}$ and $\veb^{(n)}$ grow at most cubically in $(s,a)$, uniformly in $n$.
\item For every $n\geq0$, \[ \left| \int_\S\int_\A \ve_A^{(n)}(s,a) \pi^*(a|s) \nu(\rd a) \rho^*(\rd s) \right| \leq\ve_1, \quad\text{and}\quad \left| \int_\S\int_\A \veb^{(n)}(s,a) \pi^*(a|s) \nu(\rd a) \rho^*(\rd s) \right| \leq\ve_1. \]
\item For every $n\geq0$, \[ \int_\S\int_\A \left( |\ve_A^{(n)}(s,a)|^2 + |\veb^{(n)}(s,a)|^2 \right) \pi^{(n)}(a|s) \nu(\rd a) \rho^{(n+1)}(\rd s) \leq \ve_2.
\]
\item Define the Fisher matrix
\[
F_n:=\int_\S\int_\A \nt\log\pi^{(n)}(a|s)\,\nt\log\pi^{(n)}(a|s)^\top\,
\pi^{(n)}(a|s)\,\nu(\rd a)\,\rho^{(n+1)}(\rd s)\in\R^{p\times p}.
\]
There exists $\mu>0$ such that $(w^{(n)})^\top F_n\, w^{(n)}\geq\mu\,|w^{(n)}|^2$ for every $n\geq0$.
\end{enumerate}
\end{assumption}

\begin{theorem}[Convergence of discrete parametrized policy optimization] \label{thm:discrete_parametrized_policy_optimization}
Assume \Cref{ass:wellposed_policy_evaluation}, \Cref{ass:assumption_pi_theta} and \Cref{assump:discrete_policy_optimization}, and let $\lam>0$. Suppose that there exists an optimal policy $\pi^*\in\Pi$, and let $\rho^*:=\rho^{\pi^*}$. Then there exists a constant $\dta_0>0$, independent of $n$ and $\dta$, such that, whenever $0<\dta\leq\dta_0$ and $\lam\dta\leq1$,
\begin{equation}\label{eq:main_result}
\begin{aligned}
&J[\pi^*]-J[\pi^{(n)}]+\frac{\lam}{\beta}\int_\S\KL{\pi_s^*}{\pi_s^{(n)}}\rho^*(\rd s)\\
&\qquad\leq(1-\lam\dta)^n\left(J[\pi^*]-J[\pi^{(0)}]+\frac{\lam}{\beta}\int_\S\KL{\pi_s^*}{\pi_s^{(0)}}\rho^*(\rd s)\right)+\frac{2\ve_1}{\beta}+\frac{\ve_2}{\lam\beta}.
\end{aligned}
\end{equation}
\end{theorem}

\begin{proof}[Proof sketch]
Set $\ve^{(n)}:=\ve_A^{(n)}+\veb^{(n)}$, so that $q^{(n)}:=w^{(n)\top}\nt\log\pi^{(n)}=A^{(n)}+\ve^{(n)}$. Taylor's formula along the segment $\vartheta_u=\theta^{(n)}+u\dta w^{(n)}$, $u\in[0,1]$, gives
\[
\log\frac{\pi^{(n+1)}}{\pi^{(n)}}=\dta\,q^{(n)}+R^{(n)},
\qquad
\pi^{(n+1)}=\pi^{(n)}+\dta\,q^{(n)}\pi^{(n)}+\widetilde R^{(n)},
\]
with second-order remainders proportional to $\dta^2|w^{(n)}|^2$. Let $\mathcal E_n$ be defined as
\begin{equation}\label{eq:definition_E_n}
    \mathcal E_n:=J[\pi^*]-J[\pi^{(n)}]+\frac{\lam}{\beta}\int_\S\KL{\pi_s^*}{\pi_s^{(n)}}\rho^*(\rd s).
\end{equation}
By the performance difference lemma, $\mathcal E_n=\frac1\beta\int_\S\int_\A A^{(n)}\pi^*\nu(\rd a)\rho^*(\rd s)$. We combine three facts: the first expansion inserted in the variation of the relative entropy with respect to $\pi^*$; the second expansion inserted in the performance difference lemma for $J[\pi^{(n+1)}]-J[\pi^{(n)}]$; and the centering $\int_\A q^{(n)}\pi^{(n)}\nu(\rd a)=0$. Together they yield the exact identity
\begin{equation}\label{eq:equality_En1}
    \mathcal E_{n+1}=(1-\lam\dta)\,\mathcal E_n+I_1+I_2+T_n,
\end{equation}
where
\[
I_1:=-\frac{\lam\dta}{\beta}\int_\S\int_\A\ve^{(n)}\pi^*\nu(\rd a)\rho^*(\rd s),
\qquad
I_2:=-\frac{\dta}{\beta}\int_\S\int_\A A^{(n)}q^{(n)}\pi^{(n)}\nu(\rd a)\rho^{(n+1)}(\rd s),
\]
and $T_n$ collects the second-order terms, and satisfies $|T_n|\leq C_*\dta^2|w^{(n)}|^2$. Using \Cref{assump:discrete_policy_optimization}(3) for $I_1$, the identity $A^{(n)}=q^{(n)}-\ve^{(n)}$ together with \Cref{assump:discrete_policy_optimization}(4) for $I_2$, and \Cref{assump:discrete_policy_optimization}(5) for $T_n$, we obtain
\[
I_1\leq\frac{2\lam\dta\ve_1}{\beta},\qquad
I_2\leq-\frac{\dta}{2\beta}\,w^{(n)\top}F_nw^{(n)}+\frac{\dta\ve_2}{\beta},\qquad
T_n\leq\frac{C_*\dta^2}{\mu}\,w^{(n)\top}F_nw^{(n)}.
\]
Therefore, for $\dta\leq\mu/(2\beta C_*)$ the term $T_n$ is absorbed by the negative part of the bound on $I_2$, and substituting into \eqref{eq:equality_En1} we obtain
\begin{equation}\label{eq:clear_bound_en1}
    \mathcal E_{n+1}\leq(1-\lam\dta)\,\mathcal E_n+\dta\,\frac{2\lam\ve_1+\ve_2}{\beta}.
\end{equation}
Iterating this recursion, which is possible since $0<\lam\dta\leq1$, gives \eqref{eq:main_result}.
\end{proof}

\begin{remark}\label{rem:discrete_convergence}
Compared with the policy-flow analyses in \cite{zhou2025policy,sethi2025entropy}, \eqref{eq:main_result} quantifies the effect of finite actor steps. It also includes the score-regression error induced by policy parametrization, extending beyond the function-space updates studied in \cite{reisinger2023linear,KERIMKULOV2025104765}. The result covers the nonlinear dynamics considered here under the stated assumptions, whereas \cite{giegrich2024convergence,zhang2025infinite,wang2026reinforcement} exploit LQ structure and \cite{agarwal2021theory} studies compatible approximation in discrete-time MDPs.

\Cref{assump:discrete_policy_optimization}(5) can be removed. However, in that case the term $T_n$ in \eqref{eq:equality_En1} can no longer be absorbed by $I_2$. Using only $|w^{(n)}|\leq K$, it is bounded by $C_*K^2\dta^2$, which adds this term to the right-hand side of \eqref{eq:clear_bound_en1} and, after iteration, an additional residual of order $\dta/\lam$ in \eqref{eq:main_result}; see \Cref{rem:no_fisher}. This extra term cannot be avoided in general, as the following example shows. Consider the entropy-regularized LQR of \Cref{subsec:LQR_example} in the one-dimensional case, with the two parametrizations $\pi^\theta_1(\cdot|s)=\mathcal N\big((\theta_1+\theta_2^2)s,\ \lam/N\big)$ and $\pi^\theta_2(\cdot|s)=\mathcal N\big(\theta s,\ \lam/N\big)$. Both describe the same family of Gaussian policies, which contains the optimal policy. The Fisher matrix associated with $\pi^\theta_1$ has rank one and is singular for every $\theta$, whereas for $\pi^\theta_2$ the Fisher matrix is a positive scalar. In this example the advantage is available in closed form, so $\ve_A^{(n)}=0$, and $w^{(n)}$ can be chosen explicitly so that $\veb^{(n)}=0$. Consequently, \Cref{assump:discrete_policy_optimization}(3)--(4) hold for every $\ve_1,\ve_2>0$. 
\begin{figure}[ht]
    \centering
    \begin{subfigure}[t]{0.49\textwidth}
        \centering
        \includegraphics[width=0.73\textwidth]{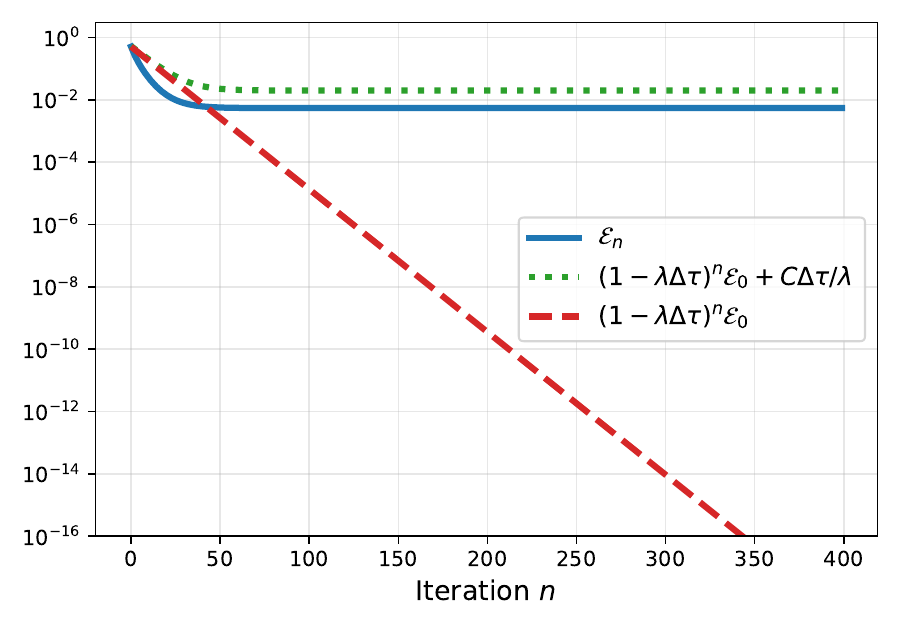}
        \caption{$\pi^\theta_1$ (singular Fisher).}
        \label{fig:fisher_singular_a}
    \end{subfigure}
    \hfill
    \begin{subfigure}[t]{0.49\textwidth}
        \centering
        \includegraphics[width=0.73\textwidth]{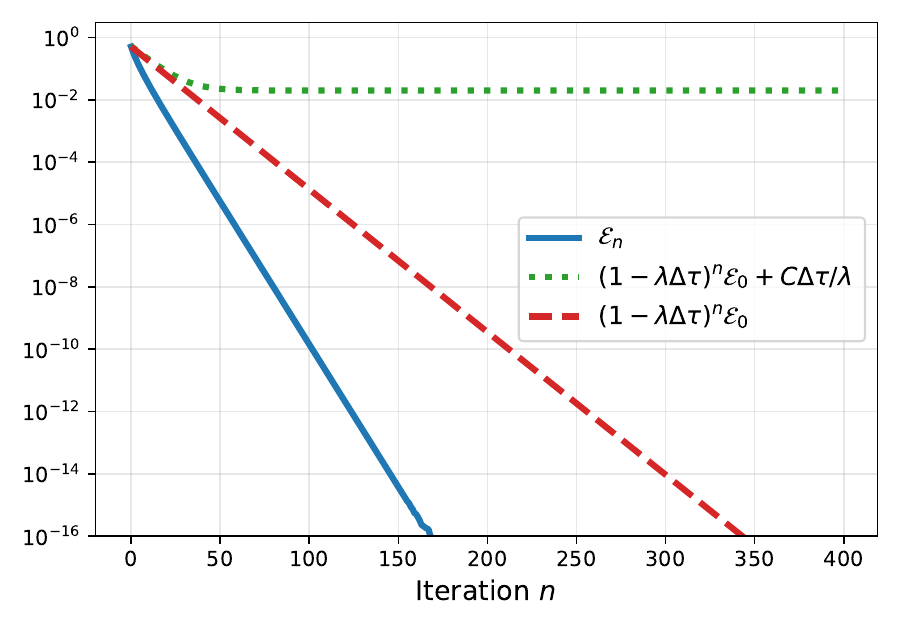}
        \caption{$\pi^\theta_2$ (non-singular Fisher).}
        \label{fig:fisher_singular_b}
    \end{subfigure}
   \caption{Convergence of $\mathcal E_n$ with vanishing errors.}
    \label{fig:fisher_singular}
\end{figure}
\Cref{fig:fisher_singular} reports the resulting values of $\mathcal{E}_n$ defined in \eqref{eq:definition_E_n}.

For the singular parametrization $\pi^\theta_1$, $\mathcal E_n$ converges to a positive limit while $(1-\lam\dta)^n\mathcal E_0\to0$, even though all errors vanish. Hence \eqref{eq:main_result} cannot hold without \Cref{assump:discrete_policy_optimization}(5), whereas the estimate $\mathcal E_n\leq(1-\lam\dta)^n\mathcal E_0+C\dta/\lam$ remains valid. For $\pi^\theta_2$, whose Fisher matrix is uniformly positive, $\mathcal E_n$ stays below $(1-\lam\dta)^n\mathcal E_0$, as predicted by \eqref{eq:main_result}.
\end{remark}

\color{black}

\section{Data driven algorithm}\label{sec:algorithm}

\subsection{Discrete observation setting}

We consider a continuous-time controlled system with unknown coefficients $b$ and $\sigma$, observed only at discrete times. The learner has access to transition data
\begin{equation}\label{traj-data}
    \{s_i,a_i,r_i,s_i'\}_{i=1}^n,
\end{equation}
where $r_i=r(s_i,a_i)$ and $s_i'$ denotes the state observed after an interval of length $\dt$. Each transition is generated by
\begin{equation}\label{eq:controlled_SDE}
    \rd s_t = b(s_t,a_t)\,\rd t + \sigma(s_t,a_t)\,\rd B_t,
\end{equation}
with the action $a_t=a_i$ held fixed throughout the observation interval. Neither $b$, $\sigma$, nor $r$ is assumed known.

For $(s,a)\in\S\times\A$, denote the corresponding transition kernel by
\begin{equation}\label{discrete dynamics}
\rho_{\Delta t}(B\mid s,a):=\mathbb P\!\left(s_{t+\Delta t}\in B\,\middle|\, s_t=s,\;a_\tau=a,\ \tau\in[t,t+\Delta t)\right),\quad B\in\mathcal B(\S).
\end{equation}
The data in \eqref{traj-data} may consist of one or several trajectories, or independent transition pairs.

Our objective remains the continuous-time control problem \eqref{eq:J}, and $\dt$ is only the observation interval. This distinguishes our setting from a discrete-time MDP, where the transition kernel is primitive and the objective is defined on the observation grid. Here, our goal is to exploit the continuous-time structure using only discrete observations.

\subsection{PhiBE}

We follow the PhiBE framework of \cite{zhuphibe_evaluation,zhu2025optimal}, which exploits the continuous-time structure using discrete transition data. Given the transition kernel $\rho_\dt$ defined in \eqref{discrete dynamics}, the state-action coefficients are defined by
\begin{equation}\label{eq:def_bhat_sigmahat}
    \hat{b}(s,a) = \E_{s'\sim\rho_\dt(\cdot|s,a)}\l[\tfrac{s'-s}{\dt}\,\middle|\,s,a\r], \quad
    \hat{\Sigma}(s,a) = \E_{s'\sim\rho_\dt(\cdot|s,a)}\l[\tfrac{(s'-s)(s'-s)^\top}{\dt}\,\middle|\,s,a\r].
\end{equation}
These coefficients define the state-action PhiBE generator
\[
    [\mLhat V](s,a)
    :=
    \hat b(s,a)\cdot\nabla_s V(s)
    +
    \frac12\hat\Sigma(s,a):\nabla_s^2 V(s).
\]
For a fixed policy $\pi\in\Pi$, define the averaged coefficients
\[
    \hat b^\pi(s)
    :=
    \int_\A \hat b(s,a)\pi(a|s)\nu(\rd a),
    \qquad
    \hat\Sigma^\pi(s)
    :=
    \int_\A \hat\Sigma(s,a)\pi(a|s)\nu(\rd a).
\]
The corresponding policy-averaged PhiBE generator is defined by
\[
    \mathcal L_{\hat b^\pi,\hat\Sigma^\pi}V(s)
    :=
    \int_\A[\mLhat V](s,a)\pi(a|s)\nu(\rd a).
\]
The PhiBE policy-evaluation equation is then given by
\begin{equation}\label{eq:phibe}
    \beta\hat V^\pi(s)
    =
    r_\lambda^\pi(s)
    +
    \mathcal L_{\hat b^\pi,\hat\Sigma^\pi}\hat V^\pi(s).
\end{equation}

The coefficients in \eqref{eq:def_bhat_sigmahat} can be estimated directly from discrete transitions. For each observation $(s_i,a_i,r_i,s_i')$, define
\begin{equation}\label{eq:approximation_b_and_simga_hat}
    \hat b_i:=\frac{s_i'-s_i}{\dt},
    \qquad
    \hat\Sigma_i:=\frac{(s_i'-s_i)(s_i'-s_i)^\top}{\dt}.
\end{equation}
These quantities provide sample contributions to the conditional expectations in \eqref{eq:def_bhat_sigmahat}. In particular, they can be averaged implicitly through empirical Galerkin or residual formulations of \eqref{eq:phibe}, yielding the data-driven policy-evaluation methods developed below.

\begin{remark}\label{remark:formulations_sde}
    The formulation \eqref{eq:average_SDE} is a Markovian realization of the policy-averaged generator. Consider the controlled dynamics \eqref{eq:controlled_SDE}, where actions are sampled from the randomized feedback policy at discrete decision times and held fixed between consecutive decision times. Under suitable regularity conditions, as the sampling interval tends to zero, the resulting controlled process converges weakly to the exploratory dynamics \eqref{eq:average_SDE}; see \cite{jia2026accuracy,bender2025continuous} and the related discussion in \cite{carmona2025reconcilingdiscretetimemixedpolicies}.
\end{remark}

\smallbreak
\noindent\textbf{Why PhiBE.} PhiBE has two main advantages. First, it does not require the unknown coefficients $b$ and $\sigma$ in \eqref{eq:J}--\eqref{eq:average_SDE}. This is important because the underlying continuous-time dynamics are generally not identifiable from discrete observations: even for linear systems, infinitely many continuous-time models may induce the same transition kernel $\rho_\dt$ \cite{zhu2025optimal}. Second, unlike the discrete-time MDP surrogate
\[
    \sup_\pi \E\l[\sum_{k=0}^\infty e^{-\beta k\dt} r(s_k,a_k)\dt  \Big|  s_0=s\r], \quad s_{k+1}\sim\rho_\dt(\cdot|s_k,a_k), \quad a_k\sim\pi(\cdot|s_k),
\]
PhiBE embeds the local generator structure of $\rho_\dt$ through $\mLhat$ instead of treating the kernel as primitive. This yields more accurate policies when the reward is oscillatory and the dynamics vary smoothly---natural in physical applications---and, in the undiscounted LQR case, PhiBE recovers the optimal policy exactly, whereas the MDP surrogate is only first-order accurate in $\dt$ \cite{zhu2025optimal}.

Equation \eqref{eq:phibe} can be solved in two ways. A \emph{model-based} approach first estimates $\rho_\dt$ (or $\hat b,\hat\Sigma$) and then solves the resulting PDE using these
estimated quantities. We instead pursue a \emph{model-free} approach that works directly with the samples $(s_i,a_i,r_i,s_i')$ through \eqref{eq:approximation_b_and_simga_hat}.

\subsection{Model-free policy gradient algorithm}

We optimize a parametrized policy $\{\pi_\th\}_{\th\in\Theta}$ through
\begin{equation}\label{eq:vanilla_direction}
    \th_{k+1}=\th_k+\eta w_k,
\end{equation}
where $w_k$ may correspond to vanilla policy gradient, natural policy gradient, TRPO, or PPO; see \Cref{subsec:gradiet_based_opt}. These updates require an estimate of the advantage of the current policy.

Given the PhiBE critic $\hat V^\pi$ from \eqref{eq:phibe}, we define the PhiBE advantage
\begin{equation}\label{eq:adv_phibe}
\hat A^\pi(s,a) := r(s,a)-\lambda\log\pi(a|s) +[\mLhat\hat V^\pi](s,a)-\beta\hat V^\pi(s).
\end{equation}
Thus, PhiBE provides the critic and advantage directly from discrete transition data while retaining the continuous-time generator structure.

For linear value-function classes, \eqref{eq:phibe} can be solved by the continuous-time counterpart of least-squares temporal difference; for nonlinear classes, one may use empirical residual minimization or the PhiBE-difference and BFF-PhiBE methods \cite{zhuphibe_evaluation}.

\smallbreak

In practice, at iteration $k$ the data are collected under the current policy $\pi_{\th_k}$, that is, $\mathcal D_k=\{(s_i, a_i\sim \pi_{\th_k}(\cdot|s_i), r_i, s_i')\}_{i=1}^{n}.$ The evaluation step solves \eqref{eq:phibe} with the coefficients \eqref{eq:approximation_b_and_simga_hat} (see \Cref{remark:on-policy-evaluation}), giving $\hat V^{\pi_{\th_k}}$. Denoting by $\Delta s_i:=s_i'-s_i$, we estimate the advantage at each sample by
\begin{equation}\label{eq:empirical_advantage}
    \widehat A^{\pi_{\th_k}}(s_i,a_i):=r_i-\lambda\log\pi_{\th_k}(a_i|s_i)+\frac{\Delta s_i}{\dt}\cdot\nb_s\hat V^{\pi_{\th_k}}(s_i)+\frac{1}{2\dt}\Delta s_i^\top\nb_s^2\hat V^{\pi_{\th_k}}(s_i)\Delta s_i-\beta\hat V^{\pi_{\th_k}}(s_i),
\end{equation}
whose first- and second-order increment terms estimate the drift and diffusion contributions, respectively. The vanilla direction and update are
\begin{equation}\label{eq:on-policy_empirical_gradient}
    \widehat w_k:=\frac{1}{\beta n}\sum_{i=1}^n \widehat A^{\pi_{\th_k}}(s_i,a_i)\,\nb_\th\log\pi_{\th_k}(a_i|s_i),\qquad \th_{k+1}=\th_k+\eta\widehat w_k.
\end{equation}
If the samples come from finite trajectories rather than the discounted occupancy measure, the factor $\frac1n$ can be replaced by the weights $\beta e^{-\beta t_i}\dt$. Alternative update rules, such as natural policy gradient, TRPO, and PPO (see \Cref{subsec:gradiet_based_opt}), and other evaluation schemes replace these two steps without further change. The LQR specialization is Algorithm \ref{alg:on_policy_lqr}.

\begin{remark}[Solving the PhiBE evaluation equations]\label{remark:on-policy-evaluation}
The evaluation PDE \eqref{eq:phibe} has data-accessible coefficients and can be solved by any PDE method. We use a linear Galerkin approximation, detailed in \Cref{subsec:policy_eval} and implemented for the LQR example in \Cref{subsec:LQR_example}; neural-network solvers are equally admissible, and a PINN is used in the robot example of \Cref{subsec:example_differential_wheeled}.
\end{remark}

\begin{algorithm}
\caption{Model-free continuous-time policy gradient from discrete data}
\label{alg:general_phibe_pg}
\begin{algorithmic}[1]
\State \textbf{Input:} policy class $\{\pi_\th\}_{\th\in\Theta}$, initial parameter $\th_0$, learning rate $\eta$, observation step $\dt$, number of iterations $K$.
\For{$k=0,1,\ldots,K-1$}
    \State \textbf{Data.} Collect fresh transitions $\mathcal D_k=\{(s_i,a_i,r_i,s_i')\}_{i=1}^n$ at resolution $\dt$, with $a_i\sim\pi_{\th_k}(\cdot|s_i)$.
    \State \textbf{Local coefficients.} Compute the estimators $\hat b_i$ and $\hat\Sigma_i$ from \eqref{eq:approximation_b_and_simga_hat}.
    \State \textbf{Critic.} Solve the PhiBE evaluation equation \eqref{eq:phibe} using the collected transitions, obtaining $\hat V^{\pi_{\th_k}}$.
    \State \textbf{Advantage.} Evaluate \eqref{eq:empirical_advantage}.
    \State \textbf{Actor.} Compute $\widehat w_k$ from \eqref{eq:on-policy_empirical_gradient} and update $\th_k$ using PG, NPG, TRPO, or PPO; see \Cref{subsec:gradiet_based_opt}.
\EndFor
\State \textbf{Return:} $\th_K$.
\end{algorithmic}
\end{algorithm}

\subsection{Online model-free actor--critic}\label{subsec:online_actor_critic}

The method in \Cref{alg:general_phibe_pg} evaluates the current policy using a batch of transitions before updating the actor. We consider an online algorithm in which the critic and actor are updated from each new transition. Let $\hat V_\xi(s)=\xi^\top\Phi(s)$ be the linear value approximation considered in \Cref{subsec:policy_eval}. Let $(s_k,a_k,r_k,s_k')$ be a transition with $s_k\sim\rho^{\pi_{\th_k}}$ and $a_k\sim\pi_{\th_k}(\cdot|s_k)$. Evaluating $\widehat A^{\pi_{\th_k}}(s_k,a_k)$ using the current critic $\hat V_{\xi_k}$, we update the critic by
\begin{equation}\label{eq:online_critic_update}
\xi_{k+1}=\xi_k+\alpha_k\widehat A^{\pi_{\th_k}}(s_k,a_k)\Phi(s_k).
\end{equation}

The update \eqref{eq:online_critic_update} is a stochastic approximation of the Galerkin orthogonality condition $\E[\widehat A^{\pi_{\th_k}}(s,a)\Phi(s)]=0$, with the policy held fixed in the corresponding evaluation problem. The same advantage estimate provides a single-transition approximation of the policy gradient in \eqref{eq:PG_score_discounted}, leading to the actor update
\begin{equation}\label{eq:online_actor_update}
\th_{k+1}=\th_k+\frac{\eta_k}{\beta}\widehat A^{\pi_{\th_k}}(s_k,a_k)\nb_\th\log\pi_{\th_k}(a_k|s_k).
\end{equation}

We choose the critic and actor learning rates $\alpha_k$ and $\eta_k$, respectively, such that $\eta_k/\alpha_k\to0$, so that the critic evolves on a faster time scale. The separation of learning rates allows the critic to track the policy-evaluation problem associated with a slowly evolving actor, following the two-timescale stochastic approximation framework \cite{konda2003actorcritic,zeng2024twotimescale}. 

PPO can also be used online: the critic is updated after each transition, while the actor parameter $\theta$ is updated through a few steps on the clipped objective using a short window of fresh current-policy data.

\begin{algorithm}
\caption{Online model-free actor--critic}\label{alg:online_actor_critic}
\begin{algorithmic}[1]
\State \textbf{Input:} initial parameters $\th_0,\xi_0$, learning rates $(\alpha_k)_{k\geq0}$ and $(\eta_k)_{k\geq0}$, observation interval $\dt$, number of updates $K$.
\For{$k=0,1,\ldots,K-1$}
    \State \textbf{Data.} Obtain $(s_k,a_k,r_k,s_k')$, with $s_k\sim\rho^{\pi_{\th_k}}$ and $a_k\sim\pi_{\th_k}(\cdot|s_k)$.
    \State \textbf{Local coefficients.} Compute $\hat b_k$ and $\hat\Sigma_k$ from \eqref{eq:approximation_b_and_simga_hat}.
    \State \textbf{Advantage.} Evaluate $\widehat A^{\pi_{\th_k}}(s_k,a_k)$ using $\hat V_{\xi_k}$.
    \State \textbf{Critic.} Update $\xi_{k+1}$ by \eqref{eq:online_critic_update}.
    \State \textbf{Actor.} Update $\th_{k+1}$ by \eqref{eq:online_actor_update}.
\EndFor
\State \textbf{Return:} $\th_K,\xi_K$.
\end{algorithmic}
\end{algorithm}

\section{Numerical results}\label{sec:numerical_results}

We evaluate the proposed methods on two problems: the entropy-regularized linear--quadratic regulator (LQR), and a nonlinear differential wheeled robot. In the LQR experiments we compare vanilla policy gradient (PG), natural policy gradient (NPG), TRPO, and PPO; the robot experiments use PPO. 

\subsection{Entropy-regularized LQR}\label{subsec:LQR_example}

We consider the entropy-regularized linear--quadratic regulator, for which the drift, diffusion, and running reward are given by
\[
    b(s,a)=As+Ba,
    \qquad
    \sigma(s,a)=\sigma\in\R,
    \qquad
    r(s,a)=-\frac12\left(Ms^2+Na^2\right).
\]
We assume $\beta>2A$, $M,N,\lambda>0$, and $B\neq0$. The optimal value is $V^*(s)=\frac12k_2^*s^2+k_1^*s+k_0^*$, and the optimal policy is $\pi^*(a|s)=
    \mathcal N\left(
        \frac{B}{N}(k_2^*s+k_1^*),\frac{\lambda}{N}
    \right),$ where
\begin{equation*}
\begin{aligned}
    k_2^*&=\frac{N}{B^2}\left[
        \frac\beta2-A-
        \sqrt{\left(A-\frac\beta2\right)^2+\frac{MB^2}{N}}
    \right],\quad
    k_1^*&=0,\quad
    k_0^*&=\frac{\lambda}{2\beta}
        \log\left(\frac{2\pi\lambda}{N}\right)
        +\frac{\sigma^2}{2\beta}k_2^*.
\end{aligned}
\end{equation*}

\paragraph{Policy class and error metric.}
We consider the policy parametrization $\pi_\omega(a|s)=\mathcal N(a\mid\omega_1s+\omega_2,\omega_3^2)$, with $\omega=(\omega_1,\omega_2,\omega_3)\in\R\times\R\times\R_+$. For a fixed policy $\pi_\omega$ satisfying $\frac{\beta}{2}>A+B\omega_1$, the corresponding value function is $V^\omega(s)=\frac12k_2^\omega s^2+k_1^\omega s+k_0^\omega$, where the coefficients $k_2^\omega$, $k_1^\omega$, and $k_0^\omega$ depend on $\omega$ and are given explicitly in \Cref{subsec:lqr_setting}.

\paragraph{Critic and advantage construction.} The policy class is common to all experiments, whereas the critic and advantage depend on the available information. In the known dynamics examples we evaluate analytically the advantage from \eqref{eq:advantage_representation}. With a known transition law we use the LQR specialization of PhiBE, which computes $\hat b$ and sets $\hat\Sigma=0$. The actor update in every case uses the discounted state weighting in \eqref{eq:PG_score_discounted}; only the construction of the critic and advantage changes.

\paragraph{Discounted policy gradient samples.} To approximate \eqref{eq:PG_score_discounted}, we draw $\tau_i\sim\operatorname{Exp}(\beta)$ independently for $i=1,\ldots,n$, simulate $S_{\tau_i}$ under the current relaxed policy from $S_0\sim\rho_0$, and draw $A_i\sim\pi_\omega(\cdot\mid S_{\tau_i})$. Since $S_{\tau_i}\sim\rho^{\pi_\omega}$, the empirical estimator is
\begin{equation}\label{eq:lqr_mc_gradient}
    \widehat{\nabla_\omega J}(\omega)
    :=
    \frac{1}{\beta n}
    \sum_{i=1}^n
    A^{\pi_\omega}(S_{\tau_i},A_i)
    \nabla_\omega\log\pi_\omega(A_i\mid S_{\tau_i}).
\end{equation}
It is unbiased when the exact advantage is used. In the linear model, $S_{\tau_i}$ is sampled from its exact distribution, so no Euler--Maruyama approximation is needed. In the data-driven experiments the exact advantage in \eqref{eq:lqr_mc_gradient} is replaced by its PhiBE estimate.

\paragraph{Default parameters.} Unless stated otherwise, $A=-1$, $B=1$, $\sigma=M=N=\beta=1$, and $\lambda=0.1$. In the known-dynamics and known-transition experiments, we take $\rho_0=\mathcal N(0,1)$ and $\omega^0=(0.5,1,1)$. Each actor iteration uses $n=10{,}000$ discounted-occupancy samples. The update parameters are $\eta_{\rm PG}=\eta_{\rm NPG}=0.1$, $\delta_{\rm TRPO}=0.005$, and $\eta_{\rm PPO}=0.005$, with five PPO epochs and $\epsilon=0.2$.

\subsubsection{Known dynamics}\label{known_dynamics_results}

\begin{figure}
    \centering
    \begin{subfigure}[t]{0.35\textwidth}
        \centering
        \includegraphics[width=\textwidth]{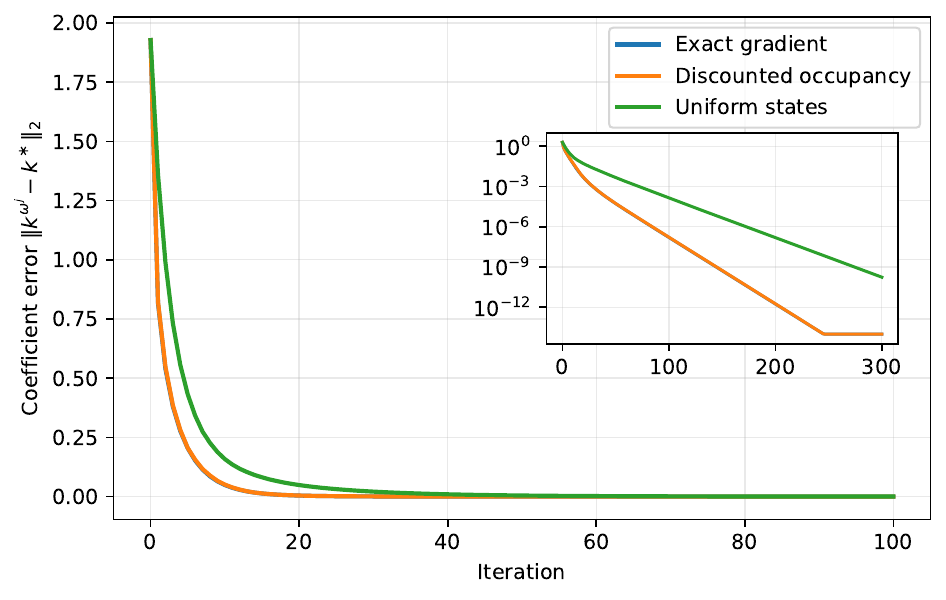}
        \caption{Exact, discounted-occupancy, and uniform-state gradients.}
        \label{known_dy_gradient_comparison}
    \end{subfigure}
    \begin{subfigure}[t]{0.35\textwidth}
        \centering
        \includegraphics[width=\textwidth]{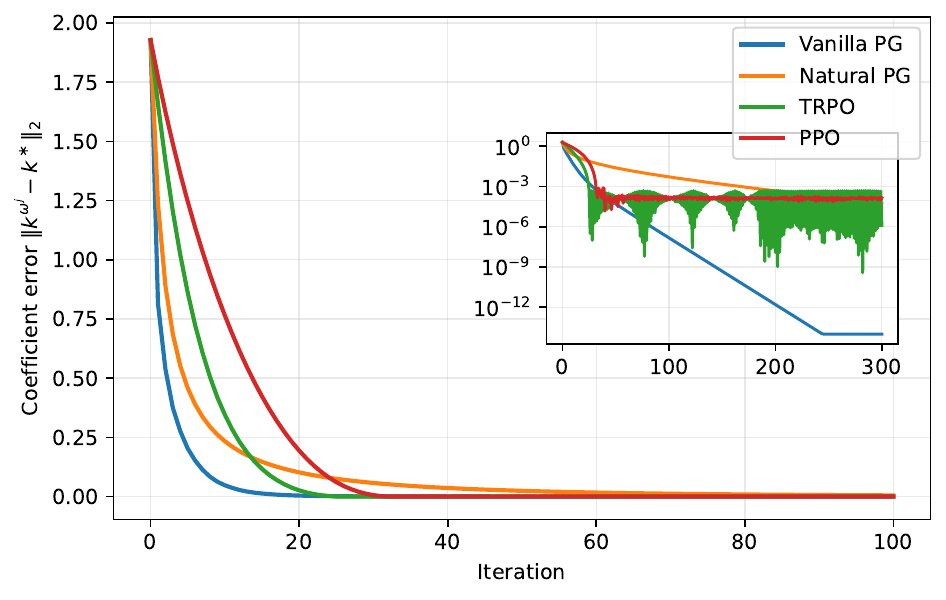}
        \caption{Default configuration.}
        \label{known_dy_default}
    \end{subfigure}
    \caption{Known-dynamics experiments: gradient sampling and default configuration.}
    \label{fig:known_dy_changedynamics}
\end{figure}

Figure~\ref{fig:known_dy_changedynamics}\subref{known_dy_gradient_comparison} compares the exact policy gradient with the Monte Carlo estimators obtained from discounted-occupancy and uniform state sampling. The discounted-occupancy estimator, which is unbiased by construction, is almost indistinguishable from the exact gradient. In contrast, uniform state sampling produces a systematically slower curve. Although it still converges in this scalar example, it uses a different state weighting and therefore gives a biased estimate of the policy gradient. Figure~\ref{fig:known_dy_changedynamics}\subref{known_dy_default} compares the four policy optimization methods under the default configuration. All methods rapidly reduce the error, with differences in their long-run behavior becoming visible mainly in the logarithmic inset.

Across the additional experiments, we observe that the stability of the policy update depends strongly on the effective step size relative to the scaling of the dynamics. Changes in $A$ mainly affect the transient behavior, whereas larger values of $B$ or $\sigma$ can make a fixed vanilla-PG step substantially less stable. NPG, TRPO, and PPO are more robust under these changes. Smaller effective steps improve stability at the cost of slower convergence, while larger steps accelerate the initial decrease but can produce oscillations or leave the admissible region. In the tested PPO configurations, changing the clipping ratio has a comparatively small effect. We refer to \Cref{sec:extra_num_ex} for the corresponding parameter analysis.

\subsubsection{Optimal-PhiBE with known discrete-time dynamics}
\label{sec:lqr_phibe}

We next assume that the continuous-time coefficients are unavailable but the transition law over $\Delta t=0.1$ is known:
\begin{equation}\label{eq:explicit_LQR_transition}
    S_{t+\Delta t}\mid(S_t=s,a_t=a)
    \sim\mathcal N\left(
        e^{A\Delta t}s+
        \frac{e^{A\Delta t}-1}{A}Ba,
        \frac{e^{2A\Delta t}-1}{2A}\sigma^2
    \right).
\end{equation}
Its first increment moment gives
\[
    \hat b(s,a)=\hat As+\hat Ba,
    \qquad
    \hat A=\frac{e^{A\Delta t}-1}{\Delta t},
    \qquad
    \hat B=\frac{B(e^{A\Delta t}-1)}{A\Delta t}.
\]
Following the LQR construction of Optimal-PhiBE~\cite{zhu2025optimal}, we set $\hat\Sigma=0$ and use the drift-only surrogate.
This is a deliberate specialization: the additive diffusion changes the value level but not the optimal LQR feedback. The critic, advantage, admissibility condition, and discounted occupancy used in this experiment are therefore all computed from the same Optimal-PhiBE surrogate, whereas the error is measured with respect to $k^*$.

\begin{figure}
\centering
\begin{subfigure}[t]{0.32\textwidth}
    \centering
    \includegraphics[width=\textwidth]{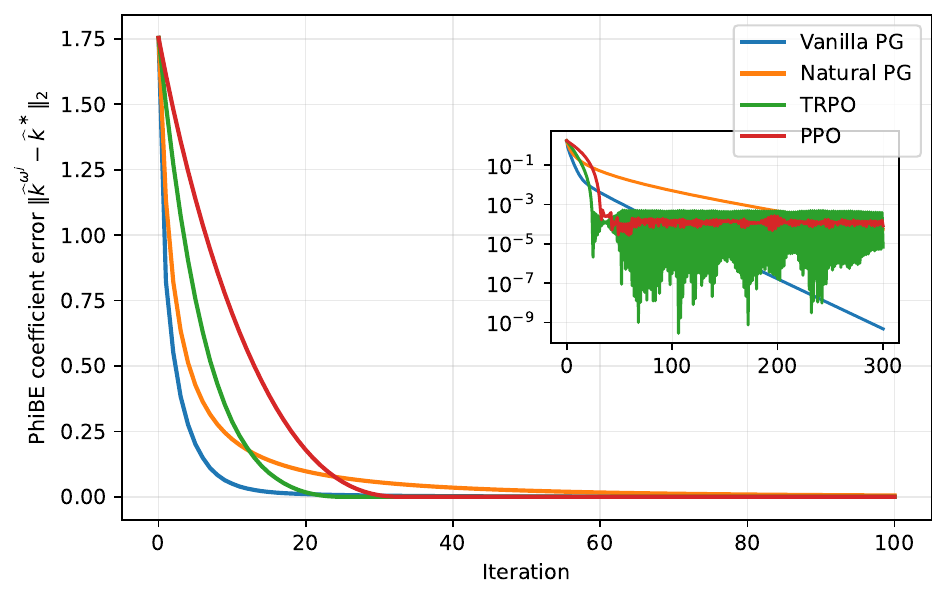}
    \caption{}
    \label{fig:PhiBE_change_A_B}
\end{subfigure}
\begin{subfigure}[t]{0.32\textwidth}
    \centering
    \includegraphics[width=\textwidth]{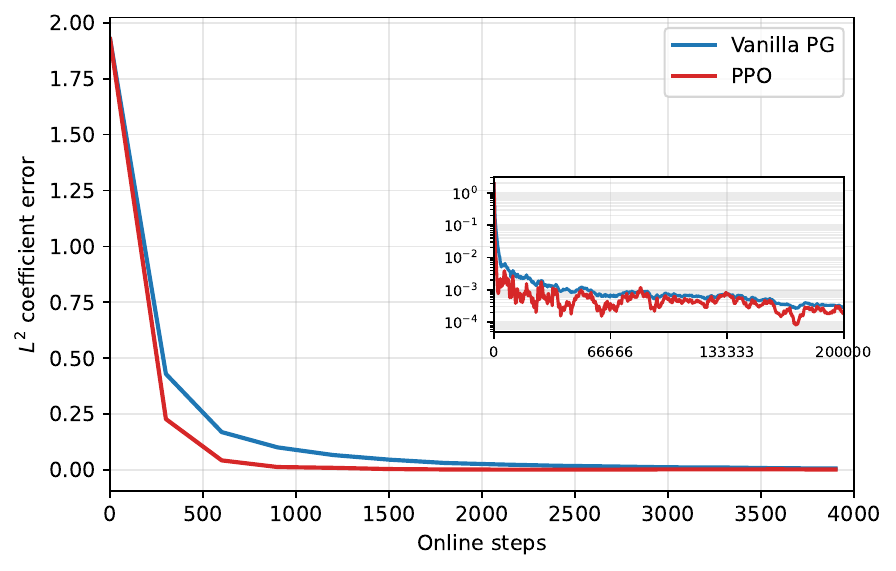}
    \caption{}
    \label{fig:online_data_driven}
\end{subfigure}
\begin{subfigure}[t]{0.32\textwidth}
    \centering
    \includegraphics[width=\textwidth]{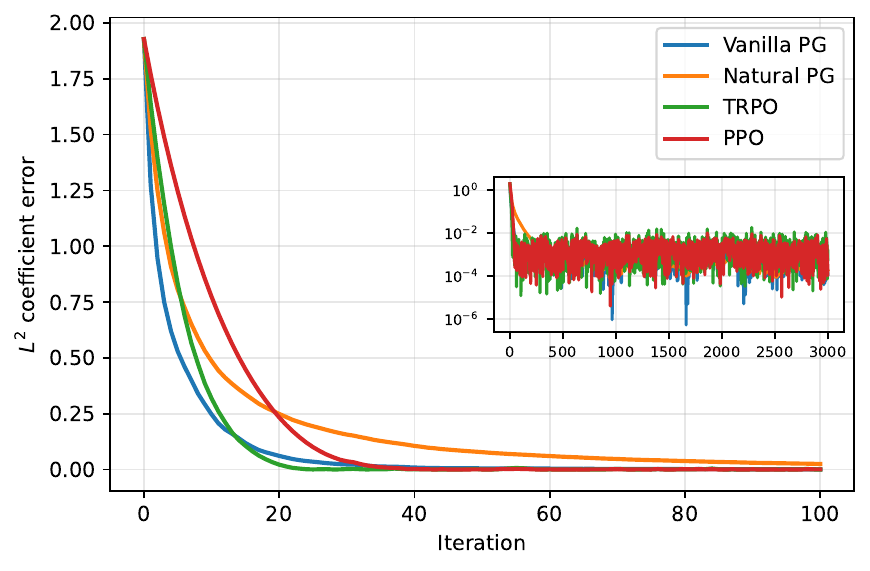}
    \caption{}
    \label{fig:data_driven_change_dynamics}
\end{subfigure}
\caption{Comparison of PhiBE implementations for the LQR example. Left: PhiBE with known discrete-time transition dynamics. Center: online data-driven PhiBE with Vanilla PG and PPO. Right: batch data-driven PhiBE.}
\label{fig:DD}
\end{figure}

Figure~\ref{fig:PhiBE_change_A_B} shows the behavior of the four policy optimization methods under the default configuration. All methods rapidly reduce the error with respect to $k^*$, reproducing the behavior observed with known continuous-time dynamics. Across the additional experiments, the same qualitative conclusions remain valid when the dynamics and update parameters are varied. Changes in $A$ and $B$ affect the transient behavior and the sensitivity to the policy update, while the effective step size continues to determine the main speed--stability trade-off. Smaller steps produce smoother but slower convergence, whereas larger steps reach the low-error region faster but lead to a more visible oscillatory floor. The tested PPO clipping ratios have a comparatively small effect. Overall, at $\Delta t=0.1$, the Optimal-PhiBE surrogate preserves the main optimization behavior observed in the known-dynamics experiments. We refer to \Cref{sec:extra_num_ex} for the corresponding parameter analysis.

\subsubsection{Data-driven: Model-free PhiBE}\label{sec:lqr_on_policy}

In the data driven experiment, every actor iteration uses a fresh batch from the current discounted occupancy measure. We sample actions from $a_i\sim\pi_{\omega^k}(\cdot|S_i)$ and observe one local transition over $\Delta t$. Hence the critic and advantage use fresh current-policy data, and the actor has the state weighting required by \eqref{eq:lqr_mc_gradient}. We use the state basis $\phi(s)=(1,s,s^2)^\top$, $n=1{,}000$ transitions per iteration, and $\Delta t=0.1$. As in the Optimal-PhiBE experiment, the critic and advantage use only the drift estimate $\hat b_i=\Delta S_i/\Delta t$ and set $\hat\Sigma=0$: since the diffusion is additive, it only affects the constant coefficient of the value function, so the advantage and the optimal feedback are unchanged. The default update parameters are $\eta_{\rm PG}=\eta_{\rm NPG}=0.05$, $\delta_{\rm TRPO}=0.005$, and $\eta_{\rm PPO}=0.005$, with five PPO epochs and $\epsilon=0.2$. The implementation used in these experiments is detailed in Algorithm~\ref{alg:on_policy_lqr} in \Cref{sec:extra_num_ex}.

Figure~\ref{fig:DD}\subref{fig:data_driven_change_dynamics} shows the behavior under the default configuration. All four methods rapidly reduce the coefficient error and enter a low-error region. In contrast to the known-dynamics experiments, however, the error does not continue toward numerical precision. Since a new finite-sample critic and advantage are estimated at every actor iteration, the curves fluctuate around a nonzero statistical error level.

The additional experiments show that this behavior is robust to changes in $A$ and $B$, while high diffusion is substantially more challenging because it increases the variance of the drift estimates $\hat b_i=\Delta S_i/\Delta t$. The effective update size controls the usual speed--stability trade-off: smaller steps give smoother but slower transients, whereas larger steps amplify the iteration-to-iteration fluctuations. Changing the PPO clipping ratio has a comparatively small effect in the tested range. Overall, these experiments indicate that the long-run variability is mainly associated with the finite-sample critic and advantage estimation rather than with a failure of policy improvement. We refer to \Cref{sec:extra_num_ex} for the corresponding parameter discussion.

\subsubsection{Data-driven: online model-free PhiBE}\label{sec:lqr_online}

We next study the online actor--critic on the same LQR problem. At each step, we sample a state from the discounted occupancy measure of the current policy, draw an action from that policy, and observe one local transition over $\Delta t=0.1$. The critic uses $\phi(s)=(1,s,s^2)^\top$ and is updated after every transition. Vanilla PG also updates the actor after every transition, whereas PPO holds the actor fixed for 2 transitions and then performs two clipped-update epochs. Each method uses $200{,}000$ local transitions. Further implementation details are given in \Cref{subsec:lqr_setting}.

Figure~\ref{fig:DD}\subref{fig:online_data_driven} shows the default comparison. In the displayed run, PPO reduces the coefficient error faster initially, while both methods reach small errors of the same order. The logarithmic inset shows further gradual improvement with fluctuations over the full observation horizon. An extended parameter analysis is reported in \Cref{subsec:lqr_setting}.

\paragraph{Comparison across regimes.}
The known-transition Optimal-PhiBE surrogate retains the qualitative optimization behavior observed with exact dynamics. The data-driven batch and online implementations also reduce the coefficient error substantially, although their finite-data trajectories fluctuate. The online results do not establish that PPO clipping itself produces the faster transient: PPO also differs from Vanilla PG in its actor-update frequency and effective step size. The additional experiments in \Cref{subsec:lqr_setting} show where the online method remains effective and where changes in the dynamics make estimation substantially more difficult.

\subsection{Differential wheeled robot control}
\label{subsec:example_differential_wheeled}

We next consider a nonlinear control problem in which a differential wheeled robot moves on the plane. In contrast with the LQR benchmark, the drift is nonlinear in the heading and the value function is approximated by a neural network.

\subsubsection{Problem setting and implementation}

The state is $s=(x,y,\theta)^\top$, where $(x,y)$ is the position and $\theta$ is the heading, identified modulo $2\pi$. The control $a\in\R$ is the angular velocity, and $B$ is a three-dimensional standard Brownian motion. With constant forward speed $v=1$ and diffusion coefficient $\sigma_d>0$, the dynamics are
\begin{equation}
    \label{eq:robot_dynamics}
    \rd s_t
    =
    \begin{pmatrix}
        \cos\theta_t&
        \sin\theta_t&
        a_t
    \end{pmatrix}^{\!\top}\rd t
    +\sigma_d\,\rd B_t .
\end{equation}
The robot starts at $(x_0,y_0)=(0,0)$, with $\theta_0\sim\mathcal U[-\pi,\pi)$, and the target is $(x_f,y_f)=(1,1)$. For $\varepsilon_r>0$, define the smoothed distance to the target by
\begin{equation*}
    d_{\varepsilon_r}(x,y)
    :=
    \sqrt{(x-x_f)^2+(y-y_f)^2+\varepsilon_r^2}-\varepsilon_r .
\end{equation*}
The running reward is $r(s,a)
    =
    \alpha\exp\!\left(-\mu d_{\varepsilon_r}(x,y)\right)
    -\gamma a^2 .$ Thus, proximity to the target is rewarded while large angular velocities are penalized. The target is a soft reward maximizer rather than an absorbing state, so a trajectory is not terminated when it enters the displayed target neighborhood. We use the Gaussian policy
\begin{equation}
    \label{eq:robot_policy}
    \pi_\Theta(\rd a\mid s)
    =\mathcal N\!\left(m_\Theta(s),\sigma_a^2\right)(\rd a),
    \qquad
    m_\Theta(s)=3\pi\tanh(f_\Theta(s)),
\end{equation}
where $\sigma_a$ is fixed. The critic $V_\phi$ and the policy mean are represented by neural networks. Both receive $(\widetilde x,\widetilde y,\sin\theta,\cos\theta)$ as input, where the spatial coordinates are scaled to $[-1,1]$ on the collocation domain. This enforces periodicity in the heading and avoids a discontinuity at $\theta=\pm\pi$.

\paragraph{Known dynamics.}
For the Gaussian policy in \eqref{eq:robot_policy}, the policy-averaged drift and regularized reward in the evaluation equation are available in closed form: $ b^{\pi_\Theta}(s)=
    (
        \cos\theta , \sin\theta ,m_\Theta(s)
    )^{\!\top},$ and
\begin{align*}
    r_\lambda^{\pi_\Theta}(s)
    =
    \alpha e^{-\mu d_{\varepsilon_r}(x,y)}
     -\gamma\bigl(m_\Theta(s)^2+\sigma_a^2\bigr)
    +\frac{\lambda}{2}\log\!\left(2\pi e\sigma_a^2\right).
\end{align*}
Accordingly, the critic minimizes the mean squared residual of
\begin{equation}
    \label{eq:robot_residual_known}
    \mathcal R_{\rm KD}(s)
    =
    \beta V_\phi(s)
    -r_\lambda^{\pi_\Theta}(s)
    -b^{\pi_\Theta}(s)^\top\nb_sV_\phi(s)
    -\frac{\sigma_d^2}{2}\Delta V_\phi(s).
\end{equation}
The action average is therefore taken before the residual is squared. Once $V_\phi$ has been fitted, the actor uses the continuous-time advantage \eqref{eq:advantage_representation}, evaluated with the known coefficients.

\paragraph{Data-driven dynamics.}
The learner now observes only one-step tuples $(s_i,a_i,r_i,s_i')$ with interval $\dt$, where $a_i\sim\pi_\Theta(\cdot\mid s_i)$. With $\Delta s_i=s_i'-s_i$ and the angular increment taken modulo $2\pi$, the critic minimizes the mean squared residual of
\begin{equation}
    \label{eq:robot_residual_dd}
    \mathcal R_{\rm DD}(s_i)
    =
    \beta V_\phi(s_i)
    -\bigl(r_i-\lambda\log\pi_\Theta(a_i\mid s_i)\bigr)
    -\frac{\Delta s_i}{\dt}\cdot\nb_sV_\phi(s_i)
    -\frac{1}{2\dt}\Delta s_i^\top\nb_s^2V_\phi(s_i)\Delta s_i ,
\end{equation}
which is the data-driven counterpart of \eqref{eq:robot_residual_known}: the policy-averaged reward and differential operators are replaced by local one-transition estimates constructed from \eqref{eq:approximation_b_and_simga_hat}. Since this residual is squared, its minimization is affected by double sampling; see \Cref{rem:double_sampling}. The actor uses \eqref{eq:adv_phibe}, evaluated with the same one-transition estimates at each sampled pair $(s_i,a_i)$.

\paragraph{Actor sampling and training parameters.}
In both regimes, actor states are sampled from the discounted occupancy measure using the exponential-time construction described in \eqref{eq:lqr_mc_gradient}. Actions and likelihood ratios are generated under a frozen copy of the current policy, and the policy is then updated with the PPO objective \eqref{eq:PPO_direction}. Advantages are standardized within each batch. The critic collocation distribution is used only to solve the evaluation equation; it is not substituted for the discounted occupancy measure in the actor update. We set
\[
    \alpha=5,\quad \mu=1,\quad \gamma=0.05,\quad
    \varepsilon_r=10^{-2},\quad \lambda=0.1,\quad
    \beta=1,\quad \sigma_a=0.1,\quad \sigma_d=0.02
\]
in both regimes. For the data-driven experiment, the observation interval is $\dt=0.01$. The value and policy functions are represented by neural networks and trained with Adam. Complete network architectures, collocation distributions, optimization parameters, batch sizes, training schedules, and validation procedures are reported in \Cref{sec:extra_num_ex}.

\subsubsection{Known-dynamics results}

\begin{figure}
    \centering
    \begin{subfigure}[t]{0.24\textwidth}
        \centering
        \includegraphics[width=\textwidth]{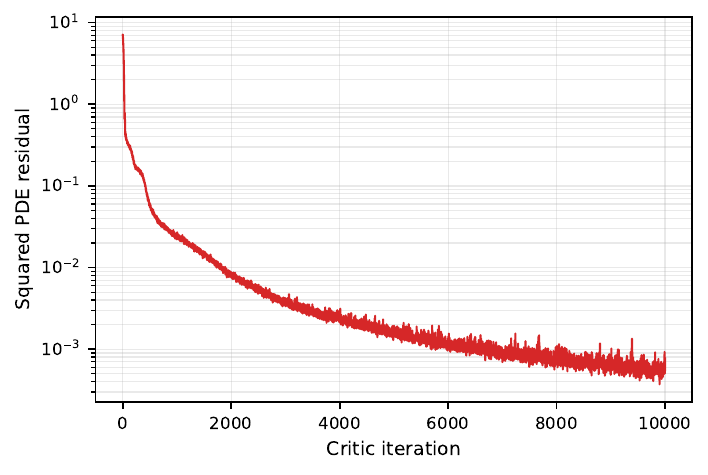}
        \caption{}
        \label{fig:robot_pe_kd_loss}
    \end{subfigure}
    \begin{subfigure}[t]{0.24\textwidth}
        \centering
        \includegraphics[width=\textwidth]{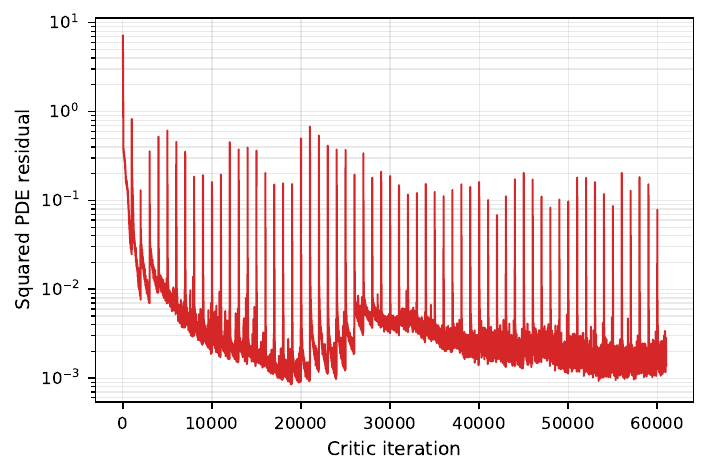}
        \caption{}
        \label{fig:robot_kd_loss}
    \end{subfigure}
    \begin{subfigure}[t]{0.24\textwidth}
        \centering
        \includegraphics[width=\textwidth]{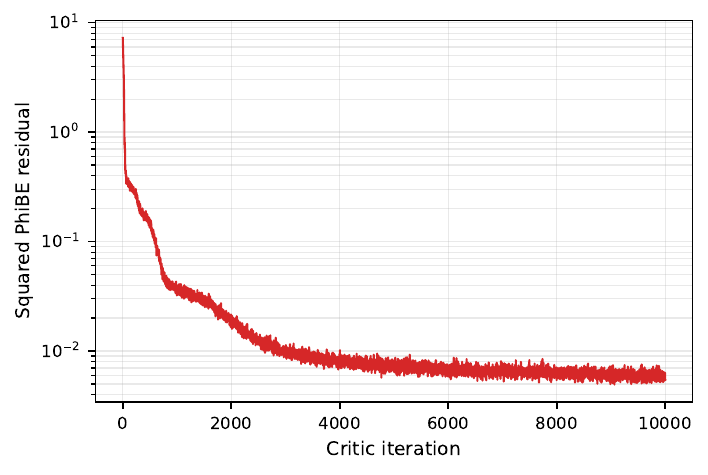}
        \caption{}
        \label{fig:robot_pe_dd_loss}
    \end{subfigure}
    \begin{subfigure}[t]{0.24\textwidth}
        \centering
        \includegraphics[width=\textwidth]{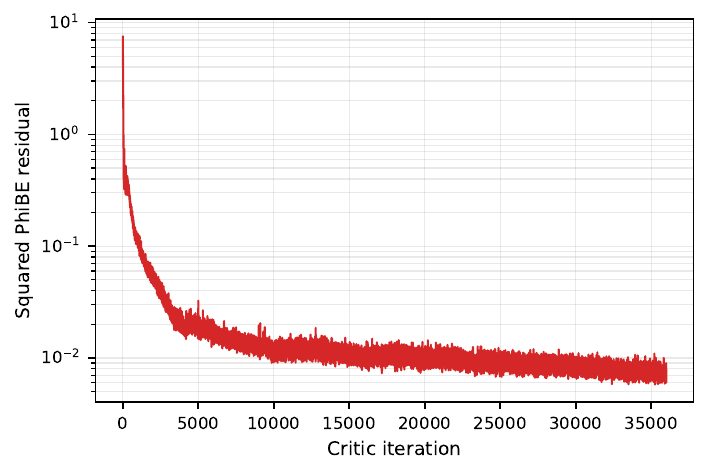}
        \caption{}
        \label{fig:robot_dd_loss}
    \end{subfigure}
    \caption{Critic residuals for the differential wheeled robot. (a) Fixed-policy evaluation with known dynamics; (b) policy optimization with known dynamics; (c) fixed-policy evaluation using data-driven PhiBE; (d) data-driven policy optimization.}\label{fig:robot_critic_residuals}
\end{figure}

\paragraph{Policy evaluation.}
We first fit the critic for a fixed randomly initialized policy. Figure~\ref{fig:robot_critic_residuals}\subref{fig:robot_pe_kd_loss} shows that the squared PDE residual decreases steadily during training and reaches a low level after $10{,}000$ critic iterations. To verify that this reduction corresponds to an accurate value approximation, we also compare the fitted critic with an independent Monte Carlo estimate of the same entropy-regularized discounted return. Across the tested headings, the fitted and reference values have nearly identical high-value regions and contour geometry, showing that the neural critic captures the main spatial and heading-dependent structure of the value function. The corresponding value-function comparisons are reported in \Cref{sec:extra_num_ex}.

\paragraph{Policy optimization.}
Figure~\ref{fig:robot_critic_residuals}\subref{fig:robot_kd_loss} shows the critic residual during joint policy optimization. The curve concatenates successive critic-fitting blocks, and each policy update changes the evaluation equation. Consequently, the residual can increase at the beginning of a new block and then decrease again as the critic adapts to the updated policy. The repeated recovery to a low residual shows that the critic continues to track the evolving policy throughout training.

The final fitted value also closely reproduces the Monte Carlo reference for the learned policy. At the same time, the mean-policy trajectories evolve from paths largely determined by their initial headings to heading-dependent curves directed toward the target neighborhood. Since the target is a soft reward maximizer, the forward speed is fixed, and no terminal condition is imposed, trajectories may continue curving after reaching this region. Together, these experiments show that the critic remains accurate during policy improvement and that the learned feedback policy successfully redirects different initial headings toward the high-reward region. The value-function and trajectory comparisons are reported in \Cref{sec:extra_num_ex}.

\subsubsection{Data-driven results}

The data-driven critic and actor use only rewards and observed increments in their PhiBE quantities. The analytical drift and diffusion in \eqref{eq:robot_dynamics} are used only by the simulator that supplies the transitions and by the post-training Monte Carlo validation.

\paragraph{Policy evaluation.}
We first evaluate a fixed policy using only one-step transition data. Figure~\ref{fig:robot_critic_residuals}\subref{fig:robot_pe_dd_loss} shows that the squared PhiBE residual decreases by several orders of magnitude before fluctuating around a small nonzero level. In contrast with the known-dynamics PDE residual, this statistical floor is expected because the reward and local PhiBE quantities are estimated from finitely many transitions and the squared one-sample residual is affected by the double-sampling effect described in \Cref{rem:double_sampling}.

Independent Monte Carlo validation shows that the fitted critic nevertheless recovers the main structure of the value function. Across the tested headings, the fitted values reproduce the location of the high-value region and its systematic change with the robot orientation, although some differences in contour geometry and value level remain. Thus, the nonzero residual floor does not prevent the data-driven critic from recovering a meaningful approximation of the policy value. The corresponding fitted--reference comparisons are reported in \Cref{sec:extra_num_ex}.

\begin{remark}[Double sampling]\label{rem:double_sampling}
Let $R_i(V)$ denote the one-sample residual \eqref{eq:robot_residual_dd}, and let $\mathcal R(V):=\beta V-r_\lambda^{\pi_\Theta}-\mathcal L_{\hat b^{\pi_\Theta},\hat\Sigma^{\pi_\Theta}}V$ be the residual of \eqref{eq:phibe}. Since $\E[R_i(V)\,|\,s_i]=\mathcal R(V)(s_i)$,
\[
    \E\bigl[R_i(V)^2\,\big|\,s_i\bigr]
    =
    \mathcal R(V)(s_i)^2
    +
    \operatorname{Var}\bigl(R_i(V)\,\big|\,s_i\bigr).
\]
The conditional variance depends on $V$ through $\nb_sV$ and $\nb_s^2V$, so the minimizer of the mean squared one-sample residual differs in general from that of the mean squared PhiBE residual; an unbiased estimate of the latter would require two independent transitions from the same state. This is the double-sampling problem. The Galerkin critics of \Cref{subsec:LQR_example} are not affected: the residual enters linearly and is multiplied by test functions of $s_i$, or of $(s_i,a_i)$, so the empirical Galerkin system is a consistent estimate of the exact one. In the present example, the part of the conditional variance that depends on $V$ is generated by the Brownian increment, of order $\sigma_d^2\dt^{-1}|\nb_sV|^2$, and by the action noise, of order $\sigma_a^2$. Both are small for $\sigma_d=0.02$, $\dt=0.01$ and $\sigma_a=0.1$, although the former increases as $\dt$ decreases. Since this variance term does not depend on the number of samples, it also keeps the empirical loss away from zero.
\end{remark}

\paragraph{Policy optimization.}
Figure~\ref{fig:robot_critic_residuals}\subref{fig:robot_dd_loss} shows the critic residual during data-driven policy optimization. Despite the critic being reconstructed from sampled transitions as the policy changes, the residual decreases substantially and eventually fluctuates around a small positive level. This behavior is consistent with the statistical floor already observed in fixed-policy evaluation.

The additional experiments show that the learned value retains the main geometry of the independent Monte Carlo reference and, more importantly, that policy improvement translates into effective feedback control. Starting from trajectories largely determined by their initial headings, the learned policy progressively redirects all tested headings toward the target neighborhood. This behavior is already visible well before the final training epoch and remains stable at the end of training. Thus, the data-driven PhiBE critic provides sufficiently accurate policy-improvement directions to learn the nonlinear steering task using only observed transitions. The corresponding value-function and trajectory comparisons are reported in \Cref{sec:extra_num_ex}.

\section{Conclusion}\label{sec:conclusion}

We developed a policy gradient framework for continuous-time reinforcement learning with theoretical convergence guarantees and data-driven implementations. Starting from a performance difference lemma, we derived policy gradient formulas for randomized and parametrized policies, established exponential convergence of the entropy-regularized policy flow, and obtained convergence bounds for discrete updates that account for critic and policy-regression errors. By incorporating PhiBE, we constructed actor--critic methods that learn from discrete observations while preserving the continuous-time dynamics. Numerical experiments on an LQR problem and a nonlinear wheeled robot demonstrated the effectiveness of the proposed approach.
\section*{Acknowledgments}
M. Zhou acknowledges support from a research grant from Peking University. Y. Zhu and M. Hernandez are supported in part by NSF grant No. 2529107.  

\bibliographystyle{plain}
\bibliography{ref}

@article{wang2020reinforcement,
  author = {Wang, Haoran and Zariphopoulou, Thaleia and Zhou, Xun Yu},
  title = {{Reinforcement learning in continuous time and space: A stochastic control approach}},
  journal = {Journal of Machine Learning Research},
  volume = {21},
  number = {198},
  pages = {1--34},
  year = {2020},
  url = {https://www.jmlr.org/papers/v21/19-144.html},
}

@article{jia2023q,
  author = {Jia, Yanwei and Zhou, Xun Yu},
  title = {{q-Learning in continuous time}},
  journal = {Journal of Machine Learning Research},
  volume = {24},
  number = {161},
  pages = {1--61},
  year = {2023},
  url = {https://www.jmlr.org/papers/v24/22-0755.html},
}

@misc{zhu2025optimal,
  author = {Zhu, Yuhua and Zhang, Yuming and Zhang, Haoyu},
  title = {{Optimal-PhiBE: A PDE-based Model-free framework for Continuous-time Reinforcement Learning}},
  year = {2025},
  eprint = {2506.05208},
  archivePrefix = {arXiv},
  url = {https://arxiv.org/abs/2506.05208},
}

@article{sethi2025entropy,
  author = {Sethi, Deven and {\v{S}}i{\v{s}}ka, David and Zhang, Yufei},
  title = {{Entropy annealing for policy mirror descent in continuous time and space}},
  journal = {SIAM Journal on Control and Optimization},
  volume = {63},
  number = {4},
  pages = {3006--3041},
  year = {2025},
  publisher = {SIAM},
  doi = {10.1137/24m166591x},
}

@article{zhou2025policy,
  author = {Zhou, Mo and Lu, Jianfeng},
  title = {{A policy gradient framework for stochastic optimal control problems with global convergence guarantee}},
  journal = {SIAM Journal on Control and Optimization},
  volume = {63},
  number = {4},
  pages = {2605--2631},
  year = {2025},
  publisher = {SIAM},
  doi = {10.1137/23m1570739},
}

@article{agarwal2021theory,
  author = {Agarwal, Alekh and Kakade, Sham M and Lee, Jason D and Mahajan, Gaurav},
  title = {{On the theory of policy gradient methods: Optimality, approximation, and distribution shift}},
  journal = {Journal of Machine Learning Research},
  volume = {22},
  number = {98},
  pages = {1--76},
  year = {2021},
  url = {https://www.jmlr.org/papers/v22/19-736.html},
}

@article{jia2022policy,
  author = {Jia, Yanwei and Zhou, Xun Yu},
  title = {{Policy gradient and actor-critic learning in continuous time and space: Theory and algorithms}},
  journal = {Journal of Machine Learning Research},
  volume = {23},
  number = {275},
  pages = {1--50},
  year = {2022},
  url = {https://www.jmlr.org/papers/v23/21-1387.html},
}

@inproceedings{zhao2023policy,
  author = {Zhao, Hanyang and Tang, Wenpin and Yao, David},
  title = {{Policy optimization for continuous reinforcement learning}},
  booktitle = {Advances in Neural Information Processing Systems},
  volume = {36},
  pages = {13637--13663},
  year = {2023},
  doi = {10.52202/075280-0601},
  url = {https://proceedings.neurips.cc/paper_files/paper/2023/hash/2c53bc01e30711a08f6ac86919193022-Abstract-Conference.html},
}

@article{tang2022exploratory,
  author = {Tang, Wenpin and Zhang, Yuming Paul and Zhou, Xun Yu},
  title = {{Exploratory HJB equations and their convergence}},
  journal = {SIAM Journal on Control and Optimization},
  volume = {60},
  number = {6},
  pages = {3191--3216},
  year = {2022},
  publisher = {SIAM},
  doi = {10.1137/21m1448185},
}

@article{zhao2022data,
  author = {Zhao, Jingang},
  title = {{Data-driven adaptive dynamic programming for optimal control of continuous-time multicontroller systems with unknown dynamics}},
  journal = {IEEE Access},
  volume = {10},
  pages = {41503--41511},
  year = {2022},
  publisher = {IEEE},
  doi = {10.1109/access.2022.3168032},
}

@article{yang2014reinforcement,
  author = {Yang, Xiong and Liu, Derong and Wang, Ding},
  title = {{Reinforcement learning for adaptive optimal control of unknown continuous-time nonlinear systems with input constraints}},
  journal = {International Journal of Control},
  volume = {87},
  number = {3},
  pages = {553--566},
  year = {2014},
  publisher = {Taylor \& Francis},
  doi = {10.1080/00207179.2013.848292},
}

@article{jiang2012computational,
  author = {Jiang, Yu and Jiang, Zhong-Ping},
  title = {{Computational adaptive optimal control for continuous-time linear systems with completely unknown dynamics}},
  journal = {Automatica},
  volume = {48},
  number = {10},
  pages = {2699--2704},
  year = {2012},
  publisher = {Elsevier},
  doi = {10.1016/j.automatica.2012.06.096},
}

@inproceedings{kim2020hamilton,
  author = {Kim, Jeongho and Yang, Insoon},
  title = {{Hamilton-Jacobi-Bellman equations for Q-learning in continuous time}},
  booktitle = {Proceedings of the 2nd Conference on Learning for Dynamics and Control},
  series = {Proceedings of Machine Learning Research},
  volume = {120},
  pages = {739--748},
  year = {2020},
  publisher = {PMLR},
  url = {https://proceedings.mlr.press/v120/kim20b.html},
}

@article{song2023reaching,
  author = {Song, Yunlong and Romero, Angel and M{\"u}ller, Matthias and Koltun, Vladlen and Scaramuzza, Davide},
  title = {{Reaching the limit in autonomous racing: Optimal control versus reinforcement learning}},
  journal = {Science Robotics},
  volume = {8},
  number = {82},
  pages = {eadg1462},
  year = {2023},
  publisher = {American Association for the Advancement of Science},
  doi = {10.1126/scirobotics.adg1462},
}

@book{pham2009continuous,
  author = {Pham, Huy{\^e}n},
  title = {{Continuous-time stochastic control and optimization with financial applications}},
  series = {Stochastic Modelling and Applied Probability},
  volume = {61},
  year = {2009},
  publisher = {Springer},
  doi = {10.1007/978-3-540-89500-8},
}

@article{sethi2002optimal,
  author = {Sethi, Suresh P and Yan, Houmin and Zhang, Hanqin and Zhang, Qing},
  title = {{Optimal and hierarchical controls in dynamic stochastic manufacturing systems: A survey}},
  journal = {Manufacturing \& Service Operations Management},
  volume = {4},
  number = {2},
  pages = {133--170},
  year = {2002},
  publisher = {INFORMS},
  doi = {10.1287/msom.4.2.133.281},
}

@article{lin1998optimal,
  author = {Lin, Feng and Brandt, Robert D},
  title = {{An optimal control approach to robust control of robot manipulators}},
  journal = {IEEE Transactions on Robotics and Automation},
  volume = {14},
  number = {1},
  pages = {69--77},
  year = {1998},
  publisher = {IEEE},
  doi = {10.1109/70.660845},
}

@article{KERIMKULOV2025104765,
  author = {Kerimkulov, Bekzhan and {\v{S}}i{\v{s}}ka, David and Szpruch, {\L}ukasz and Zhang, Yufei},
  title = {{Mirror descent for stochastic control problems with measure-valued controls}},
  journal = {Stochastic Processes and their Applications},
  volume = {190},
  pages = {104765},
  year = {2025},
  issn = {0304-4149},
  doi = {10.1016/j.spa.2025.104765},
}

@article{giegrich2024convergence,
  author = {Giegrich, Michael and Reisinger, Christoph and Zhang, Yufei},
  title = {{Convergence of policy gradient methods for finite-horizon exploratory linear-quadratic control problems}},
  journal = {SIAM Journal on Control and Optimization},
  volume = {62},
  number = {2},
  pages = {1060--1092},
  year = {2024},
  publisher = {SIAM},
  doi = {10.1137/22m1533517},
}

@article{reisinger2021regularity,
  author = {Reisinger, Christoph and Zhang, Yufei},
  title = {{Regularity and stability of feedback relaxed controls}},
  journal = {SIAM Journal on Control and Optimization},
  volume = {59},
  number = {5},
  pages = {3118--3151},
  year = {2021},
  publisher = {SIAM},
  doi = {10.1137/20m1312435},
}

@misc{zhou2025learning,
  author = {Zhou, Mo and Zhou, Haosheng and Hu, Ruimeng},
  title = {{Learning Mean-Field Games through Mean-Field Actor-Critic Flow}},
  year = {2025},
  eprint = {2510.12180},
  archivePrefix = {arXiv},
  url = {https://arxiv.org/abs/2510.12180},
}

@inproceedings{schulman2015trpo,
  author = {Schulman, John and Levine, Sergey and Moritz, Philipp and Jordan, Michael I. and Abbeel, Pieter},
  title = {{Trust Region Policy Optimization}},
  booktitle = {Proceedings of the 32nd International Conference on Machine Learning},
  series = {Proceedings of Machine Learning Research},
  volume = {37},
  pages = {1889--1897},
  year = {2015},
  publisher = {PMLR},
  url = {https://proceedings.mlr.press/v37/schulman15.html},
}

@misc{schulman2017ppo,
  author = {Schulman, John and Wolski, Filip and Dhariwal, Prafulla and Radford, Alec and Klimov, Oleg},
  title = {{Proximal Policy Optimization Algorithms}},
  year = {2017},
  eprint = {1707.06347},
  archivePrefix = {arXiv},
  url = {https://arxiv.org/abs/1707.06347},
}

@article{zhuphibe_evaluation,
  author = {Zhu, Yuhua},
  title = {{PhiBE: A PDE-Based Bellman Equation for Continuous-Time Policy Evaluation}},
  journal = {Journal of Machine Learning Research},
  volume = {27},
  number = {211},
  pages = {1--68},
  year = {2026},
  url = {https://www.jmlr.org/papers/v27/24-0734.html},
}

@book{oksendal2003sde,
  author = {{\O}ksendal, Bernt},
  title = {{Stochastic Differential Equations: An Introduction with Applications}},
  series = {Universitext},
  year = {2003},
  edition = {6},
  publisher = {Springer},
  address = {Berlin},
  doi = {10.1007/978-3-642-14394-6},
}

@book{flemingSoner2006,
  author = {Fleming, Wendell H. and Soner, Halil Mete},
  title = {{Controlled Markov Processes and Viscosity Solutions}},
  series = {Stochastic Modelling and Applied Probability},
  volume = {25},
  year = {2006},
  edition = {2},
  publisher = {Springer},
  address = {New York},
  doi = {10.1007/0-387-31071-1},
}

@article{jia2022policyeval,
  author = {Jia, Yanwei and Zhou, Xun Yu},
  title = {{Policy evaluation and temporal-difference learning in continuous time and space: A martingale approach}},
  journal = {Journal of Machine Learning Research},
  volume = {23},
  number = {154},
  pages = {1--55},
  year = {2022},
  url = {https://www.jmlr.org/papers/v23/21-0947.html},
}

@inproceedings{kakade2001natural,
  author = {Kakade, Sham M.},
  title = {{A natural policy gradient}},
  booktitle = {Advances in Neural Information Processing Systems},
  volume = {14},
  pages = {1531--1538},
  year = {2001},
  url = {https://proceedings.neurips.cc/paper/2001/hash/4b86abe48d358ecf194c56c69108433e-Abstract.html},
}

@inproceedings{sutton1999policygradient,
  author = {Sutton, Richard S. and McAllester, David and Singh, Satinder and Mansour, Yishay},
  title = {{Policy gradient methods for reinforcement learning with function approximation}},
  booktitle = {Advances in Neural Information Processing Systems},
  volume = {12},
  pages = {1057--1063},
  year = {1999},
  url = {https://proceedings.neurips.cc/paper_files/paper/1999/hash/464d828b85b0bed98e80ade0a5c43b0f-Abstract.html},
}

@book{kunita1990flows,
  author = {Kunita, Hiroshi},
  title = {{Stochastic flows and stochastic differential equations}},
  series = {Cambridge Studies in Advanced Mathematics},
  volume = {24},
  year = {1990},
  publisher = {Cambridge University Press, Cambridge},
  isbn = {0-521-35050-6},
  url = {https://assets.cambridge.org/97805213/50501/sample/9780521350501ws.pdf},
}

@book{mao2008sde,
  author = {Mao, Xuerong},
  title = {{Stochastic differential equations and applications}},
  year = {2008},
  edition = {Second},
  publisher = {Horwood Publishing Limited, Chichester},
  isbn = {978-1-904275-34-3},
  doi = {10.1533/9780857099402},
}

@book{krylov1980controlled,
  author = {Krylov, N. V.},
  title = {{Controlled diffusion processes}},
  series = {Applications of Mathematics},
  volume = {14},
  year = {1980},
  publisher = {Springer-Verlag},
  isbn = {0-387-90461-1},
  doi = {10.1007/978-3-540-70914-5},
  note = {Translated from the Russian by A. B. Aries},
}

@article{carmona2025reconcilingdiscretetimemixedpolicies,
  author = {Carmona, Ren{'e} and Lauri{\`e}re, Mathieu},
  title = {{Reconciling Discrete-Time Mixed Policies and Continuous-Time Relaxed Controls in Reinforcement Learning and Stochastic Control}},
  journal = {SIAM Review},
  volume = {68},
  number = {4},
  pages = {823--849},
  year = {2026},
  doi = {10.1137/25M1782856},
}

@article{reisinger2023linear,
  author = {Reisinger, Christoph and Stockinger, Wolfgang and Zhang, Yufei},
  title = {{Linear Convergence of a Policy Gradient Method for Some Finite Horizon Continuous Time Control Problems}},
  journal = {SIAM Journal on Control and Optimization},
  volume = {61},
  number = {6},
  pages = {3526--3558},
  year = {2023},
  doi = {10.1137/22M1492180},
}

@article{siska2024gradient,
  author = {{\v{S}}i{\v{s}}ka, David and Szpruch, {\L}ukasz},
  title = {{Gradient Flows for Regularized Stochastic Control Problems}},
  journal = {SIAM Journal on Control and Optimization},
  volume = {62},
  number = {4},
  pages = {2036--2070},
  year = {2024},
  doi = {10.1137/20M1373645},
}

@misc{zhou2024solving,
  author = {Zhou, Mo and Lu, Jianfeng},
  title = {{Solving time-continuous stochastic optimal control problems: Algorithm design and convergence analysis of actor-critic flow}},
  year = {2024},
  eprint = {2402.17208},
  archivePrefix = {arXiv},
  url = {https://arxiv.org/abs/2402.17208},
}

@article{jia2026accuracy,
  author = {Jia, Yanwei and Ouyang, Du and Zhang, Yufei},
  title = {{Accuracy of Discretely Sampled Stochastic Policies in Continuous-Time Reinforcement Learning}},
  journal = {SIAM Journal on Control and Optimization},
  volume = {64},
  number = {3},
  pages = {1889--1929},
  year = {2026},
  doi = {10.1137/25M1742199},
}

@article{zhang2025infinite,
  author = {Zhang, Xinpei and Jia, Guangyan},
  title = {{Convergence of Policy Gradient for Stochastic Linear Quadratic Optimal Control Problems in Infinite Horizon}},
  journal = {Journal of Mathematical Analysis and Applications},
  volume = {547},
  number = {1},
  pages = {129264},
  year = {2025},
  doi = {10.1016/j.jmaa.2025.129264},
}

@inproceedings{lillicrap2016continuous,
  author = {Lillicrap, Timothy P. and Hunt, Jonathan J. and Pritzel, Alexander and Heess, Nicolas and Erez, Tom and Tassa, Yuval and Silver, David and Wierstra, Daan},
  title = {{Continuous Control with Deep Reinforcement Learning}},
  booktitle = {International Conference on Learning Representations},
  year = {2016},
  url = {https://arxiv.org/abs/1509.02971},
}

@inproceedings{vieillard2022implicit,
  author = {Vieillard, Nino and Andrychowicz, Marcin and Raichuk, Anton and Pietquin, Olivier and Geist, Matthieu},
  title = {{Implicitly Regularized RL with Implicit Q-values}},
  booktitle = {Proceedings of the 25th International Conference on Artificial Intelligence and Statistics},
  series = {Proceedings of Machine Learning Research},
  volume = {151},
  pages = {1380--1402},
  year = {2022},
  publisher = {PMLR},
  url = {https://proceedings.mlr.press/v151/vieillard22a.html},
}

@inproceedings{kakade2002approximately,
  author = {Kakade, Sham M. and Langford, John},
  title = {{Approximately Optimal Approximate Reinforcement Learning}},
  booktitle = {Proceedings of the Nineteenth International Conference on Machine Learning},
  pages = {267--274},
  year = {2002},
  publisher = {Morgan Kaufmann},
  url = {https://mlanthology.org/icml/2002/kakade2002icml-approximately/},
}

@article{wang2026reinforcement,
  author = {Wang, Boyu and Gao, Xuefeng and Li, Lingfei},
  title = {{Reinforcement learning for continuous-time optimal execution: actor--critic algorithm and error analysis}},
  journal = {Finance and Stochastics},
  volume = {30},
  number = {2},
  pages = {597--655},
  year = {2026},
  publisher = {Springer},
  doi = {10.1007/s00780-026-00589-5},
}

@article{basei2022logarithmic,
  author = {Basei, Matteo and Guo, Xin and Hu, Anran and Zhang, Yufei},
  title = {{Logarithmic regret for episodic continuous-time linear-quadratic reinforcement learning over a finite-time horizon}},
  journal = {Journal of Machine Learning Research},
  volume = {23},
  number = {178},
  pages = {1--34},
  year = {2022},
  url = {https://www.jmlr.org/papers/v23/20-664.html},
}

@article{denkert2025control,
  author = {Denkert, Robert and Pham, Huy{\^e}n and Warin, Xavier},
  title = {{Control randomisation approach for policy gradient and application to reinforcement learning in optimal switching}},
  journal = {Applied Mathematics \& Optimization},
  volume = {91},
  number = {1},
  pages = {9},
  year = {2025},
  publisher = {Springer},
  doi = {10.1007/s00245-024-10207-5},
}

@article{zhou2021actor,
  author = {Zhou, Mo and Han, Jiequn and Lu, Jianfeng},
  title = {{Actor-critic method for high dimensional static Hamilton--Jacobi--Bellman partial differential equations based on neural networks}},
  journal = {SIAM Journal on Scientific Computing},
  volume = {43},
  number = {6},
  pages = {A4043--A4066},
  year = {2021},
  publisher = {SIAM},
  doi = {10.1137/21m1402303},
}

@article{bender2025continuous,
  author = {Bender, Christian and Thuan, Nguyen Tran},
  title = {{Continuous time reinforcement learning: A random measure approach}},
  journal = {Stochastic Processes and their Applications},
  volume = {194},
  pages = {104848},
  year = {2026},
  publisher = {Elsevier},
  doi = {10.1016/j.spa.2025.104848},
}

@inproceedings{Baird1994ReinforcementLI,
  author = {Baird, Leemon C.},
  title = {{Reinforcement learning in continuous time: advantage updating}},
  booktitle = {Proceedings of 1994 IEEE International Conference on Neural Networks (ICNN'94)},
  volume = {4},
  pages = {2448--2453},
  year = {1994},
  doi = {10.1109/ICNN.1994.374604},
}

@inproceedings{tallec2019,
  author = {Tallec, Corentin and Blier, L{\'e}onard and Ollivier, Yann},
  title = {{Making Deep {Q}-Learning Methods Robust to Time Discretization}},
  booktitle = {Proceedings of the 36th International Conference on Machine Learning (ICML)},
  series = {Proceedings of Machine Learning Research},
  volume = {97},
  pages = {6096--6104},
  year = {2019},
  publisher = {PMLR},
  url = {https://proceedings.mlr.press/v97/tallec19a.html},
}

@article{konda2003actorcritic,
  author = {Konda, Vijay R. and Tsitsiklis, John N.},
  title = {{On Actor-Critic Algorithms}},
  journal = {SIAM Journal on Control and Optimization},
  volume = {42},
  number = {4},
  pages = {1143--1166},
  year = {2003},
  doi = {10.1137/S0363012901385691},
}

@article{zeng2024twotimescale,
  author = {Zeng, Sihan and Doan, Thinh T. and Romberg, Justin},
  title = {{A Two-Time-Scale Stochastic Optimization Framework with Applications in Control and Reinforcement Learning}},
  journal = {SIAM Journal on Optimization},
  volume = {34},
  number = {1},
  pages = {946--976},
  year = {2024},
  doi = {10.1137/22M150277X},
}

\appendix
\section{Appendix}

\subsection{Remarks and assumptions equivalence}

We begin this section by writing \Cref{ass:wellposed_policy_evaluation} and \Cref{ass:assumption_pi_theta} as explicit inequalities, in the form in which they are used in the proofs, and by recording two consequences that are invoked repeatedly. We then discuss the well-posedness of the exploratory SDE \eqref{eq:average_SDE} and of the linear equation used for policy evaluation.

\begin{proposition}[Regularity and growth of the value function]
\label{prop:value_function_regularity}
Assume \Cref{ass:wellposed_policy_evaluation} and let $\pi\in\Pi$. Then the value function \eqref{eq:value_function} belongs to $C^2(\S)$, solves \eqref{eq:Bellman_discounted}, and satisfies
\begin{equation}\label{eq:value_polynomial_growth}
    |V^\pi(s)|\leq C_\pi(1+|s|^2),
    \qquad
    |\ns V^\pi(s)|\leq C_\pi(1+|s|),
    \qquad
    \|\ns^2V^\pi(s)\|\leq C_\pi(1+|s|).
\end{equation}
Moreover, $V^\pi$ is the only function of class $C^2$ satisfying \eqref{eq:value_polynomial_growth} that solves \eqref{eq:Bellman_discounted}.
\end{proposition}
 
\begin{proof}
Throughout, $X_t=X_t^s$ denotes the solution of \eqref{eq:average_SDE} with $X_0=s$, and $C$ is a constant that may change from line to line.
 
\emph{Step 1: growth of $V^\pi$.} By \eqref{eq:closed_loop_reward_regularity} and Taylor's theorem, $|r_\lambda^\pi(s)|\leq|r_\lambda^\pi(0)|+C(1+|s|^2)\leq C(1+|s|^2)$. Together with \eqref{eq:state_moment_growth} and Jensen's inequality this gives
\[
    |V^\pi(s)|
    \leq
    \int_0^\infty e^{-\beta t}\E\left[|r_\lambda^\pi(X_t)|\right]\rd t
    \leq
    C(1+|s|^2)\int_0^\infty e^{-(\beta-\kappa_\pi/4)t}\rd t
    \leq
    C(1+|s|^2),
\]
the integral being finite by \eqref{eq:discount_condition}, which gives $\kappa_\pi\leq2\nu_\pi+\kappa_\pi<\beta$.
 
\emph{Step 2: derivatives of the flow.} By \Cref{lem:closed_loop_rates}, writing $J_t:=\ns X_t^s$ and $H_t:=\ns^2X_t^s$, we have
\begin{equation}\label{eq:flow_moments_used}
    \left(\E\|J_t\|^2\right)^{1/2}
    \leq
    \left(\E\|J_t\|^4\right)^{1/4}
    \leq
    C_\pi e^{\nu_\pi t},
    \qquad
    \left(\E\|H_t\|^2\right)^{1/2}
    \leq
    C_{\pi,\eps}e^{(2\nu_\pi+\eps)t},
\end{equation}
for every $\eps>0$, both uniformly in $s$.

\emph{Step 3: derivatives of $V^\pi$.} By \eqref{eq:closed_loop_reward_regularity}, $r_\lambda^\pi$ is of class $C^2$ with $|\ns r_\lambda^\pi(x)|\leq C(1+|x|)$ and $\|\ns^2r_\lambda^\pi(x)\|\leq C$. Combining this with \eqref{eq:state_moment_growth} and \eqref{eq:flow_moments_used}, the difference quotients of $r_\lambda^\pi(X_t^s)$ in $s$ are uniformly integrable on compact sets of $s$, uniformly for $t$ in compact sets, and the resulting integrands are dominated by an $e^{-\beta t}$-integrable function by \eqref{eq:discount_condition}. Differentiating \eqref{eq:value_function} twice under the expectation and the time integral is therefore legitimate and gives
\[
\begin{aligned}
    \ns V^\pi(s)
    =
    \E\int_0^\infty e^{-\beta t}J_t^\top\ns r_\lambda^\pi(X_t)\,\rd t,\quad
    \ns^2V^\pi(s)
    =
    \E\int_0^\infty e^{-\beta t}
    \left[
        J_t^\top\ns^2r_\lambda^\pi(X_t)J_t
        +
        \ns r_\lambda^\pi(X_t)^\top H_t
    \right]\rd t .
\end{aligned}
\]
In particular $V^\pi\in C^2(\S)$. By \eqref{eq:state_moment_growth} and Jensen's inequality, $\bigl(\E(1+|X_t|)^2\bigr)^{1/2}\leq C(1+|s|)e^{\kappa_\pi t/8}$. Hence, by the Cauchy--Schwarz inequality and \eqref{eq:flow_moments_used},
\[
    |\ns V^\pi(s)|
    \leq
    C\int_0^\infty e^{-\beta t}
    \left(\E|J_t|^2\right)^{1/2}
    \left(\E(1+|X_t|)^2\right)^{1/2}\rd t
    \leq
    C(1+|s|)\int_0^\infty e^{-(\beta-\nu_\pi-\kappa_\pi/8)t}\rd t
    \leq
    C(1+|s|),
\]
and, using \eqref{eq:flow_moments_used} for the second term,
\[
    \|\ns^2V^\pi(s)\|
    \leq
    C\int_0^\infty e^{-\beta t}
    \left[
        \E|J_t|^2
        +
        \left(\E(1+|X_t|)^2\right)^{1/2}\left(\E|H_t|^2\right)^{1/2}
    \right]\rd t
    \leq
    C(1+|s|)\int_0^\infty e^{-(\beta-2\nu_\pi-\kappa_\pi/8-\eps)t}\rd t,
\]
which is finite for $\eps$ small enough by \eqref{eq:discount_condition}. This proves \eqref{eq:value_polynomial_growth}.
 
\emph{Step 4: $V^\pi$ solves \eqref{eq:Bellman_discounted}.} Fix $T>0$. By the Markov property,
\[
    V^\pi(s)
    =
    \E\left[\int_0^Te^{-\beta t}r_\lambda^\pi(X_t)\,\rd t\right]
    +
    e^{-\beta T}\E\left[V^\pi(X_T)\right].
\]
Since $V^\pi\in C^2$ satisfies \eqref{eq:value_polynomial_growth} and $\|\sigma^\pi(x)\|\leq C(1+|x|)$, the stochastic integral appearing in It\^o's formula applied to $e^{-\beta t}V^\pi(X_t)$ is a true martingale on $[0,T]$ by \eqref{eq:state_moment_growth}, so
\[
    e^{-\beta T}\E\left[V^\pi(X_T)\right]-V^\pi(s)
    =
    \E\int_0^Te^{-\beta t}
    \left(\L_{b^\pi,\Sigma^\pi}V^\pi-\beta V^\pi\right)(X_t)\,\rd t .
\]
Adding the two identities gives
\[
    \E\int_0^Te^{-\beta t}
    \left(\L_{b^\pi,\Sigma^\pi}V^\pi-\beta V^\pi+r_\lambda^\pi\right)(X_t)\,\rd t=0,
    \qquad T>0 .
\]
The integrand is continuous in $x$ and dominated by $C(1+|x|^3)$, which is uniformly integrable along $X_t$ for small $t$ by \eqref{eq:state_moment_growth}. Dividing by $T$ and letting $T\to0^+$ yields $\L_{b^\pi,\Sigma^\pi}V^\pi(s)-\beta V^\pi(s)+r_\lambda^\pi(s)=0$.
 
\emph{Step 5: uniqueness.} Let $u\in C^2(\S)$ satisfy \eqref{eq:value_polynomial_growth} and solve \eqref{eq:Bellman_discounted}. Applying It\^o's formula to $e^{-\beta t}u(X_t)$ as in Step 4 and letting $T\to\infty$, the terminal term vanishes because $e^{-\beta T}\E|u(X_T)|\leq Ce^{-(\beta-\kappa_\pi/4)T}(1+|s|^2)\to0$, and we obtain $u(s)=\E\int_0^\infty e^{-\beta t}r_\lambda^\pi(X_t)\rd t=V^\pi(s)$.
\end{proof}

Two consequences of \Cref{prop:value_function_regularity} are used repeatedly. First, by \eqref{eq:advantage_representation}, \eqref{eq:coeff_growth}, 	\Cref{ass:wellposed_policy_evaluation}(ii) and \eqref{eq:value_polynomial_growth},
\begin{equation}\label{eq:advantage_growth}
    |A^\pi(s,a)|\leq C_\pi(1+|s|+|a|)^3,
    \qquad (s,a)\in\S\times\A,
\end{equation}
the cubic term coming from $\Sigma(s,a):\ns^2V^\pi(s)$. Second, by \eqref{eq:polynomial_integration},
\begin{equation}\label{eq:advantage_integrability}
    \int_\A|A^\pi(s,a)|\pi(a|s)\nu(\rd a)\leq C_\pi(1+|s|)^3,
    \qquad s\in\S .
\end{equation}
 
\begin{remark}[On the bound for the Hessian]\label{rem:hessian_sharpness}
The linear growth of $\ns^2V^\pi$ in \eqref{eq:value_polynomial_growth} cannot be improved to a uniform bound without further structure. The term responsible is $\ns r_\lambda^\pi(X_t)^\top H_t$ in Step 3, which is of order $1+|s|$ whenever the second derivatives of the closed-loop coefficients do not vanish. For instance, the Hessian is bounded, and the advantage is quadratic, in the two situations that occur in practice: when $b^\pi$ is affine and $\sigma^\pi$ is constant, so that $H_t\equiv0$, and when $\ns r_\lambda^\pi$ is bounded.
\end{remark}

\begin{remark}[Polynomial integration]\label{rem:polynomial_integration}
Two consequences of \Cref{ass:wellposed_policy_evaluation} are used throughout. First, integrating a function of joint degree at most eight against an admissible policy gives a function of the state of the same degree: if $|\Psi(s,a)|\leq C_\Psi(1+|s|+|a|)^k$ with $k\leq8$, then
\begin{equation}\label{eq:polynomial_integration}
    \int_\A|\Psi(s,a)|\pi(a|s)\nu(\rd a)\leq C(1+|s|)^k,
    \qquad s\in\S .
\end{equation}
Indeed, $(1+|s|+|a|)^k\leq2^k\bigl((1+|s|)^k+|a|^k\bigr)$, and by Jensen's inequality and \eqref{eq:policy_moment},
\[
    \int_\A|a|^k\pi(a|s)\nu(\rd a)
    \leq
    \left(\int_\A|a|^8\pi(a|s)\nu(\rd a)\right)^{k/8}
    \leq
    C(1+|s|)^k .
\]
Second, by \eqref{eq:state_moment_growth}, \eqref{eq:discount_condition} and Fubini's theorem, the discounted occupancy measure integrates every polynomial of degree at most eight:
\begin{equation}\label{eq:occupancy_moment}
    \int_\S(1+|s|^8)\rho^\pi(\rd s)
    \leq
\frac{\beta C_\pi}{\beta-\kappa_\pi}\int_\S(1+|s|^8)\rho_0(\rd s)<\infty .
\end{equation}
In particular, the relative entropy between two admissible policies satisfies $\KL{\widehat\pi_s}{\pi_s}\leq C(1+|s|)^4$, by \eqref{eq:policy_moment}, 	\Cref{ass:wellposed_policy_evaluation}(ii) and \eqref{eq:polynomial_integration}, and is therefore integrable against $\rho^{\widehat\pi}$.
\end{remark}
 

\begin{remark}[Well-posedness of the exploratory diffusion]\label{rem:wellposed_average_sde}
By \eqref{eq:coeff_growth} and \eqref{eq:polynomial_integration}, the averaged coefficients satisfy
\begin{equation}\label{eq:averaged_linear_growth}
    |b^\pi(s)|+\|\sigma^\pi(s)\|\leq C_\pi(1+|s|),
    \qquad s\in\S .
\end{equation}
Moreover, splitting
\[
    b^\pi(s)-b^\pi(s')
    =
    \int_\A\bigl(b(s,a)-b(s',a)\bigr)\pi(a|s)\nu(\rd a)
    +
    \int_\A b(s',a)\bigl(\pi(a|s)-\pi(a|s')\bigr)\nu(\rd a),
\]
the first integral is bounded by $C(1+|s|+|s'|)|s-s'|$ by \Cref{ass:wellposed_policy_evaluation}(i) and \eqref{eq:policy_moment}, and the second by $C(1+|s'|)(1+|s|+|s'|)^8|s-s'|$ by \eqref{eq:policy_weighted_lipschitz}; hence $b^\pi$ is Lipschitz on bounded sets, and the same argument applies to $\Sigma^\pi$. By \Cref{ass:wellposed_policy_evaluation}(i) we have $\Sigma^\pi\succeq\underline\theta I_d$, and since the principal square-root map is Lipschitz on $\{A\succeq\underline\theta I_d\}$, the matrix $\sigma^\pi$ is Lipschitz on bounded sets as well. Consequently \eqref{eq:average_SDE} has a unique strong solution for every initial condition, nonexplosive by \eqref{eq:averaged_linear_growth}; see, for instance, \cite[Chapter 5]{oksendal2003sde}.
\end{remark}

\begin{lemma}[Growth rates of the averaged dynamics]\label{lem:closed_loop_rates}
Assume \Cref{ass:wellposed_policy_evaluation}(i)--(iii) and let $\pi\in\Pi$. Set
\begin{equation}\label{eq:dissipativity}
    \nu_\pi
    :=
    \max\left\{0,\
    \sup_{s\in\S}\ \sup_{|v|=1}\
    \Bigl(\inner{\ns b^\pi(s)v}{v}+\tfrac32\|\ns\sigma^\pi(s)\|^2\Bigr)\right\},
\end{equation}
which is finite because $\ns b^\pi$ and $\ns\sigma^\pi$ are bounded. Then there are $\kappa_\pi\geq0$, depending only on the constant in \eqref{eq:averaged_linear_growth}, and $C_\pi>0$ such that, for every $t\geq0$ and $s\in\S$,
\begin{equation}\label{eq:state_moment_growth}
    \E\left[1+|s_t|^8\,\big|\,s_0=s\right]
    \leq
    C_\pi(1+|s|)^8e^{\kappa_\pi t},
\end{equation}
and, denoting by $J_t:=\ns s_t$ and $H_t:=\ns^2s_t$ the first two derivatives of the flow map $s\mapsto s_t$,
\begin{equation}\label{eq:flow_derivative_growth}
    \E\|J_t\|^4
    \leq
    C_\pi e^{4\nu_\pi t},
    \qquad
    \E\|H_t\|^2
    \leq
    C_{\pi,\eps}e^{(4\nu_\pi+\eps)t}
    \quad\text{for every }\eps>0,
\end{equation}
both uniformly in $s$.
\end{lemma}
 
\begin{proof}
Estimate \eqref{eq:state_moment_growth} is the classical moment bound for a stochastic differential equation whose coefficients have linear growth: applying It\^o's formula to $(1+|s_t|^2)^4$, using \eqref{eq:averaged_linear_growth} and Gronwall's inequality gives the claim, with $\kappa_\pi$ an explicit multiple of the constant in \eqref{eq:averaged_linear_growth}; see, for instance, \cite[Chapter 2]{mao2008sde} or \cite[Chapter 2]{krylov1980controlled}.
 
For \eqref{eq:flow_derivative_growth}, note first that under \Cref{ass:wellposed_policy_evaluation}(iii) the coefficients of \eqref{eq:average_SDE} are twice continuously differentiable with bounded derivatives, so the stochastic flow $s\mapsto s_t$ is twice continuously differentiable and $J$ and $H$ solve the linear equations
\[
\begin{aligned}
    \rd J_t
    &=
    \ns b^\pi(s_t)J_t\,\rd t
    +
    \ns\sigma^\pi(s_t)J_t\,\rd W_t,
    &&J_0=I_d,\\
    \rd H_t
    &=
    \bigl(\ns b^\pi(s_t)H_t+\ns^2b^\pi(s_t)(J_t,J_t)\bigr)\rd t
    +
    \bigl(\ns\sigma^\pi(s_t)H_t+\ns^2\sigma^\pi(s_t)(J_t,J_t)\bigr)\rd W_t,
    &&H_0=0;
\end{aligned}
\]
see \cite[Chapter 4]{kunita1990flows}. Fix $v\in\R^d$ with $|v|=1$ and set $Y_t:=J_tv$. Applying It\^o's formula to $|Y_t|^4$ and using \eqref{eq:dissipativity} gives $\frac{\rd}{\rd t}\E|Y_t|^4
    \leq
    4\nu_\pi\E|Y_t|^4.$
Hence $\E|J_tv|^4\leq e^{4\nu_\pi t}$. Applying this estimate to a fixed basis of $\R^d$ and using equivalence of matrix norms yields
$\E\|J_t\|^4
    \leq
    C_\pi e^{4\nu_\pi t}.$
Since $\ns^2b^\pi$ and $\ns^2\sigma^\pi$ are bounded, the forcing terms in the equation for $H_t$ are bounded by $C\|J_t\|^2$. Standard moment estimates for this linear stochastic equation, together with the preceding bound, give $\E\|H_t\|^2
    \leq
    C_{\pi,\eps}e^{(4\nu_\pi+\eps)t},$ which proves \eqref{eq:flow_derivative_growth}.
\end{proof}

\subsection{Proof of the main results}

\begin{proof}[\textbf{Proof of \Cref{prop:adv}}]
By \Cref{prop:value_function_regularity}, $V^\pi\in C^2(\S)$ and satisfies \eqref{eq:value_polynomial_growth}. It\^o's formula applied to $t\mapsto e^{-\beta t}V^\pi(s_t^a)$ gives
\begin{equation}\label{eq:ito_expansion_eq}
\begin{aligned}
\rd\parentheses{e^{-\beta t}V^\pi(s_t^a)}
={}&
e^{-\beta t}
\parentheses{
    [\mL V^\pi](s_t^a,a)
    -
    \beta V^\pi(s_t^a)
}
\rd t+
e^{-\beta t}
\ns V^\pi(s_t^a)^\top
\sigma(s_t^a,a)
\rd B_t.
\end{aligned}
\end{equation}
By \eqref{eq:coeff_growth} and \eqref{eq:coeff_lipschitz}, for fixed $a$ the coefficients of $s^a$ are Lipschitz in the state and of linear growth. Applying It\^o's formula to $|s_t^a|^p$ and Gronwall's inequality therefore gives, for every $p\geq2$ and $T>0$,
\begin{equation}\label{eq:moment_estimate_fixed_action}
    \sup_{0\leq t\leq T}\E|s_t^a|^p
    \leq
    C_{p,T}(1+|s|^p+|a|^p).
\end{equation}
Since $\|\sigma(x,a)\|\leq C(1+|x|+|a|)$ and $|\ns V^\pi(x)|\leq C(1+|x|)$, the case $p=4$ of \eqref{eq:moment_estimate_fixed_action} yields
\begin{align*}
    \E\left[
    \int_0^\dt
    e^{-2\beta t}
    \left|
        \sigma(s_t^a,a)^\top\ns V^\pi(s_t^a)
    \right|^2
    \rd t
\right]
<\infty.
\end{align*}
Hence the stochastic integral is a square-integrable martingale and has zero conditional expectation. Taking expectation in \eqref{eq:ito_expansion_eq} and substituting the resulting identity into \eqref{eq:advantage_definition}, we obtain
\[
A^\pi(s,a)
=
\lim_{\dt\to0^+}
\frac{1}{\dt}
\E\left[
    \int_0^\dt e^{-\beta t}
    \Psi_a(s_t^a)
    \rd t
    \,\Big|\,
    s_0^a=s
\right],
\]
where $\Psi_a(x)
    :=
    r(x,a)
    -
    \lambda\log\pi(a|x)
    +[\mL V^\pi](x,a)
    -
    \beta V^\pi(x).$ Moreover, using the integral form of the SDE, the Cauchy--Schwarz inequality, It\^o's isometry and \eqref{eq:moment_estimate_fixed_action}, we obtain, for $t\leq1$,
\[
\begin{aligned}
\E|s_t^a-s|^2
\leq
2t\int_0^t\E|b(s_u^a,a)|^2\,\rd u
+
2\int_0^t\E\|\sigma(s_u^a,a)\|^2\,\rd u\leq
C(1+|s|^2+|a|^2)(t^2+t).
\end{aligned}
\]
Consequently, $s_t^a\to s$ in $L^2$ as $t\to 0$, and in particular in probability. For fixed $a$, the map $\Psi_a$ is continuous, by \eqref{eq:coeff_growth}, the continuity of $\log\pi$ and \eqref{eq:value_polynomial_growth}, and has at most cubic growth, $|\Psi_a(x)|\leq C_a(1+|x|^3),$ where we used 	\Cref{ass:wellposed_policy_evaluation}(ii) for the entropy term. By \eqref{eq:moment_estimate_fixed_action} with $p=6$, the family $\{\Psi_a(s_t^a)\}_{t\in[0,1]}$ is bounded in $L^2$ and therefore uniformly integrable, so $\E[\Psi_a(s_t^a)]\to\Psi_a(s)$ as $t\to0^+$. The map $t\mapsto e^{-\beta t}\E[\Psi_a(s_t^a)]$ is then bounded on $[0,1]$ and continuous at $t=0$, so its averages over $[0,\dt]$ converge to $\Psi_a(s)$, which is \eqref{eq:advantage_representation}. Finally, integrating \eqref{eq:advantage_representation} against $\pi(a|s)\nu(\rd a)$, which is legitimate by \eqref{eq:advantage_integrability}, and using \eqref{eq:regularized_reward}, \eqref{eq:averaged_coefficients} and \eqref{eq:Bellman_discounted}, we deduce \eqref{eq:centered_advantage}.
\end{proof}

\begin{lemma}[Discounted resolvent identity]\label{lem:discounted_resolvent}
Assume \Cref{ass:wellposed_policy_evaluation}. For every $\pi\in\Pi$ and every $f\in C^2(\S)$ satisfying
\begin{equation}\label{eq:resolvent_growth}
    |f(s)|\leq C_f(1+|s|^2),
    \qquad
    |\ns f(s)|\leq C_f(1+|s|),
    \qquad
    \|\ns^2f(s)\|\leq C_f(1+|s|),
\end{equation}
we have
\begin{equation*}
\int_\S
\parentheses{
    \beta f(s)
    -
    \L_{b^\pi,\Sigma^\pi}f(s)
}
\rho^\pi(\rd s)
=
\beta\int_\S f(s)\rho_0(\rd s).
\end{equation*}
\end{lemma}
 
\begin{proof}
Fix $T>0$. By \Cref{rem:wellposed_average_sde}, $\|\sigma^\pi(s)\|\leq C(1+|s|)$, so $|\sigma^\pi(s)^\top\ns f(s)|\leq C(1+|s|^2)$ and, by \eqref{eq:state_moment_growth}, the stochastic integral appearing in It\^o's formula is a square-integrable martingale on $[0,T]$. Applying It\^o's formula to $e^{-\beta t}f(s_t)$ and taking expectations conditionally on $s_0=s$ gives
\[
\begin{aligned}
\E\left[e^{-\beta T}f(s_T)\,\big|\,s_0=s\right]
-f(s)
=
\E\left[
    \int_0^T
    e^{-\beta t}
    \parentheses{
        \L_{b^\pi,\Sigma^\pi}f(s_t)
        -
        \beta f(s_t)
    }
    \rd t
    \,\Big|\,
    s_0=s
\right].
\end{aligned}
\]
We integrate with respect to $\rho_0(\rd s)$ and let $T\to\infty$. For the terminal term, \eqref{eq:resolvent_growth}, Jensen's inequality and \eqref{eq:state_moment_growth} give
\[
    e^{-\beta T}
    \int_\S
    \E\left[|f(s_T)|\,\big|\,s_0=s\right]
    \rho_0(\rd s)
    \leq
    Ce^{-(\beta-\kappa_\pi/4)T}
    \int_\S(1+|s|^2)\rho_0(\rd s)
    \longrightarrow0,
\]
by \eqref{eq:discount_condition} and because $\rho_0$ has a finite eighth moment. For the right-hand side, \Cref{rem:wellposed_average_sde} and \eqref{eq:resolvent_growth} give
\[
    \left|
        \beta f(s)-\L_{b^\pi,\Sigma^\pi}f(s)
    \right|
    \leq
    C_f(1+|s|)^3,
\]
which is $\rho^\pi$-integrable by \eqref{eq:occupancy_moment}, so dominated convergence applies. The result follows after multiplying by $\beta$, since by the definition of the discounted occupancy measure $\int_\S\E\bigl[\int_0^\infty e^{-\beta t}g(s_t)\,\rd t\,|\,s_0=s\bigr]\rho_0(\rd s)=\frac1\beta\int_\S g(s)\rho^\pi(\rd s)$ for every $\rho^\pi$-integrable $g$.
\end{proof}

\begin{proof}[Proof of \Cref{lem:performance_difference}.]
We give the details omitted from the proof sketch in the main text.
By \eqref{eq:closed_loop_reward_regularity} and Taylor's theorem,
the regularized reward satisfies
$|r_\lambda^{\widehat\pi}(s)|\leq C(1+|s|^2)$,
which is $\rho^{\widehat\pi}$-integrable by
\eqref{eq:occupancy_moment}. Hence, Fubini's theorem and the
definition of the discounted occupancy measure give
\begin{equation}\label{eq:j_widehatpi}
J[\widehat\pi]
=
\frac{1}{\beta}
\int_\S r_\lambda^{\widehat\pi}(s)
\rho^{\widehat\pi}(\rd s).
\end{equation}
Applying \Cref{lem:discounted_resolvent} to the policy
$\widehat\pi$ and the function $f=V^\pi$, which satisfies
\eqref{eq:resolvent_growth} by
\eqref{eq:value_polynomial_growth}, gives
\begin{equation}\label{eq:J_pi_expanded}
J[\pi]
=
\int_\S V^\pi(s)\rho_0(\rd s)
=
\frac{1}{\beta}
\int_\S
\parentheses{
    \beta V^\pi(s)
    -
    \L_{b^{\widehat\pi},\Sigma^{\widehat\pi}}V^\pi(s)
}
\rho^{\widehat\pi}(\rd s).
\end{equation}
Subtracting \eqref{eq:J_pi_expanded} from
\eqref{eq:j_widehatpi} yields
\[
J[\widehat\pi]-J[\pi]
=
\frac{1}{\beta}
\int_\S
\parentheses{
    r_\lambda^{\widehat\pi}(s)
    +
    \L_{b^{\widehat\pi},\Sigma^{\widehat\pi}}V^\pi(s)
    -
    \beta V^\pi(s)
}
\rho^{\widehat\pi}(\rd s).
\]
Expanding the averaged reward and generator, the expression inside
the outer integral becomes
\[
\int_\A
\Bigl(
    r(s,a)
    -
    \lambda\log\widehat\pi(a|s)
    +
    [\mL V^\pi](s,a)
    -
    \beta V^\pi(s)
\Bigr)
\widehat\pi(a|s)\nu(\rd a).
\]
This expansion is legitimate because the terms have at most the
polynomial growth controlled by
\eqref{eq:polynomial_integration} and
\eqref{eq:occupancy_moment}.

By \eqref{eq:advantage_representation},
\[
r(s,a)
+
[\mL V^\pi](s,a)
-
\beta V^\pi(s)
=
A^\pi(s,a)
+
\lambda\log\pi(a|s).
\]
Therefore,
\begin{align*}
&r(s,a)
-\lambda\log\widehat\pi(a|s)
+[\mL V^\pi](s,a)
-\beta V^\pi(s)
=
A^\pi(s,a)
-
\lambda
\log\frac{\widehat\pi(a|s)}{\pi(a|s)}.
\end{align*}
Substituting this identity above and using the definition of the
relative entropy gives
\[
J[\widehat\pi]-J[\pi]
=
\frac{1}{\beta}
\int_\S
\left[
    \int_\A
    A^\pi(s,a)\widehat\pi(a|s)\nu(\rd a)
    -
    \lambda\KL{\widehat\pi_s}{\pi_s}
\right]
\rho^{\widehat\pi}(\rd s),
\]
which is \eqref{eq:performance_difference}.
All terms are integrable by
\eqref{eq:polynomial_integration},
\eqref{eq:occupancy_moment}, and
\Cref{rem:relative_entropy_integrability}.
\end{proof}

\begin{remark}[Consequences of admissibility]
\label{rem:admissible_policy_direction}
Let $\phi\in T_\pi\Pi$, and let $\ve_0>0$ be as in
\Cref{def:admissible_policy_direction}. Since both
$\pi+\ve_0\phi$ and $\pi-\ve_0\phi$ are strictly positive, we have
\[
    |\phi(a|s)|
    \leq
    \frac{1}{\ve_0}\pi(a|s),
    \qquad
    (s,a)\in\S\times\A.
\]
In particular $\pi^\ve\leq2\pi$ for every $|\ve|\leq\ve_0$, so the constant in \eqref{eq:policy_moment} can be chosen uniformly in $\ve$ and \eqref{eq:polynomial_integration} applies to the whole family $\{\pi^\ve\}_{|\ve|\leq\ve_0}$ with a single constant.
 
Moreover, since both $\pi+\ve_0\phi$ and $\pi-\ve_0\phi$ belong to $\Pi$, they satisfy \eqref{eq:policy_moment} and \eqref{eq:policy_weighted_lipschitz}. Let $M_+$ and $M_-$ denote the constants in \eqref{eq:policy_moment} associated with $\pi+\ve_0\phi$ and $\pi-\ve_0\phi$, and let $L_+$ and $L_-$ denote the corresponding constants in \eqref{eq:policy_weighted_lipschitz}. Since $ \phi=((\pi+\ve_0\phi)-(\pi-\ve_0\phi))/(2\ve_0),$
the triangle inequality yields
\begin{equation}\label{eq:admissible_direction_bounds}
\begin{aligned}
    \int_\A
    (1+|a|^8)|\phi(a|s)|
    \nu(\rd a)
    &\leq
    \frac{M_++M_-}{2\ve_0}(1+|s|^8),\\
    \int_\A
    (1+|a|^8)
    |\phi(a|s)-\phi(a|s')|
    \nu(\rd a)
    &\leq
    \frac{L_++L_-}{2\ve_0}(1+|s|^8+|s'|^8)|s-s'|.
\end{aligned}
\end{equation}
 
A simple sufficient condition for $\phi\in T_\pi\Pi$ is the existence of
a bounded continuous function $g:\S\times\A\to\R$, globally Lipschitz in
the state variable uniformly in the action, such that
\[
    \phi(a|s)=\pi(a|s)g(s,a),
    \qquad
    \int_\A g(s,a)\pi(a|s)\nu(\rd a)=0.
\]
Indeed, $\pi^\ve(a|s)=\pi(a|s)\bigl(1+\ve g(s,a)\bigr)$, which is normalized and remains strictly positive whenever $|\ve|\|g\|_\infty<1$. The boundedness and Lipschitz continuity of $g$ also preserve the conditions in \Cref{ass:wellposed_policy_evaluation}(ii) for sufficiently small $|\ve|$, with constants uniform in $\ve$; the remaining requirement is that \Cref{ass:wellposed_policy_evaluation}(iii) hold uniformly along the perturbed family, which is a condition on the perturbed closed-loop dynamics rather than on the density $\pi^\ve$ itself.
\end{remark}

\begin{lemma}[Continuity of the discounted occupancy measure]
\label{lem:occupancy_continuity}
Assume \Cref{ass:wellposed_policy_evaluation}. Let $\pi\in\Pi$,
$\phi\in T_\pi\Pi$, and let $\pi^\ve$ be defined by
\eqref{eq:additive_policy_perturbation}. Then, for every continuous function $f:\S\to\R$ satisfying $|f(s)|\leq C_f(1+|s|^6)$,
\begin{equation}\label{eq:occupancy_weak_continuity}
    \lim_{\ve\to0}
    \int_\S f(s)\rho^{\pi^\ve}(\rd s)
    =
    \int_\S f(s)\rho^\pi(\rd s).
\end{equation}
\end{lemma}
 
\begin{proof}
Define $b^\phi(s)
    :=
    \int_\A b(s,a)\phi(a|s)\nu(\rd a),$ and $\Sigma^\phi(s)
    :=
    \int_\A \Sigma(s,a)\phi(a|s)\nu(\rd a).$ By \eqref{eq:coeff_growth}, \Cref{rem:admissible_policy_direction} and \eqref{eq:polynomial_integration}, these functions satisfy $|b^\phi(s)|\leq C(1+|s|)$ and $\|\Sigma^\phi(s)\|\leq C(1+|s|^2)$. Since $b^{\pi^\ve}=b^\pi+\ve b^\phi$ and $\Sigma^{\pi^\ve}=\Sigma^\pi+\ve\Sigma^\phi$, it follows that, for every $R>0$,
\begin{equation}\label{eq:local_uniform_coefficient_convergence}
    \sup_{|s|\leq R}
    \left(
        |b^{\pi^\ve}(s)-b^\pi(s)|
        +
        \|\Sigma^{\pi^\ve}(s)-\Sigma^\pi(s)\|
    \right)
    \leq
    C(1+R^2)|\ve|
    \longrightarrow0
    \qquad\text{as }\ve\to0.
\end{equation}
Furthermore, since $\pi^\ve\leq2\pi$, the argument of \Cref{rem:wellposed_average_sde} applies to the whole family $\{\pi^\ve\}_{|\ve|\leq\ve_0}$ with constants independent of $\ve$: the coefficients $b^{\pi^\ve}$ and $\Sigma^{\pi^\ve}$ have linear and quadratic growth respectively and are Lipschitz on bounded sets, uniformly in $\ve$. By \Cref{ass:wellposed_policy_evaluation}(i) all the matrices $\Sigma^{\pi^\ve}$ are uniformly elliptic with the same constant, and the principal square-root map satisfies $\|A^{1/2}-B^{1/2}\|\leq(2\sqrt{\underline\theta})^{-1}\|A-B\|$ for $A,B\succeq\underline\theta I_d$; hence the same local convergence and the same growth and Lipschitz control hold for $\sigma^{\pi^\ve}$.
 
Let $s_t^\ve$ and $s_t$ denote the solutions of the SDEs associated with $\pi^\ve$ and $\pi$, constructed with the same initial condition $s$ and the same Brownian motion, and set $e_t:=s_t^\ve-s_t$. Fix $T,R>0$ and let $\tau_R:=\inf\{t\geq0:|s_t|\vee|s_t^\ve|\geq R\}$. On $[0,\tau_R]$ the coefficients are Lipschitz with a constant $L_R$ independent of $\ve$, so It\^o's formula applied to $|e_t|^2$ and Gronwall's inequality give
\[
    \sup_{t\leq T}\E\left[|e_t|^2\mathbf 1_{\{\tau_R>t\}}\right]
    \leq
    C_Te^{CL_R^2T}
    \sup_{|s|\leq R}
    \left(
        |b^{\pi^\ve}(s)-b^\pi(s)|^2
        +
        \|\sigma^{\pi^\ve}(s)-\sigma^\pi(s)\|^2
    \right).
\]
On the complementary event, the Cauchy--Schwarz inequality gives
\[
    \E\left[
        |e_t|^2\mathbf 1_{\{\tau_R\leq T\}}
    \right]
    \leq
    \left(\E|e_t|^4\right)^{1/2}
    \mathbb P(\tau_R\leq T)^{1/2},
    \qquad t\leq T.
\]
By \eqref{eq:state_moment_growth}, which holds uniformly in $\ve$, we have $(\E|e_t|^4)^{1/2}\leq C_T(1+|s|^2)$. Moreover, since the coefficients have linear growth uniformly in $\ve$, the maximal moment estimate \cite[Chapter 2, Theorem 4.4]{mao2008sde} gives $\E\sup_{t\leq T}\bigl(|s_t|^8+|s_t^\ve|^8\bigr)\leq C_T(1+|s|^8)$, and Markov's inequality yields $\mathbb P(\tau_R\leq T)\leq C_T(1+|s|^8)R^{-8}$. Hence
\[
    \E\left[
        |e_t|^2\mathbf 1_{\{\tau_R\leq T\}}
    \right]
    \leq
    \frac{C_T(1+|s|^6)}{R^4},
    \qquad t\leq T.
\]
Combining the two bounds, letting $\ve\to0$ using \eqref{eq:local_uniform_coefficient_convergence}, and then letting $R\to\infty$, we obtain $\sup_{0\leq t\leq T}\E|s_t^\ve-s_t|^2\to0$ as $\ve\to0$. In particular, $s_t^\ve\to s_t$ in probability for every fixed $t\geq0$.
 
Let now $f$ be continuous with $|f(s)|\leq C_f(1+|s|^6)$. By \eqref{eq:state_moment_growth}, which holds uniformly in $\ve$ by \Cref{def:admissible_policy_direction},
\[
    \sup_{|\ve|\leq\ve_0}
    \E\left[|f(s_t^\ve)|^{4/3}\right]
    \leq
    C\sup_{|\ve|\leq\ve_0}\E\left[1+|s_t^\ve|^8\right]
    <\infty,
\]
so the family $\{f(s_t^\ve)\}_{|\ve|\leq\ve_0}$ is uniformly integrable and therefore $\E[f(s_t^\ve)]\to\E[f(s_t)]$ for every fixed $t\geq0$. Finally, set $\bar\kappa
    :=
    \sup_{|\ve|\leq\ve_0}\kappa_{\pi^\ve}.$ By the uniformity required in
\Cref{def:admissible_policy_direction}, we have
$\bar\kappa<\beta$. Hence, by
\eqref{eq:state_moment_growth} and Jensen's inequality,
\[
    \left|
        \E\left[
            f(s_t^\ve)\,\big|\,s_0=s
        \right]
    \right|
    \leq
    C(1+|s|^6)e^{3\bar\kappa t/4},
\]
uniformly in $|\ve|\leq\ve_0$. Since $3\bar\kappa/4<\beta$, the product of the right-hand side with $e^{-\beta t}$ is integrable on $[0,\infty)$ and is also $\rho_0$-integrable in $s$. Dominated convergence in time, followed by integration with respect to $\rho_0$, proves \eqref{eq:occupancy_weak_continuity}.
\end{proof}

\begin{proof}[\textbf{Proof of \Cref{thm:PG}}]
Let $\pi^\ve:=\pi+\ve\phi$. By \Cref{rem:admissible_policy_direction}, the function $g(a|s):=\phi(a|s)/\pi(a|s)$ is bounded and continuous, and $\pi^\ve(a|s)=\pi(a|s)(1+\ve g(a|s))$. Since $\pi,\pi^\ve\in\Pi$, \Cref{lem:performance_difference} applies and gives
\begin{equation}\label{eq:PG_difference_quotient_start}
J[\pi^\ve]-J[\pi]
=
\frac{1}{\beta}\int_\S\int_\A
\left(
A^\pi(s,a)
-
\lambda\log\frac{\pi^\ve(a|s)}{\pi(a|s)}
\right)
\pi^\ve(a|s)
\nu(\rd a)
\rho^{\pi^\ve}(\rd s).
\end{equation}
For $\ve\neq0$, define
\[
F_\ve(s):=
\frac{1}{\ve}
\int_\A
\left(
A^\pi(s,a)
-
\lambda\log\frac{\pi^\ve(a|s)}{\pi(a|s)}
\right)
\pi^\ve(a|s)
\nu(\rd a),
\qquad
F(s):=
\int_\A A^\pi(s,a)\phi(a|s)\nu(\rd a).
\]
Both integrals converge absolutely by \eqref{eq:advantage_integrability}, $|\phi|\leq\ve_0^{-1}\pi$ and $\pi^\ve\leq2\pi$. Dividing \eqref{eq:PG_difference_quotient_start} by $\ve$ and multiplying by $\beta$, \eqref{eq:PG_pi_discounted} is equivalent to showing that $\int_\S F_\ve(s)\rho^{\pi^\ve}(\rd s)
\longrightarrow
\int_\S F(s)\rho^\pi(\rd s).$ We decompose the difference as
\begin{equation}\label{eq:PG_error_decomposition}
\begin{aligned}
    \int_\S F_\ve(s)\rho^{\pi^\ve}(\rd s)
-
\int_\S F(s)\rho^\pi(\rd s)
&=
\int_\S\bigl(F_\ve(s)-F(s)\bigr)\rho^{\pi^\ve}(\rd s)
+
\left(
\int_\S F(s)\rho^{\pi^\ve}(\rd s)
-
\int_\S F(s)\rho^\pi(\rd s)
\right)
\\
&=:\mathsf A_\ve+\mathsf B_\ve
\end{aligned}
\end{equation}
We first prove that $\mathsf A_\ve\to0$. Using $\pi^\ve=\pi(1+\ve g)$ and \eqref{eq:centered_advantage}, we obtain
\[
F_\ve(s)
=
F(s)
-
\frac{\lambda}{\ve}
\int_\A
\pi(a|s)
\bigl(1+\ve g(a|s)\bigr)
\log\bigl(1+\ve g(a|s)\bigr)
\nu(\rd a).
\]
Let $H(x):=(1+x)\log(1+x)-x$. The function $H$ satisfies $|H(x)|\leq Cx^2$ on every compact subset of $(-1,\infty)$. Since $g$ is uniformly bounded, for all sufficiently small $|\ve|$ the values $\ve g(a|s)$ belong to a fixed compact subset of $(-1,\infty)$. Moreover, we have that $\int_\A g(a|s)\pi(a|s)\nu(\rd a)
=
\int_\A\phi(a|s)\nu(\rd a)
=
0.$ Consequently,
\[
\begin{aligned}
\left|
\frac{1}{\ve}
\int_\A
\pi(a|s)
\bigl(1+\ve g(a|s)\bigr)
\log\bigl(1+\ve g(a|s)\bigr)
\nu(\rd a)
\right|
&=
\left|
\frac{1}{\ve}
\int_\A
\pi(a|s)H\bigl(\ve g(a|s)\bigr)
\nu(\rd a)
\right|\leq
C|\ve|\|g\|_\infty^2,
\end{aligned}
\]
uniformly in $s$, because $\pi(\cdot|s)$ is a probability density. Therefore, since $\rho^{\pi^\ve}(\S)=1$, we have $|\mathsf A_\ve|
\leq
\sup_{s\in\S}|F_\ve(s)-F(s)|
\longrightarrow0.$ We next prove that $\mathsf B_\ve\to0$. By \eqref{eq:advantage_integrability} and $|\phi|\leq\ve_0^{-1}\pi$,
\[
|F(s)|
\leq
\frac{1}{\ve_0}
\int_\A
|A^\pi(s,a)|\pi(a|s)
\nu(\rd a)
\leq
C(1+|s|)^3,
\]
so $F$ has at most cubic growth. We now check that $F$ is continuous. Write $A^\pi=\Psi-\lambda\log\pi$, where
\[
    \Psi(s,a)
    :=
    r(s,a)+b(s,a)\cdot\ns V^\pi(s)+\tfrac12\Sigma(s,a):\ns^2V^\pi(s)-\beta V^\pi(s),
\]
so that $|\Psi(s,a)|\leq C(1+|s|+|a|)^3$ by \eqref{eq:coeff_growth} and \eqref{eq:value_polynomial_growth}. For the term involving $\Psi$ we decompose
\[
\begin{aligned}
&\int_\A\Psi(s',a)\phi(a|s')\nu(\rd a)
-
\int_\A\Psi(s,a)\phi(a|s)\nu(\rd a)\\
&\qquad=
\int_\A\bigl(\Psi(s',a)-\Psi(s,a)\bigr)\phi(a|s')\nu(\rd a)
+
\int_\A\Psi(s,a)\bigl(\phi(a|s')-\phi(a|s)\bigr)\nu(\rd a).
\end{aligned}
\]
For $|s|,|s'|\leq R$, the Lipschitz bounds \Cref{ass:wellposed_policy_evaluation}(i) and the continuity of $\ns V^\pi$ and $\ns^2V^\pi$ give $|\Psi(s',a)-\Psi(s,a)|\leq C_R(1+|a|^2)\eta_R(|s-s'|)$ with $\eta_R(\varrho)\to0$ as $\varrho\to0$, so the first integral tends to zero by \eqref{eq:admissible_direction_bounds}. The second tends to zero by the second bound in \eqref{eq:admissible_direction_bounds}, since $|\Psi(s,\cdot)|\leq C_s(1+|a|^3)\leq C_s(1+|a|^8)$ for fixed $s$. For the entropy term, using $\phi=\pi g$, we write $\phi(a|s)\log\pi(a|s)=g(a|s)\pi(a|s)\log\pi(a|s).$ If $s'\to s$, then
\[
\begin{aligned}
&\int_\A
g(a|s')\pi(a|s')\log\pi(a|s')
\nu(\rd a)
-
\int_\A
g(a|s)\pi(a|s)\log\pi(a|s)
\nu(\rd a)\\
&=
\int_\A
g(a|s')
\left(
\pi(a|s')\log\pi(a|s')
-
\pi(a|s)\log\pi(a|s)
\right)
\nu(\rd a)+
\int_\A
\left(
g(a|s')-g(a|s)
\right)
\pi(a|s)\log\pi(a|s)
\nu(\rd a).
\end{aligned}
\]
The first term converges to zero by \eqref{eq:policy_entropy_lipschitz} and the boundedness of $g$, while the second converges to zero by dominated convergence, the dominating function $2\|g\|_\infty|\pi\log\pi|(\cdot|s)$ being $\nu$-integrable by \eqref{eq:policy_moment}. Thus $F$ is continuous with at most cubic growth, so \Cref{lem:occupancy_continuity} applies and yields
\[
\mathsf B_\ve
=
\int_\S F(s)\rho^{\pi^\ve}(\rd s)
-
\int_\S F(s)\rho^\pi(\rd s)
\longrightarrow0.
\]
 
Combining the convergence of $\mathsf A_\ve$ and $\mathsf B_\ve$ in \eqref{eq:PG_error_decomposition}, we conclude that
\[
\lim_{\ve\to0}\frac{J[\pi^\ve]-J[\pi]}{\ve}
=
\frac{1}{\beta}\int_\S F(s)\rho^\pi(\rd s)
=
\frac{1}{\beta}\int_\S\int_\A
A^\pi(s,a)\phi(a|s)
\nu(\rd a)
\rho^\pi(\rd s),
\]
which proves \eqref{eq:PG_pi_discounted}.
\end{proof}

\begin{proof}[\textbf{Proof of \Cref{cor:parametric_policy_gradient}}]
Fix $\theta\in\Theta$ and a compact set $K\subset\Theta$ containing a neighbourhood of $\theta$, so that $\theta+z\in K$ for all sufficiently small $|z|$. All the constants below depend on $K$ but not on $z$ or $s$. By the definition of the gradient, to prove \eqref{eq:PG_omega_discounted} it suffices to show that, for every increment $z\in\R^p$,
\begin{equation}\label{eq:equiv_characta_gradient}
    J[\pi^{\theta+z}]-J[\pi^\theta]
=
\frac{1}{\beta}\left(
\int_\S\int_\A
A^{\pi^\theta}(s,a)
\nt\pi^\theta(a|s)
\nu(\rd a)
\rho^{\pi^\theta}(\rd s)
\right)\cdot z
+
o(|z|)
\end{equation}
as $z\to0$.
 
\emph{Step 1: expansion of the log-policy.} By the fundamental theorem of calculus,
\begin{equation}\label{eqr:expansion_logpi}
\log\frac{\pi^{\theta+z}(a|s)}{\pi^\theta(a|s)}
=
\int_0^1
\nt\log\pi^{\theta+t z}(a|s)\cdot z
\,\rd t
=
\nt\log\pi^\theta(a|s)\cdot z
+
R_z(s,a),
\end{equation}
where \eqref{eq:score_parameter_continuity} gives
\begin{equation}\label{eq:parametric_log_remainder}
|R_z(s,a)|
\leq
|z|\,\o_K(|z|)(1+|s|^2+|a|^2).
\end{equation}
Since $\pi^\theta,\pi^{\theta+z}\in\Pi$, \Cref{lem:performance_difference} applies and gives
\[
J[\pi^{\theta+z}]-J[\pi^\theta]
=
\frac{1}{\beta}\int_\S F_z(s)\rho^{\pi^{\theta+z}}(\rd s),\quad \text{where}\quad F_z(s)
:=
\int_\A
\left(
A^{\pi^\theta}(s,a)
-
\lambda
\log\frac{\pi^{\theta+z}(a|s)}{\pi^\theta(a|s)}
\right)
\pi^{\theta+z}(a|s)
\nu(\rd a).
\]
 
\emph{Step 2: expansion of the density.} Applying the fundamental theorem of calculus to $\vartheta\mapsto\pi^\vartheta(a|s)$ and using $\nt\pi^\vartheta=\pi^\vartheta\nt\log\pi^\vartheta$, we obtain
\begin{equation}\label{eq:parametric_policy_expansion}
\pi^{\theta+z}(a|s)
=
\pi^\theta(a|s)
+
\nt\pi^\theta(a|s)\cdot z
+
R_z^\pi(a|s),
\qquad
R_z^\pi(a|s)
=
\int_0^1
\bigl(
    \nt\pi^{\theta+tz}
    -
    \nt\pi^{\theta}
\bigr)(a|s)\cdot z
\,\rd t .
\end{equation}
We claim that
\begin{equation}\label{eq:parametric_policy_remainder}
\int_\A
(1+|s|+|a|)^3
|R_z^\pi(a|s)|
\nu(\rd a)
=
o(|z|)(1+|s|^8),
\end{equation}
with $o(|z|)$ uniform in $s$. Set $\vartheta_t:=\theta+tz\in K$ for $t\in[0,1]$, and write
\[
    \nt\pi^{\vartheta_t}-\nt\pi^{\theta}
    =
    \pi^{\vartheta_t}\bigl(\nt\log\pi^{\vartheta_t}-\nt\log\pi^{\theta}\bigr)
    +
    \bigl(\pi^{\vartheta_t}-\pi^{\theta}\bigr)\nt\log\pi^{\theta}.
\]
By \eqref{eq:score_parameter_continuity}, the first term is bounded by $\o_K(|z|)(1+|s|^2+|a|^2)\pi^{\vartheta_t}(a|s)$. For the second term, the fundamental theorem of calculus, $\nt\pi^{\vartheta}=\pi^{\vartheta}\nt\log\pi^{\vartheta}$ and \eqref{eq:score_polynomial_growth} give
\begin{equation}\label{eq:parametric_density_increment}
    \left|\pi^{\vartheta_t}(a|s)-\pi^{\theta}(a|s)\right|
    \leq
    C|z|\int_0^1(1+|s|^2+|a|^2)\pi^{\vartheta_u}(a|s)\,\rd u,
\end{equation}
while $|\nt\log\pi^\theta(a|s)|\leq C(1+|s|^2+|a|^2)$. Inserting these bounds into the definition of $R_z^\pi$, and using Tonelli's theorem together with \eqref{eq:polynomial_integration}, which holds uniformly for the policies $\pi^{\vartheta_u}$, $u\in[0,1]$, by \Cref{ass:assumption_pi_theta}, we obtain
\[
    \int_\A
    (1+|s|+|a|)^3
    |R_z^\pi(a|s)|
    \nu(\rd a)
    \leq
    C|z|\o_K(|z|)(1+|s|)^5
    +
    C|z|^2(1+|s|)^7,
\]
which proves \eqref{eq:parametric_policy_remainder}. Similarly, \eqref{eq:parametric_density_increment} with $t=1$ gives
\begin{equation}\label{eq:parametric_density_lipschitz}
    \int_\A
    (1+|a|^2)
    \left|
        \pi^{\theta+z}(a|s)-\pi^\theta(a|s)
    \right|
    \nu(\rd a)
    \leq
    C|z|(1+|s|^4).
\end{equation}
 
\emph{Step 3: continuity of the occupancy measure.} By \eqref{eq:coeff_growth} and \eqref{eq:parametric_density_lipschitz},
\[
    |b^{\pi^{\theta+z}}(s)-b^{\pi^\theta}(s)|
    +
    \|\Sigma^{\pi^{\theta+z}}(s)-\Sigma^{\pi^\theta}(s)\|
    \leq
    C|z|(1+|s|^6),
\]
so the coefficients converge uniformly on compact sets as $z\to0$. The compact-uniform constants of \Cref{ass:assumption_pi_theta} give growth, local Lipschitz and state-moment bounds that are uniform for $\theta+z\in K$. Repeating the proof of \Cref{lem:occupancy_continuity} verbatim therefore yields
\begin{equation}\label{eq:parametric_occupancy_continuity}
\int_\S f(s)\rho^{\pi^{\theta+z}}(\rd s)
\longrightarrow
\int_\S f(s)\rho^{\pi^\theta}(\rd s)
\end{equation}
for every continuous $f$ with $|f(s)|\leq C_f(1+|s|^6)$.
 
\emph{Step 4: first-order expansion of $F_z$.} By \eqref{eq:parametric_policy_expansion}, \eqref{eq:centered_advantage} and \eqref{eq:advantage_growth},
\[
\int_\A
A^{\pi^\theta}(s,a)
\pi^{\theta+z}(a|s)
\nu(\rd a)
=
\left(
\int_\A
A^{\pi^\theta}(s,a)
\nt\pi^\theta(a|s)
\nu(\rd a)
\right)\cdot z
+
r_{A,z}(s),
\qquad
|r_{A,z}(s)|\leq o(|z|)(1+|s|^8),
\]
where the remainder was estimated with \eqref{eq:parametric_policy_remainder}, whose weight matches the growth \eqref{eq:advantage_growth} of the advantage. For the entropy term, we first show that the score is centered. Since $|\nt\pi^\theta(a|s)|=|\nt\log\pi^\theta(a|s)|\,\pi^\theta(a|s)\leq C(1+|s|^2+|a|^2)\pi^\theta(a|s)$ by \eqref{eq:score_polynomial_growth}, the vector $v(s):=\int_\A\nt\pi^\theta(a|s)\nu(\rd a)\in\R^p$ is well defined by \eqref{eq:polynomial_integration}. Integrating \eqref{eq:parametric_policy_expansion} in $a$, and using that $\pi^{\theta+z}(\cdot|s)$ and $\pi^\theta(\cdot|s)$ are probability densities, we get $1=1+v(s)\cdot z+\int_\A R_z^\pi(a|s)\nu(\rd a)$. Hence, by \eqref{eq:parametric_policy_remainder},
\[
    |v(s)\cdot z|
    =
    \left|\int_\A R_z^\pi(a|s)\nu(\rd a)\right|
    \leq
    \int_\A(1+|s|+|a|)^3|R_z^\pi(a|s)|\nu(\rd a)
    =
    o(|z|)(1+|s|^8).
\]
If $v(s)\neq0$, choosing $z=\ve\,v(s)/|v(s)|$ with $\ve>0$ gives $\ve|v(s)|\leq o(\ve)(1+|s|^8)$. Dividing by $\ve$ and letting $\ve\to0$ yields $|v(s)|=0$, a contradiction. Therefore $v(s)=0$, that is,
\begin{equation}\label{eq:score_centering_parametric}
    \int_\A\nt\log\pi^\theta(a|s)\pi^\theta(a|s)\nu(\rd a)
    =
    \int_\A\nt\pi^\theta(a|s)\nu(\rd a)
    =0 .
\end{equation}
Combining \eqref{eqr:expansion_logpi}, \eqref{eq:score_centering_parametric}, \eqref{eq:parametric_log_remainder} and \eqref{eq:parametric_density_lipschitz},
\[
\begin{aligned}
\int_\A
\log\frac{\pi^{\theta+z}(a|s)}{\pi^\theta(a|s)}
\pi^{\theta+z}(a|s)
\nu(\rd a)
&=
\left(\int_\A\nt\log\pi^\theta(a|s)\bigl(\pi^{\theta+z}-\pi^\theta\bigr)(a|s)\nu(\rd a)\right)\cdot z
+
\int_\A R_z(s,a)\pi^{\theta+z}(a|s)\nu(\rd a)\\
&=
r_{\mathrm{KL},z}(s),
\qquad
|r_{\mathrm{KL},z}(s)|
\leq
o(|z|)(1+|s|^6),
\end{aligned}
\]
where the first integral is $\mathcal O(|z|)(1+|s|^6)$ by \eqref{eq:score_polynomial_growth} and \eqref{eq:parametric_density_lipschitz}, since $1+|s|^2+|a|^2\leq(1+|s|^2)(1+|a|^2)$, and the second is $|z|\o_K(|z|)C(1+|s|^2)$ by \eqref{eq:parametric_log_remainder} and \eqref{eq:polynomial_integration}. Consequently,
\begin{equation}\label{eq:parametric_integrand_expansion}
F_z(s)
=
\left(
\int_\A
A^{\pi^\theta}(s,a)
\nt\pi^\theta(a|s)
\nu(\rd a)
\right)\cdot z
+
r_z(s),
\qquad
|r_z(s)|\leq o(|z|)(1+|s|^8).
\end{equation}
Since the eighth moment of $\rho^{\pi^{\theta+z}}$ is bounded uniformly for $\theta+z\in K$, by \eqref{eq:occupancy_moment} and the compact-uniform constants of \Cref{ass:assumption_pi_theta}, integrating \eqref{eq:parametric_integrand_expansion} gives
\begin{equation}\label{eq:almost_last_characterization}
    J[\pi^{\theta+z}]-J[\pi^\theta]
=
\frac{1}{\beta}\left(
\int_\S\int_\A
A^{\pi^\theta}(s,a)
\nt\pi^\theta(a|s)
\nu(\rd a)
\rho^{\pi^{\theta+z}}(\rd s)
\right)\cdot z
+
o(|z|).
\end{equation}
 
\emph{Step 5: passage to the limit in the occupancy measure.} Consider the vector-valued map
\[
    G(s):=
    \int_\A
    A^{\pi^\theta}(s,a)
    \nt\pi^\theta(a|s)
    \nu(\rd a).
\]
By \eqref{eq:advantage_growth}, \eqref{eq:score_polynomial_growth} and \eqref{eq:polynomial_integration}, $|G(s)|\leq C(1+|s|)^5$, since the integrand has joint degree five. For the continuity of $G$ we argue as in the proof of \Cref{thm:PG}, with $\phi$ replaced by $\nt\pi^\theta=\pi^\theta\nt\log\pi^\theta$. Writing again $A^{\pi^\theta}=\Psi-\lambda\log\pi^\theta$, the term involving $\Psi$ is treated with the decomposition
\[
\begin{aligned}
\int_\A\Psi(s,a)\bigl(\nt\pi^\theta(a|s')-\nt\pi^\theta(a|s)\bigr)\nu(\rd a)&=
\int_\A\Psi(s,a)\bigl(\pi^\theta(a|s')-\pi^\theta(a|s)\bigr)\nt\log\pi^\theta(a|s')\nu(\rd a)\\
&
+
\int_\A\Psi(s,a)\pi^\theta(a|s)\bigl(\nt\log\pi^\theta(a|s')-\nt\log\pi^\theta(a|s)\bigr)\nu(\rd a),
\end{aligned}
\]
whose first term tends to zero by \eqref{eq:policy_weighted_lipschitz}, the integrand having degree at most eight in $a$ for fixed $s,s'$, and whose second term tends to zero by \eqref{eq:score_state_lipschitz} and \eqref{eq:polynomial_integration}; the remaining piece, obtained by replacing $\Psi(s,\cdot)$ by $\Psi(s',\cdot)$, is handled exactly as in the proof of \Cref{thm:PG}. For the entropy term, put
\[
    W(s):=\int_\A\bigl(\pi^\theta\log\pi^\theta\bigr)(a|s)\,\nt\log\pi^\theta(a|s)\nu(\rd a),
\]
and decompose
\[
\begin{aligned}
W(s')-W(s)
={}&
\int_\A
\left[
\bigl(\pi^\theta\log\pi^\theta\bigr)(a|s')
-
\bigl(\pi^\theta\log\pi^\theta\bigr)(a|s)
\right]
\nt\log\pi^\theta(a|s')
\nu(\rd a)\\
&+
\int_\A
\bigl(\pi^\theta\log\pi^\theta\bigr)(a|s)
\left[
\nt\log\pi^\theta(a|s')-\nt\log\pi^\theta(a|s)
\right]
\nu(\rd a).
\end{aligned}
\]
The first term is bounded by $C(1+|s'|^2)L_{\pi^\theta}(1+|s|^8+|s'|^8)|s-s'|$ by \eqref{eq:score_polynomial_growth} and \eqref{eq:policy_entropy_lipschitz}, and the second by $C(1+|s|^2+|s'|^2)|s-s'|(1+|s|^4)$ by \eqref{eq:score_state_lipschitz} and the bound $\|\pi_s\log\pi_s\|_2\leq C(1+|s|)^4$ stated after \eqref{eq:policy_entropy_lipschitz}. Both tend to zero as $s'\to s$, so $G$ is continuous. Therefore \eqref{eq:parametric_occupancy_continuity}, applied componentwise to $G$, gives
\[
\int_\S G(s)\rho^{\pi^{\theta+z}}(\rd s)
=
\int_\S G(s)\rho^{\pi^\theta}(\rd s)
+
o(1).
\]
Multiplying this identity by $z/\beta$ and substituting it into \eqref{eq:almost_last_characterization} yields \eqref{eq:equiv_characta_gradient}. Finally, substituting $\nt\pi^\theta(a|s)=\pi^\theta(a|s)\nt\log\pi^\theta(a|s)$ into \eqref{eq:PG_omega_discounted} gives \eqref{eq:PG_score_discounted}.
\end{proof}

\begin{proof}[\textbf{Proof of \Cref{thm:PG_flow_exponential_convergence}}]
Define
\[
\mathcal E(\tau)
:=
J[\pi^*]-J[\pi^\tau]
+
\frac{\lam}{\beta}
\int_\S
\KL{\pi_s^*}{\pi_s^\tau}
\rho^*(\rd s).
\]By \Cref{rem:relative_entropy_integrability}, $\mathcal E(\tau)$ is finite for every $\tau\geq0$. Applying \Cref{lem:performance_difference} to $\pi^\tau$ and $\pi^*$ gives
\begin{equation}\label{eq:PD_thm}
\mathcal E(\tau)
=
\frac{1}{\beta}\int_\S\int_\A
A^\tau(s,a)
\pi^*(a|s)
\nu(\rd a)
\rho^*(\rd s).
\end{equation}

Since $\pi^\tau$ is strictly positive, \eqref{eq:PG_flow} implies
\begin{equation}\label{eq:PG_flow_log}
\pta\log\pi^\tau(a|s)
=
A^\tau(s,a).
\end{equation}
By \Cref{ass:flow_regularity}, for every $T>0$ there exists $C_T>0$ such that
\begin{equation}\label{eq:uniform_time_score_bound}
\left|
\pta\log\pi^\tau(a|s)
\right|
\leq
C_T(1+|s|+|a|)^3,
\qquad
\tau\in[0,T],\ (s,a)\in\S\times\A.
\end{equation}
We next compute the derivative of $\mathcal E$. By \Cref{ass:flow_regularity}, \Cref{cor:parametric_policy_gradient} applies to the family $\tau\mapsto\pi^\tau$ with $p=1$ (with a right derivative at $\tau=0$) and, together with \eqref{eq:PG_flow}, gives
\begin{equation}\label{eq:dJpi_dtau}
\begin{aligned}
\pta J[\pi^\tau]
&=
\frac{1}{\beta}\int_\S\int_\A
A^\tau(s,a)
\pta\pi^\tau(a|s)
\nu(\rd a)
\rho^\tau(\rd s)=
\frac{1}{\beta}\int_\S\int_\A
|A^\tau(s,a)|^2
\pi^\tau(a|s)
\nu(\rd a)
\rho^\tau(\rd s),
\end{aligned}
\end{equation}
the last integral being finite because $|A^\tau|^2$ has joint degree six, so that \eqref{eq:polynomial_integration} and \eqref{eq:occupancy_moment} apply. Moreover, since $\pi^*$ and $\rho^*$ do not depend on $\tau$, differentiation under the integral sign and \eqref{eq:PG_flow_log} give
\begin{equation}\label{eq:dKL_dtau}
\begin{aligned}
\pta
\int_\S
\KL{\pi_s^*}{\pi_s^\tau}
\rho^*(\rd s)
&=
-
\int_\S\int_\A
A^\tau(s,a)
\pi^*(a|s)
\nu(\rd a)
\rho^*(\rd s)=
-\beta\,\mathcal E(\tau),
\end{aligned}
\end{equation}
where the last equality follows from \eqref{eq:PD_thm}. Differentiation under the integral sign is justified on every interval $[0,T]$ by \eqref{eq:PG_flow_log} and \eqref{eq:uniform_time_score_bound}, since
\[
    \left| \pta\log\pi^\tau(a|s) \right| \pi^*(a|s)
    =
    |A^\tau(s,a)| \pi^*(a|s)
    \leq
    C_T(1+|s|+|a|)^3\pi^*(a|s),
\]
and the right-hand side is integrable with respect to $\nu(\rd a)\rho^*(\rd s)$. Combining \eqref{eq:dJpi_dtau} and \eqref{eq:dKL_dtau}, we obtain
\[
\begin{aligned}
\pta\mathcal E(\tau)
&=
-\pta J[\pi^\tau]
+
\frac{\lam}{\beta}
\pta
\int_\S
\KL{\pi_s^*}{\pi_s^\tau}
\rho^*(\rd s)=
-
\frac{1}{\beta}\int_\S\int_\A
|A^\tau(s,a)|^2
\pi^\tau(a|s)
\nu(\rd a)
\rho^\tau(\rd s)
-
\lam\mathcal E(\tau)
\leq
-\lam\mathcal E(\tau).
\end{aligned}
\]
Gronwall's inequality gives $\mathcal E(\tau)\leq e^{-\lam\tau}\mathcal E(0)$, which is precisely \eqref{eq:exponential_decay}.
\end{proof}

\begin{proof}[\textbf{Proof of \Cref{thm:discrete_parametrized_policy_optimization}}]
Throughout the proof, $C>0$ denotes a constant that may change from line to line but is independent of $n$, $\dta$, $s$, $a$ and of the interpolation parameter $u\in[0,1]$ introduced below. For every $n\geq0$ set
\[
\ve^{(n)}:=\ve_A^{(n)}+\veb^{(n)},
\qquad
q^{(n)}(s,a):=w^{(n)\top}\nt\log\pi^{(n)}(a|s).
\]
By the definitions of the two errors,
\begin{equation}\label{eq:dpo_q_decomposition}
q^{(n)}(s,a)=A^{(n)}(s,a)+\ve^{(n)}(s,a).
\end{equation}

\emph{Preliminary bounds.} We collect the estimates used below. By \Cref{assump:discrete_policy_optimization}(1), all parameters considered below lie in the compact set $\Theta_0$. By \eqref{eq:score_polynomial_growth} with $K=\Theta_0$ and by \Cref{assump:discrete_policy_optimization}(1), for every $\vartheta\in\Theta_0$,
\begin{equation}\label{eq:dpo_score_bounds}
|\nt\log\pi^\vartheta(a|s)|\leq C(1+|s|+|a|)^2,
\qquad
\|\nt^2\log\pi^\vartheta(a|s)\|\leq C(1+|s|+|a|)^2,
\end{equation}
where we used $1+|s|^2+|a|^2\leq(1+|s|+|a|)^2$. By \eqref{eq:advantage_growth}, with a constant uniform on $\Theta_0$, and by \Cref{assump:discrete_policy_optimization}(2),
\begin{equation}\label{eq:dpo_cubic_growth}
|A^{(n)}(s,a)|+|\ve^{(n)}(s,a)|\leq C(1+|s|+|a|)^3 .
\end{equation}
By \Cref{ass:assumption_pi_theta}, the constants in \eqref{eq:polynomial_integration} and \eqref{eq:occupancy_moment} are uniform over $\Theta_0$. Hence, for $k\leq8$ and every $\vartheta\in\Theta_0$,
\begin{equation}\label{eq:dpo_moments}
\int_\A(1+|s|+|a|)^k\pi^\vartheta(a|s)\nu(\rd a)\leq C(1+|s|)^k,
\qquad
\int_\S(1+|s|)^k\rho^{(n+1)}(\rd s)+\int_\S(1+|s|)^k\rho^*(\rd s)\leq C,
\end{equation}
and the first bound also holds with $\pi^\vartheta$ replaced by $\pi^*\in\Pi$. Finally, $\int_\A A^{(n)}\pi^{(n)}\nu(\rd a)=0$ by \eqref{eq:centered_advantage}, and $\int_\A\nt\log\pi^{(n)}\pi^{(n)}\nu(\rd a)=0$ by \eqref{eq:score_centering_parametric}. Multiplying the latter by $w^{(n)}$ and using \eqref{eq:dpo_q_decomposition},
\begin{equation}\label{eq:dpo_centering}
\int_\A A^{(n)}(s,a)\pi^{(n)}(a|s)\nu(\rd a)
=\int_\A q^{(n)}(s,a)\pi^{(n)}(a|s)\nu(\rd a)
=\int_\A \ve^{(n)}(s,a)\pi^{(n)}(a|s)\nu(\rd a)=0 .
\end{equation}
All the integrals written below are finite by \eqref{eq:dpo_score_bounds}--\eqref{eq:dpo_moments} and \Cref{rem:relative_entropy_integrability}.

\medskip
\emph{Step 1: score variation.}
Fix $n\geq0$ and $(s,a)\in\S\times\A$. For $u\in[0,1]$ define the segment
\[
\vartheta_u:=\theta^{(n)}+u\,\dta\,w^{(n)} .
\]
By \eqref{eq:theta_update}, $\vartheta_0=\theta^{(n)}$ and $\vartheta_1=\theta^{(n+1)}$, and by \Cref{assump:discrete_policy_optimization}(1), $\vartheta_u\in\Theta_0$ for all $u\in[0,1]$. Set
\[
Z_n(s,a):=\log\pi^{(n)}(a|s),
\qquad
\varphi(u):=\log\pi^{\vartheta_u}(a|s),\quad u\in[0,1],
\]
so that $\varphi(0)=Z_n(s,a)$ and $\varphi(1)=Z_{n+1}(s,a)$. Since $\frac{\rd}{\rd u}\vartheta_u=\dta\,w^{(n)}$, the chain rule gives
\[
\varphi'(u)=\dta\,w^{(n)\top}\nt\log\pi^{\vartheta_u}(a|s),
\qquad
\varphi''(u)=\dta^2\,w^{(n)\top}\nt^2\log\pi^{\vartheta_u}(a|s)\,w^{(n)} .
\]
In particular, $\varphi'(0)=\dta\,q^{(n)}(s,a)$. Integrating by parts,
$\int_0^1(1-u)\varphi''(u)\,\rd u=-\varphi'(0)+\varphi(1)-\varphi(0)$, which is Taylor's formula with integral remainder:
\[
\varphi(1)-\varphi(0)=\varphi'(0)+\int_0^1(1-u)\varphi''(u)\,\rd u .
\]
Together with \eqref{eq:dpo_q_decomposition}, this yields the exact identity
\begin{equation}\label{eq:dpo_log_expansion}
Z_{n+1}(s,a)-Z_n(s,a)=\dta\left(A^{(n)}(s,a)+\ve^{(n)}(s,a)\right)+R^{(n)}(s,a),
\end{equation}
\[
R^{(n)}(s,a):=\dta^2\int_0^1(1-u)\,w^{(n)\top}\nt^2\log\pi^{\vartheta_u}(a|s)\,w^{(n)}\,\rd u .
\]
By \eqref{eq:dpo_score_bounds} and $\int_0^1(1-u)\rd u=\frac12$,
\begin{equation}\label{eq:dpo_R_bound}
|R^{(n)}(s,a)|\leq C\,\dta^2|w^{(n)}|^2(1+|s|+|a|)^2 .
\end{equation}
Moreover, writing $Z_{n+1}-Z_n=\varphi(1)-\varphi(0)=\int_0^1\varphi'(u)\,\rd u$ and using \eqref{eq:dpo_score_bounds},
\begin{equation}\label{eq:dpo_log_increment}
|Z_{n+1}(s,a)-Z_n(s,a)|\leq C\,\dta\,|w^{(n)}|\,(1+|s|+|a|)^2 .
\end{equation}
Now multiply \eqref{eq:dpo_log_expansion} by $\pi^{(n)}(a|s)$ and integrate in $a$. On the left-hand side,
\[
\int_\A\left(Z_{n+1}-Z_n\right)\pi^{(n)}\nu(\rd a)
=\int_\A\pi^{(n)}(a|s)\log\frac{\pi^{(n+1)}(a|s)}{\pi^{(n)}(a|s)}\nu(\rd a)
=-\KL{\pi^{(n)}_s}{\pi^{(n+1)}_s}.
\]
On the right-hand side, the first-order term vanishes by \eqref{eq:dpo_centering}. Therefore
\begin{equation}\label{eq:dpo_score_variation}
\int_\A\left(Z_{n+1}(s,a)-Z_n(s,a)\right)\pi^{(n)}(a|s)\nu(\rd a)
=\int_\A R^{(n)}(s,a)\pi^{(n)}(a|s)\nu(\rd a)
=-\KL{\pi^{(n)}_s}{\pi^{(n+1)}_s}.
\end{equation}

\medskip
\emph{Step 2: variation of $\mathcal D_n$ and $\mathcal E_n$.}
Define
\[
\mathcal D_n:=\int_\S\KL{\pi^*_s}{\pi^{(n)}_s}\rho^*(\rd s),
\qquad
\mathcal E_n:=J[\pi^*]-J[\pi^{(n)}]+\frac{\lam}{\beta}\mathcal D_n ,
\]
which are finite by \Cref{rem:relative_entropy_integrability}. Applying \Cref{lem:performance_difference} with $\widehat\pi=\pi^*$ and $\pi=\pi^{(n)}$, so that $\rho^{\widehat\pi}=\rho^*$,
\[
J[\pi^*]-J[\pi^{(n)}]
=\frac1\beta\int_\S\int_\A A^{(n)}\pi^*\nu(\rd a)\rho^*(\rd s)-\frac{\lam}{\beta}\int_\S\KL{\pi^*_s}{\pi^{(n)}_s}\rho^*(\rd s).
\]
The last term is exactly $-\frac{\lam}{\beta}\mathcal D_n$: the relative entropy appears in the same direction and is integrated against the same measure. Adding $\frac\lam\beta\mathcal D_n$ to both sides gives
\begin{equation}\label{eq:dpo_energy_identity}
\mathcal E_n=\frac1\beta\int_\S\int_\A A^{(n)}(s,a)\pi^*(a|s)\nu(\rd a)\rho^*(\rd s).
\end{equation}
By the definition of $\mathcal E_n$,
\begin{equation}\label{eq:dpo_E_difference}
\mathcal E_{n+1}-\mathcal E_n=-\left(J[\pi^{(n+1)}]-J[\pi^{(n)}]\right)+\frac{\lam}{\beta}\left(\mathcal D_{n+1}-\mathcal D_n\right).
\end{equation}
We compute the variation of the relative entropy for fixed $s$. Since $\KL{\pi^*_s}{\pi^{(n)}_s}=\int_\A\pi^*\log\pi^*\,\nu(\rd a)-\int_\A\pi^* Z_n\,\nu(\rd a)$, adding and subtracting $\int_\A(Z_{n+1}-Z_n)\pi^{(n)}\nu(\rd a)$ and using \eqref{eq:dpo_score_variation},
\begin{equation}\label{eq:dpo_three_point}
\begin{aligned}
\KL{\pi^*_s}{\pi^{(n+1)}_s}-\KL{\pi^*_s}{\pi^{(n)}_s}
&=-\int_\A\left(Z_{n+1}-Z_n\right)\pi^*\,\nu(\rd a)\\
&=-\int_\A\left(Z_{n+1}-Z_n\right)\left(\pi^*-\pi^{(n)}\right)\nu(\rd a)-\int_\A\left(Z_{n+1}-Z_n\right)\pi^{(n)}\,\nu(\rd a)\\
&=-\int_\A\left(Z_{n+1}-Z_n\right)\left(\pi^*-\pi^{(n)}\right)\nu(\rd a)+\KL{\pi^{(n)}_s}{\pi^{(n+1)}_s}.
\end{aligned}
\end{equation}
We now substitute \eqref{eq:dpo_log_expansion} into \eqref{eq:dpo_three_point} and integrate against $\rho^*$. By \eqref{eq:dpo_centering}, $\int_\A(A^{(n)}+\ve^{(n)})\pi^{(n)}\nu(\rd a)=0$. Then \eqref{eq:dpo_energy_identity} gives
\begin{equation}\label{eq:dpo_D_difference}
\begin{aligned}
\mathcal D_{n+1}-\mathcal D_n
&=-\dta\int_\S\int_\A\left(A^{(n)}+\ve^{(n)}\right)\left(\pi^*-\pi^{(n)}\right)\nu(\rd a)\rho^*(\rd s)\\
&\quad-\int_\S\int_\A R^{(n)}\left(\pi^*-\pi^{(n)}\right)\nu(\rd a)\rho^*(\rd s)
+\int_\S\KL{\pi^{(n)}_s}{\pi^{(n+1)}_s}\rho^*(\rd s)\\
&=-\dta\int_\S\int_\A A^{(n)}\pi^*\,\nu(\rd a)\rho^*(\rd s)
-\dta\int_\S\int_\A\ve^{(n)}\pi^*\,\nu(\rd a)\rho^*(\rd s)\\
&\quad-\int_\S\int_\A R^{(n)}\left(\pi^*-\pi^{(n)}\right)\nu(\rd a)\rho^*(\rd s)
+\int_\S\KL{\pi^{(n)}_s}{\pi^{(n+1)}_s}\rho^*(\rd s)\\
&=-\beta\dta\,\mathcal E_n
-\dta\int_\S\int_\A\ve^{(n)}\pi^*\,\nu(\rd a)\rho^*(\rd s)\\
&\quad-\int_\S\int_\A R^{(n)}\left(\pi^*-\pi^{(n)}\right)\nu(\rd a)\rho^*(\rd s)
+\int_\S\KL{\pi^{(n)}_s}{\pi^{(n+1)}_s}\rho^*(\rd s).
\end{aligned}
\end{equation}

\medskip
\emph{Step 3: expression of $\mathcal E_{n+1}$.}
With $\vartheta_u$ as in Step 1, define $\psi(u):=\pi^{\vartheta_u}(a|s)$ for $u\in[0,1]$, so that $\psi(0)=\pi^{(n)}(a|s)$ and $\psi(1)=\pi^{(n+1)}(a|s)$. Using $\nt\pi^\vartheta=\pi^\vartheta\nt\log\pi^\vartheta$ and, differentiating once more,
\begin{equation}\label{eq:dpo_hessian_density}
\nt^2\pi^\vartheta=\pi^\vartheta\left(\nt^2\log\pi^\vartheta+\nt\log\pi^\vartheta\,\nt\log\pi^{\vartheta\top}\right),
\end{equation}
the chain rule gives
\begin{equation}\label{eq:dpo_psi_derivatives}
\psi'(u)=\dta\left(w^{(n)\top}\nt\log\pi^{\vartheta_u}(a|s)\right)\psi(u),
\qquad
\psi''(u)=\dta^2\,w^{(n)\top}\nt^2\pi^{\vartheta_u}(a|s)\,w^{(n)} ,
\end{equation}
and in particular $\psi'(0)=\dta\,q^{(n)}(s,a)\pi^{(n)}(a|s)$. Taylor's formula with integral remainder, as in Step 1, yields
\begin{equation}\label{eq:dpo_density_expansion}
\pi^{(n+1)}(a|s)=\pi^{(n)}(a|s)+\dta\,q^{(n)}(s,a)\pi^{(n)}(a|s)+\widetilde R^{(n)}(a|s),
\qquad
\widetilde R^{(n)}(a|s):=\int_0^1(1-u)\psi''(u)\,\rd u .
\end{equation}
Multiplying \eqref{eq:dpo_density_expansion} by $A^{(n)}(s,a)$, integrating in $a$, and using $\int_\A A^{(n)}\pi^{(n)}\nu(\rd a)=0$,
\begin{equation}\label{eq:dpo_A_next}
\int_\A A^{(n)}(s,a)\pi^{(n+1)}(a|s)\nu(\rd a)
=\dta\int_\A A^{(n)}q^{(n)}\pi^{(n)}\nu(\rd a)+R_A^{(n)}(s),
\qquad
R_A^{(n)}(s):=\int_\A A^{(n)}(s,a)\widetilde R^{(n)}(a|s)\nu(\rd a).
\end{equation}
Applying \Cref{lem:performance_difference} with $\widehat\pi=\pi^{(n+1)}$ and $\pi=\pi^{(n)}$, and using \eqref{eq:dpo_A_next},
\begin{equation}\label{eq:dpo_J_difference}
\begin{aligned}
&J[\pi^{(n+1)}]-J[\pi^{(n)}]
=\frac1\beta\int_\S\left(\int_\A A^{(n)}\pi^{(n+1)}\nu(\rd a)-\lam\KL{\pi^{(n+1)}_s}{\pi^{(n)}_s}\right)\rho^{(n+1)}(\rd s)\\
&=\frac{\dta}{\beta}\int_\S\int_\A A^{(n)}q^{(n)}\pi^{(n)}\nu(\rd a)\rho^{(n+1)}(\rd s)
+\frac1\beta\int_\S R_A^{(n)}\rho^{(n+1)}(\rd s)
-\frac{\lam}{\beta}\int_\S\KL{\pi^{(n+1)}_s}{\pi^{(n)}_s}\rho^{(n+1)}(\rd s).
\end{aligned}
\end{equation}
Substituting \eqref{eq:dpo_D_difference} and \eqref{eq:dpo_J_difference} into \eqref{eq:dpo_E_difference}, we obtain the exact identity
\begin{equation}\label{eq:dpo_E_next}
\mathcal E_{n+1}=(1-\lam\dta)\,\mathcal E_n+I_1+I_2+T_n,
\end{equation}
where
\begin{align*}
I_1&:=-\frac{\lam\dta}{\beta}\int_\S\int_\A\ve^{(n)}(s,a)\pi^*(a|s)\nu(\rd a)\rho^*(\rd s),\\
I_2&:=-\frac{\dta}{\beta}\int_\S\int_\A A^{(n)}(s,a)q^{(n)}(s,a)\pi^{(n)}(a|s)\nu(\rd a)\rho^{(n+1)}(\rd s),\\
T_n&:=\frac{\lam}{\beta}\int_\S\KL{\pi^{(n+1)}_s}{\pi^{(n)}_s}\rho^{(n+1)}(\rd s)
+\frac{\lam}{\beta}\int_\S\KL{\pi^{(n)}_s}{\pi^{(n+1)}_s}\rho^*(\rd s)\\
&\quad-\frac1\beta\int_\S R_A^{(n)}(s)\rho^{(n+1)}(\rd s)
-\frac{\lam}{\beta}\int_\S\int_\A R^{(n)}(s,a)\left(\pi^*(a|s)-\pi^{(n)}(a|s)\right)\nu(\rd a)\rho^*(\rd s).
\end{align*}
The term $I_1$ collects the errors under the optimal policy, $I_2$ is the improvement of the objective, and $T_n$ collects all second-order terms produced by the discretization.

\medskip
\emph{Step 4: estimates of $I_1$, $I_2$ and $T_n$.}

\emph{Estimate of $I_1$.} Splitting $\ve^{(n)}=\ve_A^{(n)}+\veb^{(n)}$ and using \Cref{assump:discrete_policy_optimization}(3),
\begin{equation}\label{eq:dpo_I1}
I_1
\leq\frac{\lam\dta}{\beta}\left(\left|\int_\S\int_\A\ve_A^{(n)}\pi^*\nu(\rd a)\rho^*(\rd s)\right|+\left|\int_\S\int_\A\veb^{(n)}\pi^*\nu(\rd a)\rho^*(\rd s)\right|\right)
\leq\frac{2\lam\dta\,\ve_1}{\beta}.
\end{equation}

\emph{Estimate of $I_2$.} In this estimate, all double integrals are taken with respect to $\pi^{(n)}(a|s)\nu(\rd a)\rho^{(n+1)}(\rd s)$. By \eqref{eq:dpo_q_decomposition}, $A^{(n)}=q^{(n)}-\ve^{(n)}$. Using the Cauchy--Schwarz inequality and $xy\leq\frac12x^2+\frac12y^2$,
\begin{align*}
\int_\S\int_\A A^{(n)}q^{(n)}
=\int_\S\int_\A|q^{(n)}|^2-\int_\S\int_\A\ve^{(n)}q^{(n)}
&\geq\int_\S\int_\A|q^{(n)}|^2-\left(\int_\S\int_\A|\ve^{(n)}|^2\right)^{1/2}\left(\int_\S\int_\A|q^{(n)}|^2\right)^{1/2}\\
&
\geq\frac12\int_\S\int_\A|q^{(n)}|^2-\frac12\int_\S\int_\A|\ve^{(n)}|^2 .
\end{align*}
Since $|\ve^{(n)}|^2\leq2|\ve_A^{(n)}|^2+2|\veb^{(n)}|^2$, \Cref{assump:discrete_policy_optimization}(4) gives $\int_\S\int_\A|\ve^{(n)}|^2\leq2\ve_2$. Hence
\begin{equation}\label{eq:dpo_I2}
I_2\leq-\frac{\dta}{2\beta}\int_\S\int_\A|q^{(n)}(s,a)|^2\pi^{(n)}(a|s)\nu(\rd a)\rho^{(n+1)}(\rd s)+\frac{\dta\,\ve_2}{\beta}.
\end{equation}
Moreover, since $|q^{(n)}|^2=w^{(n)\top}\big(\nt\log\pi^{(n)}\,\nt\log\pi^{(n)\top}\big)w^{(n)}$ and $w^{(n)}$ does not depend on $(s,a)$,
\begin{equation}\label{eq:dpo_fisher_identity}
\int_\S\int_\A|q^{(n)}(s,a)|^2\pi^{(n)}(a|s)\nu(\rd a)\rho^{(n+1)}(\rd s)=w^{(n)\top}F_n\,w^{(n)} .
\end{equation}

\emph{Estimate of $T_n$.} We bound the three types of terms in $T_n$ separately.

(i) \emph{The two relative entropies.} Fix $s$ and abbreviate $\psi_u:=\pi^{\vartheta_u}(\cdot|s)$, so that $\psi_0=\pi^{(n)}_s$ and $\psi_1=\pi^{(n+1)}_s$. Let $L:=Z_{n+1}(s,\cdot)-Z_n(s,\cdot)=\log\psi_1-\log\psi_0$. Then
\[
\int_\A(\psi_1-\psi_0)\,L\,\nu(\rd a)
=\int_\A\psi_1\log\frac{\psi_1}{\psi_0}\nu(\rd a)-\int_\A\psi_0\log\frac{\psi_1}{\psi_0}\nu(\rd a)
=\KL{\pi^{(n+1)}_s}{\pi^{(n)}_s}+\KL{\pi^{(n)}_s}{\pi^{(n+1)}_s}.
\]
Both relative entropies are nonnegative. Hence each of them is bounded by
\begin{equation}\label{eq:dpo_KL_symmetric}
\KL{\pi^{(n+1)}_s}{\pi^{(n)}_s}+\KL{\pi^{(n)}_s}{\pi^{(n+1)}_s}
\leq\int_\A|\psi_1-\psi_0|\,|L|\,\nu(\rd a).
\end{equation}
Both factors on the right-hand side are of first order in $\dta|w^{(n)}|$. For $L$ this is \eqref{eq:dpo_log_increment}. For the densities, \eqref{eq:dpo_psi_derivatives} and \eqref{eq:dpo_score_bounds} give
\[
|\psi_1-\psi_0|=\left|\int_0^1\psi'(u)\,\rd u\right|\leq C\,\dta\,|w^{(n)}|(1+|s|+|a|)^2\int_0^1\psi_u\,\rd u .
\]
Inserting both bounds into \eqref{eq:dpo_KL_symmetric}, exchanging the order of integration by Tonelli's theorem, and using \eqref{eq:dpo_moments} with $k=4$,
\[
\KL{\pi^{(n+1)}_s}{\pi^{(n)}_s}+\KL{\pi^{(n)}_s}{\pi^{(n+1)}_s}
\leq C\,\dta^2|w^{(n)}|^2\int_0^1\int_\A(1+|s|+|a|)^4\psi_u\,\nu(\rd a)\,\rd u
\leq C\,\dta^2|w^{(n)}|^2(1+|s|)^4 .
\]
Integrating against $\rho^{(n+1)}$ and $\rho^*$ and using \eqref{eq:dpo_moments},
\begin{equation}\label{eq:dpo_T_KL}
\frac{\lam}{\beta}\int_\S\KL{\pi^{(n+1)}_s}{\pi^{(n)}_s}\rho^{(n+1)}(\rd s)
+\frac{\lam}{\beta}\int_\S\KL{\pi^{(n)}_s}{\pi^{(n+1)}_s}\rho^*(\rd s)
\leq C\,\dta^2|w^{(n)}|^2 .
\end{equation}

(ii) \emph{The term $R_A^{(n)}$.} By \eqref{eq:dpo_hessian_density}, \eqref{eq:dpo_psi_derivatives} and \eqref{eq:dpo_score_bounds}, $|\psi''(u)|\leq C\,\dta^2|w^{(n)}|^2(1+|s|+|a|)^4\psi_u(a)$, and therefore
\[
|\widetilde R^{(n)}(a|s)|\leq C\,\dta^2|w^{(n)}|^2(1+|s|+|a|)^4\int_0^1\psi_u(a)\,\rd u .
\]
Combining this with \eqref{eq:dpo_cubic_growth}, Tonelli's theorem and \eqref{eq:dpo_moments} with $k=7$,
\[
|R_A^{(n)}(s)|\leq C\,\dta^2|w^{(n)}|^2\int_0^1\int_\A(1+|s|+|a|)^7\psi_u\,\nu(\rd a)\,\rd u\leq C\,\dta^2|w^{(n)}|^2(1+|s|)^7 ,
\]
and integrating against $\rho^{(n+1)}$,
\begin{equation}\label{eq:dpo_T_RA}
\frac1\beta\int_\S|R_A^{(n)}(s)|\rho^{(n+1)}(\rd s)\leq C\,\dta^2|w^{(n)}|^2 .
\end{equation}

(iii) \emph{The term $R^{(n)}$.} By \eqref{eq:dpo_R_bound} and \eqref{eq:dpo_moments} with $k=2$, applied to $\pi^*$ and $\pi^{(n)}$,
\[
\int_\A|R^{(n)}(s,a)|\left(\pi^*(a|s)+\pi^{(n)}(a|s)\right)\nu(\rd a)\leq C\,\dta^2|w^{(n)}|^2(1+|s|)^2 ,
\]
and integrating against $\rho^*$,
\begin{equation}\label{eq:dpo_T_R}
\frac{\lam}{\beta}\left|\int_\S\int_\A R^{(n)}\left(\pi^*-\pi^{(n)}\right)\nu(\rd a)\rho^*(\rd s)\right|\leq C\,\dta^2|w^{(n)}|^2 .
\end{equation}

Combining \eqref{eq:dpo_T_KL}, \eqref{eq:dpo_T_RA} and \eqref{eq:dpo_T_R}, there exists $C_*>0$, independent of $n$ and $\dta$, such that
\begin{equation}\label{eq:dpo_T}
T_n\leq|T_n|\leq C_*\,\dta^2\,|w^{(n)}|^2 .
\end{equation}
By \Cref{assump:discrete_policy_optimization}(5) and \eqref{eq:dpo_fisher_identity},
\begin{equation}\label{eq:dpo_T_fisher}
T_n\leq\frac{C_*\dta^2}{\mu}\,w^{(n)\top}F_n\,w^{(n)}
=\frac{C_*\dta^2}{\mu}\int_\S\int_\A|q^{(n)}(s,a)|^2\pi^{(n)}(a|s)\nu(\rd a)\rho^{(n+1)}(\rd s).
\end{equation}

\medskip
\emph{Step 5: rolling out and conclusion.}
Substituting \eqref{eq:dpo_I1}, \eqref{eq:dpo_I2} and \eqref{eq:dpo_T_fisher} into \eqref{eq:dpo_E_next},
\[
\mathcal E_{n+1}\leq(1-\lam\dta)\,\mathcal E_n+\frac{2\lam\dta\,\ve_1}{\beta}+\frac{\dta\,\ve_2}{\beta}
+\dta\left(\frac{C_*\dta}{\mu}-\frac{1}{2\beta}\right)\int_\S\int_\A|q^{(n)}|^2\pi^{(n)}\nu(\rd a)\rho^{(n+1)}(\rd s).
\]
Set $\dta_0:=\mu/(2\beta C_*)$, which is independent of $n$ and $\dta$. For $0<\dta\leq\dta_0$ the bracket is nonpositive, and since the integral is nonnegative, the last term can be dropped:
\begin{equation}\label{eq:dpo_recursion}
\mathcal E_{n+1}\leq(1-\lam\dta)\,\mathcal E_n+\dta\,\frac{2\lam\ve_1+\ve_2}{\beta}.
\end{equation}
Since $0<\lam\dta\leq1$, we have $0\leq1-\lam\dta<1$, so \eqref{eq:dpo_recursion} can be multiplied by powers of $1-\lam\dta$ without changing the direction of the inequality. By induction on $n$,
\[
\mathcal E_n\leq(1-\lam\dta)^n\mathcal E_0+\dta\,\frac{2\lam\ve_1+\ve_2}{\beta}\sum_{k=0}^{n-1}(1-\lam\dta)^k .
\]
Since $\sum_{k=0}^{n-1}(1-\lam\dta)^k=\frac{1-(1-\lam\dta)^n}{\lam\dta}\leq\frac{1}{\lam\dta}$, we conclude that $\mathcal E_n\leq(1-\lam\dta)^n\mathcal E_0+\frac{2\ve_1}{\beta}+\frac{\ve_2}{\lam\beta}.$ Substituting the definition of $\mathcal E_n$ proves \eqref{eq:main_result}.
\end{proof}

\begin{remark}[Without the Fisher condition]\label{rem:no_fisher}
\Cref{assump:discrete_policy_optimization}(5) is used only once in the proof, namely in \eqref{eq:dpo_T_fisher}, to absorb the discretization term $T_n$ into the improvement term $I_2$. If \Cref{assump:discrete_policy_optimization}(5) does not hold, the same argument still applies with two changes. First, in place of \eqref{eq:dpo_T_fisher}, we combine \eqref{eq:dpo_T} with the bound $|w^{(n)}|\leq K$ of \Cref{assump:discrete_policy_optimization}(2), which gives $T_n\leq C_*K^2\dta^2$. Second, in \eqref{eq:dpo_I2} the nonpositive term $-\frac{\dta}{2\beta}\int_\S\int_\A|q^{(n)}|^2\pi^{(n)}\nu(\rd a)\rho^{(n+1)}(\rd s)$ is simply discarded, which gives $I_2\leq\dta\,\ve_2/\beta$. Inserting these two bounds and \eqref{eq:dpo_I1} into \eqref{eq:dpo_E_next} yields, in place of \eqref{eq:dpo_recursion},
\[
\mathcal E_{n+1}\leq(1-\lam\dta)\,\mathcal E_n+\dta\,\frac{2\lam\ve_1+\ve_2}{\beta}+C_*K^2\dta^2 .
\]
Rolling out as in Step 5, we obtain, for every $\dta>0$ with $\lam\dta\leq1$,
\begin{align}\label{eq:estimatio_with_singular_fisher}
    \mathcal E_n\leq(1-\lam\dta)^n\mathcal E_0+\frac{2\ve_1}{\beta}+\frac{\ve_2}{\lam\beta}+\frac{C_*K^2\dta}{\lam}.
\end{align}
\end{remark}

\color{black}

\subsection{Additional comments on linear approximation and optimization}\label{subsec:policy_eval}
\subsubsection{Linear approximation in policy evaluation}
The policy evaluation step is based on the PhiBE equation
\[
    \beta \hat V^{\pi_{\th_k}}(s)=r_\lambda^{\pi_{\th_k}}(s)+\L_{\hat b^{\pi_{\th_k}},\hat\Sigma^{\pi_{\th_k}}}[\hat V^{\pi_{\th_k}}](s).
\]
To solve this policy evaluation equation from data, we use a Galerkin approximation of the PhiBE PDE. Namely, we choose a finite-dimensional approximation space generated by smooth basis functions $\Phi_1,\ldots,\Phi_m$ and seek the value function in the form
\[
    \hat V_\xi^{\pi_{\th_k}}(s)=\xi^\top\Phi(s), \qquad \Phi=(\Phi_1,\ldots,\Phi_m)^\top,\qquad \xi\in\R^m.
\]
Thus, the infinite-dimensional policy evaluation problem is reduced to the computation of the coefficient vector $\xi$. The coefficients are determined by imposing the PhiBE equation in an empirical Galerkin sense, using the samples. For each data point, set $\Delta s_i:=s_i'-s_i$ and define the empirical PhiBE generator acting on each basis function by
\[
    \widehat{\mathcal G}_i\Phi_j:=\frac{\Delta s_i}{\dt}\cdot\nb_s\Phi_j(s_i)+\frac{1}{2\dt}\Delta s_i^\top\nb_s^2\Phi_j(s_i)\Delta s_i,\qquad j=1,\ldots,m.
\]
We also write $\widehat{\mathcal G}_i\Phi:=(\widehat{\mathcal G}_i\Phi_1,\ldots,\widehat{\mathcal G}_i\Phi_m)^\top$. The empirical Galerkin approximation of the PhiBE policy evaluation equation is defined by
\[
    \frac1n\sum_{i=1}^n \Phi(s_i)\left(\beta \hat V_\xi^{\pi_{\th_k}}(s_i)-\widehat{\mathcal G}_i\hat V_\xi^{\pi_{\th_k}}-\bigl(r_i-\lambda\log\pi_{\th_k}(a_i|s_i)\bigr)\right)=0.
\]
Since $\widehat{\mathcal G}_i\hat V_\xi^{\pi_{\th_k}}=\xi^\top\widehat{\mathcal G}_i\Phi$, this condition is equivalent to the linear system $F_k\xi_k=b_k,$
where
\[
    F_k:=\frac1n\sum_{i=1}^n \Phi(s_i)\left(\beta\Phi(s_i)-\widehat{\mathcal G}_i\Phi\right)^\top\quad\text{and}\quad
    b_k:=\frac1n\sum_{i=1}^n \left(r_i-\lambda\log\pi_{\th_k}(a_i|s_i)\right)\Phi(s_i).
\]
When $F_k$ is invertible, we set $\xi_k=F_k^{-1}b_k$. If $F_k$ is singular or ill-conditioned, one may use a regularized Galerkin system or the Moore--Penrose inverse. This gives the approximate value function $\hat V^{\pi_{\th_k}}(s):=\xi_k^\top\Phi(s).$
\smallbreak

\subsubsection{Policy gradient optimization methods}
\label{subsec:gradiet_based_opt}

For observations $(s_i,a_i)$ collected at iteration $k$, let $\widehat w_k$ be the empirical policy-gradient direction defined in \eqref{eq:on-policy_empirical_gradient}. Vanilla policy gradient uses \eqref{eq:vanilla_direction} with $w_k=\widehat w_k$. For natural policy gradient, define the empirical Fisher information matrix
\begin{equation*}
\widehat{\mathcal F}_k:=\frac{1}{n}\sum_{i=1}^n\nb_\th\log\pi_{\th_k}(a_i|s_i)\nb_\th\log\pi_{\th_k}(a_i|s_i)^\top.
\end{equation*}
The natural-gradient direction is $g_k=\widehat{\mathcal F}_k^\dagger\widehat w_k$, where $\widehat{\mathcal F}_k^\dagger$ denotes the Moore--Penrose inverse. The update follows \eqref{eq:vanilla_direction} with $w_k=g_k$. For TRPO, the direction is obtained from
\begin{equation*}
g_k\in\operatorname*{argmax}_{g}\ \widehat w_k\cdot g\qquad\text{subject to}\qquad \frac{\eta^2}{2}g^\top\widehat{\mathcal F}_kg\leq\delta_{\rm TRPO},
\end{equation*}
and \eqref{eq:vanilla_direction} is applied with $w_k=g_k$; see \cite{schulman2015trpo}. For PPO, let $\varrho_\th(s,a):=\frac{\pi_\th(a|s)}{\pi_{\th_k}(a|s)}$. The empirical clipped surrogate objective is
\begin{align}\label{eq:PPO_direction}
\widehat L_{\rm PPO}(\th;\theta_k):=\frac{1}{n}\sum_{i=1}^n\min\Bigl\{&\varrho_\th(s_i,a_i)\widehat A^{\pi_{\th_k}}(s_i,a_i),\text{\rm clip}\bigl(\varrho_\th(s_i,a_i),1-\epsilon_{\rm PPO},1+\epsilon_{\rm PPO}\bigr)\widehat A^{\pi_{\th_k}}(s_i,a_i)\Bigr\},
\end{align}
where $\text{\rm clip}(x,l,u)=\max(l,\min(x,u))$ and $\epsilon_{\rm PPO}>0$. 
For PPO, starting from $\th=\th_k$, we perform several gradient-ascent steps $\th\leftarrow\th+\eta\nb_\th\widehat L_{\rm PPO}(\th;\theta_k),
$ and set $\th_{k+1}=\th$ after the final step.

\subsection{Additional details on the numerical experiments}\label{sec:extra_num_ex}
This section provides additional details on the numerical examples presented in \Cref{sec:numerical_results}.

\subsubsection{LQR setting}\label{subsec:lqr_setting}

The coefficients $k_2^\omega$, $k_1^\omega$, and $k_0^\omega$ of $V^\omega$ are explicitly given by
\begin{align*}
    k_2^\omega
    &=-\frac{M+N\omega_1^2}{\beta-2(A+B\omega_1)},\qquad
    k_1^\omega
    =\frac{B\omega_2k_2^\omega-N\omega_1\omega_2}
        {\beta-A-B\omega_1},\\
    k_0^\omega
    &=\frac1\beta\left[
        -\frac N2(\omega_2^2+\omega_3^2)
        +\lambda\left(\frac12\log(2\pi)+\log\omega_3+\frac12\right)
        +B\omega_2k_1^\omega+\frac12\sigma^2k_2^\omega
    \right].
\end{align*}

\begin{figure}
    \centering
    \begin{subfigure}[t]{0.22\textwidth}
        \centering
        \includegraphics[width=\textwidth]{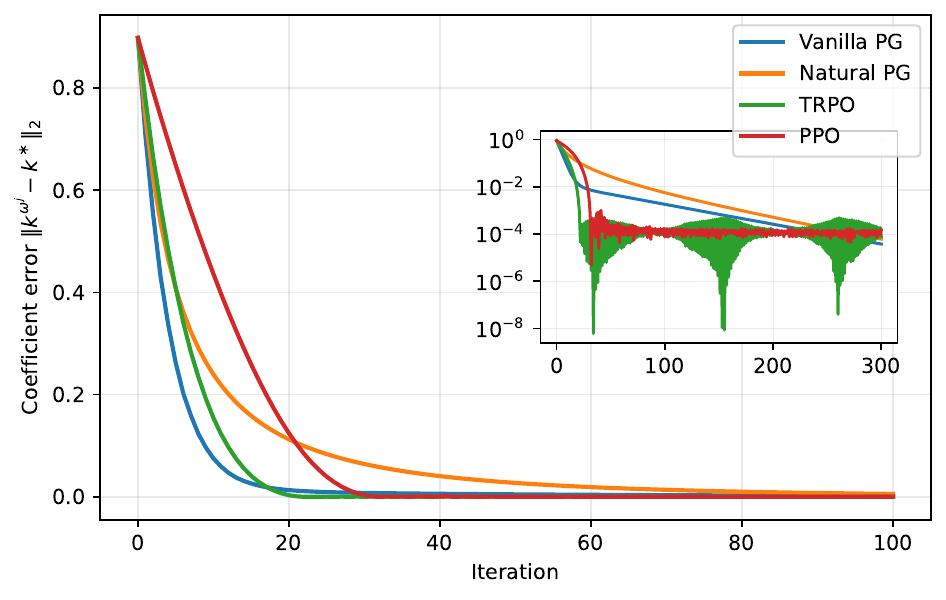}
        \caption{$A=-10$.}
        \label{known_dy_change_A}
    \end{subfigure}
    \begin{subfigure}[t]{0.22\textwidth}
        \centering
        \includegraphics[width=\textwidth]{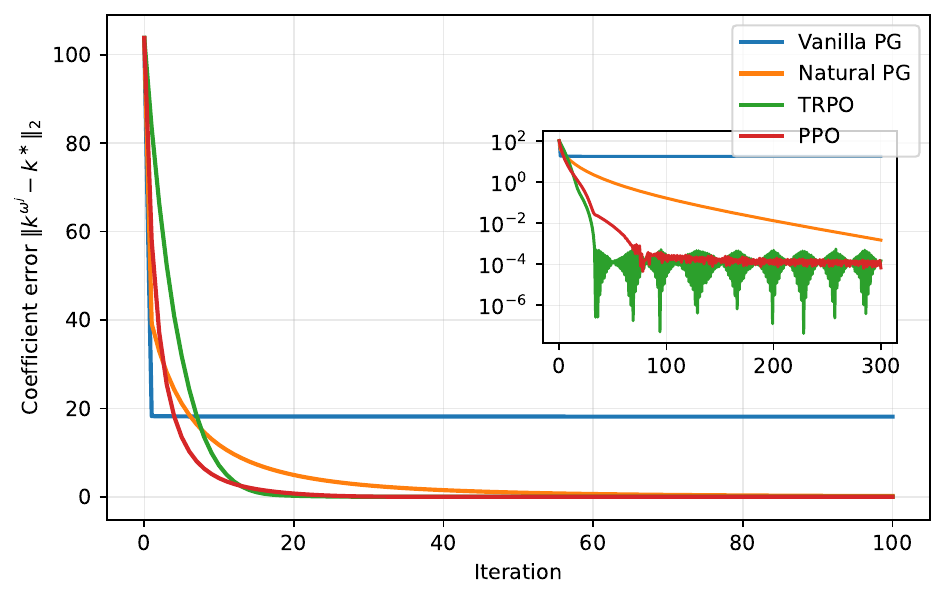}
        \caption{$B=10$, $\omega^0=(0.1,1,1)$.}
        \label{known_dy_change_B_1}
    \end{subfigure}
    \begin{subfigure}[t]{0.22\textwidth}
        \centering
        \includegraphics[width=\textwidth]{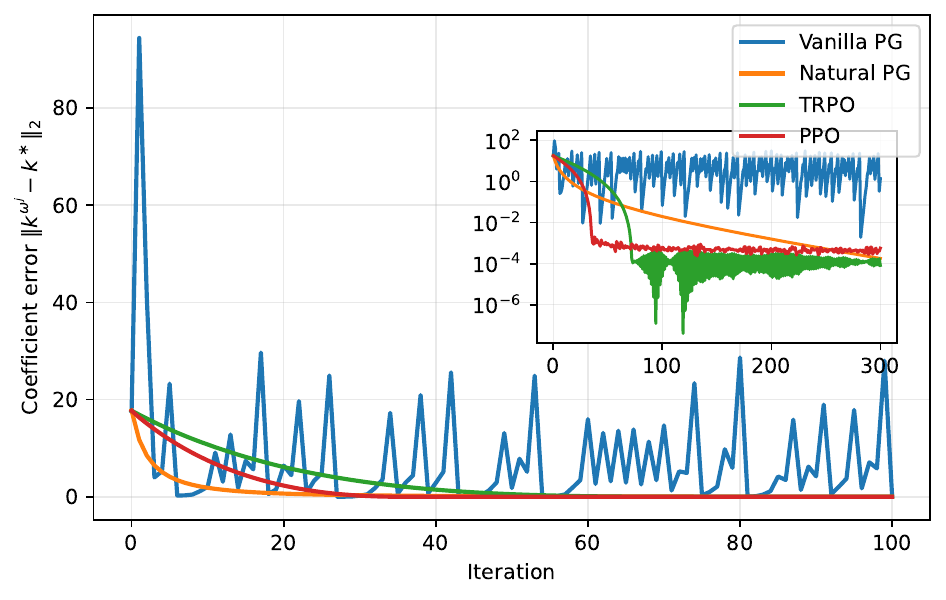}
        \caption{$\sigma=10$.}
        \label{known_dy_change_sigma}
    \end{subfigure}
    \begin{subfigure}[t]{0.22\textwidth}
        \centering
        \includegraphics[width=\textwidth]{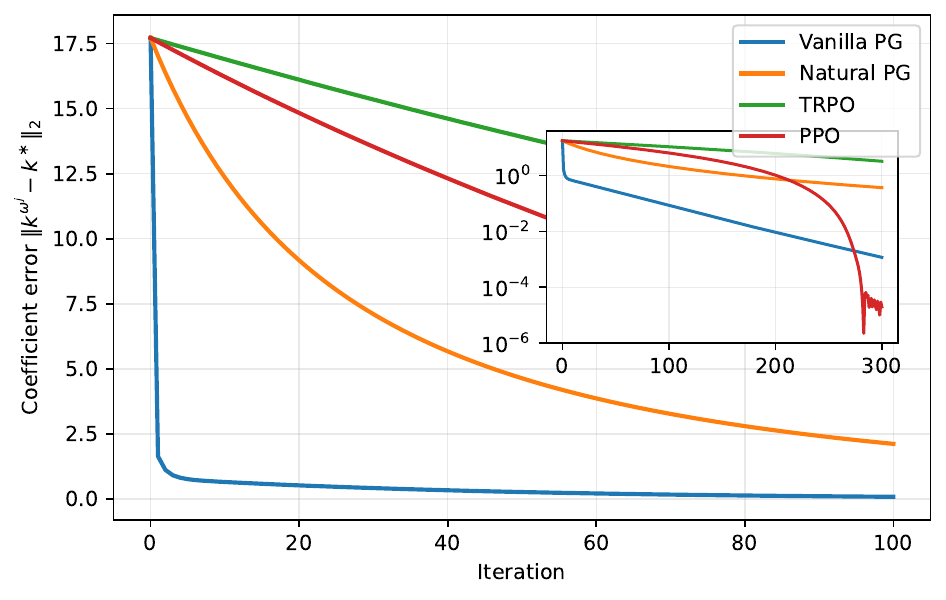}
        \caption{$\sigma=10$ and effective steps divided by $10$.}
        \label{known_dy_change_sigma_lr}
    \end{subfigure}
    \caption{Known-dynamics experiments under different choices of the dynamics parameters.}
    \label{fig:known_dy_parameter_variations}
\end{figure}

\noindent\textbf{Known dynamics.}
We first complement the known-dynamics results of \Cref{known_dynamics_results} by studying the sensitivity of the policy optimization methods to changes in the dynamics. Figure~\ref{fig:known_dy_parameter_variations} shows the effect of modifying $A$, $B$, and $\sigma$. Changing $A$ alters the transient convergence rate but does not change the overall behavior of the methods. The case $B=10$ is more discriminating: with the same fixed step used in the default configuration, vanilla PG stalls at a relatively large error, whereas NPG, TRPO, and PPO continue toward the optimum. Increasing $B$ changes the sensitivity of the closed loop to the policy parameter, so a step size that is appropriate for $B=1$ need not remain appropriate for $B=10$.

The experiments with $\sigma=10$ illustrate a different effect. In this LQR setting, $\sigma$ does not change the optimal feedback coefficients, but it broadens the state distribution used in the Monte Carlo policy gradient estimator. As a result, vanilla PG exhibits large spikes, whereas the preconditioned or constrained updates remain substantially more stable. Reducing the effective steps by a factor of ten suppresses these spikes, at the cost of slower convergence during the first iterations.

\begin{figure}
    \centering
    \begin{subfigure}[t]{0.22\textwidth}
        \centering
        \includegraphics[width=\textwidth]{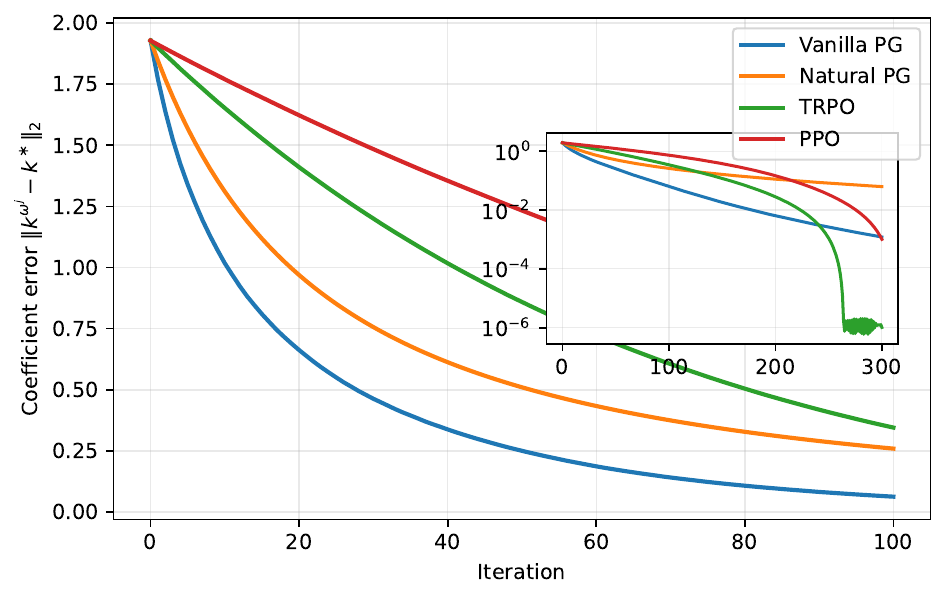}
        \caption{Effective steps divided by $10$.}
        \label{known_dy_lr_small}
    \end{subfigure}
    \begin{subfigure}[t]{0.22\textwidth}
        \centering
        \includegraphics[width=\textwidth]{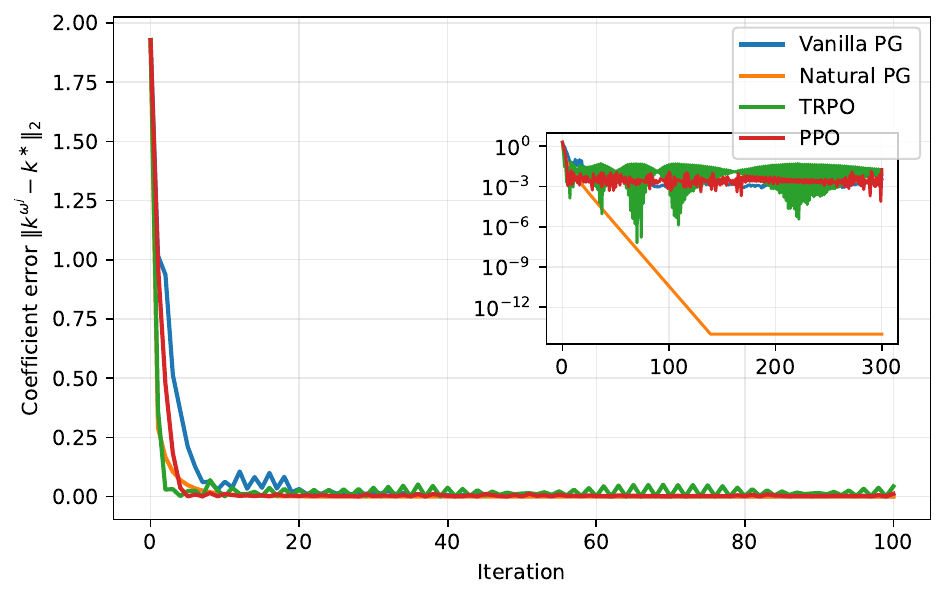}
        \caption{Effective steps multiplied by $10$.}
        \label{known_dy_lr_big}
    \end{subfigure}
    \begin{subfigure}[t]{0.22\textwidth}
        \centering
        \includegraphics[width=\textwidth]{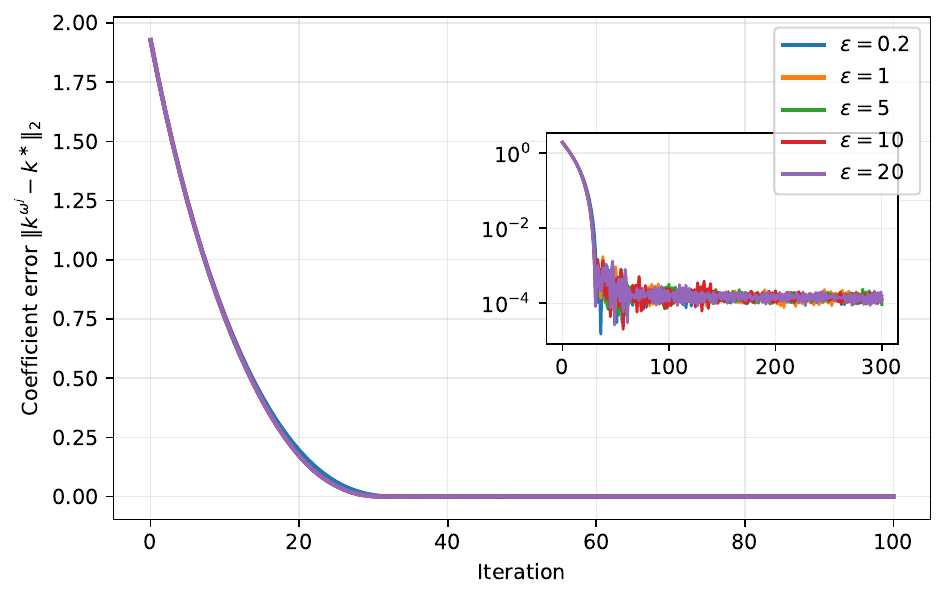}
        \caption{PPO clipping ratios.}
        \label{known_dy_ppo_epsl}
    \end{subfigure}
    \begin{subfigure}[t]{0.22\textwidth}
        \centering
        \includegraphics[width=\textwidth]{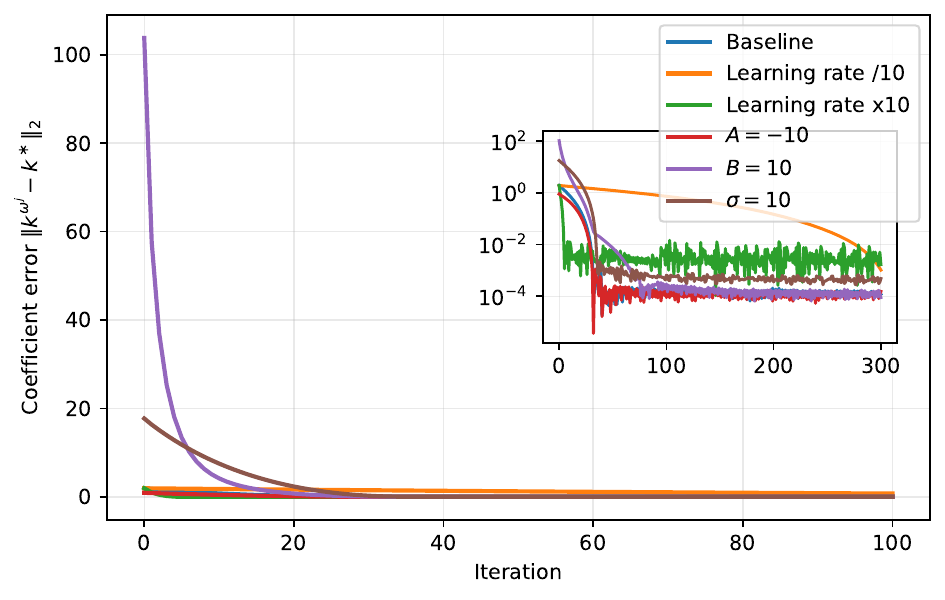}
        \caption{PPO under different settings.}
        \label{known_dy_ppo_robustness}
    \end{subfigure}
    \caption{Known-dynamics experiments: sensitivity to the update parameters.}
    \label{fig:known_dy_change_lr}
\end{figure}

We next study the sensitivity to the policy-update parameters. Figure~\ref{fig:known_dy_change_lr} confirms that the effective update size controls the main speed--stability trade-off. Dividing all steps by ten produces smooth but slower convergence. Multiplying them by ten accelerates the initial decrease but also produces oscillations and can move an iterate outside the admissible region. For TRPO, multiplying the effective step by a factor $c$ corresponds to multiplying the quadratic radius by $c^2$.

The PPO curves obtained with the tested clipping ratios are nearly identical during the main transient, with small differences appearing mainly in the long-run error. Thus, in this configuration, the clipping ratio is not the main factor limiting convergence. The learning rate and the scaling of the dynamics have a substantially larger effect on the observed stability.

\begin{figure}
    \centering
    \begin{subfigure}[t]{0.38\textwidth}
        \centering
        \includegraphics[width=\textwidth]{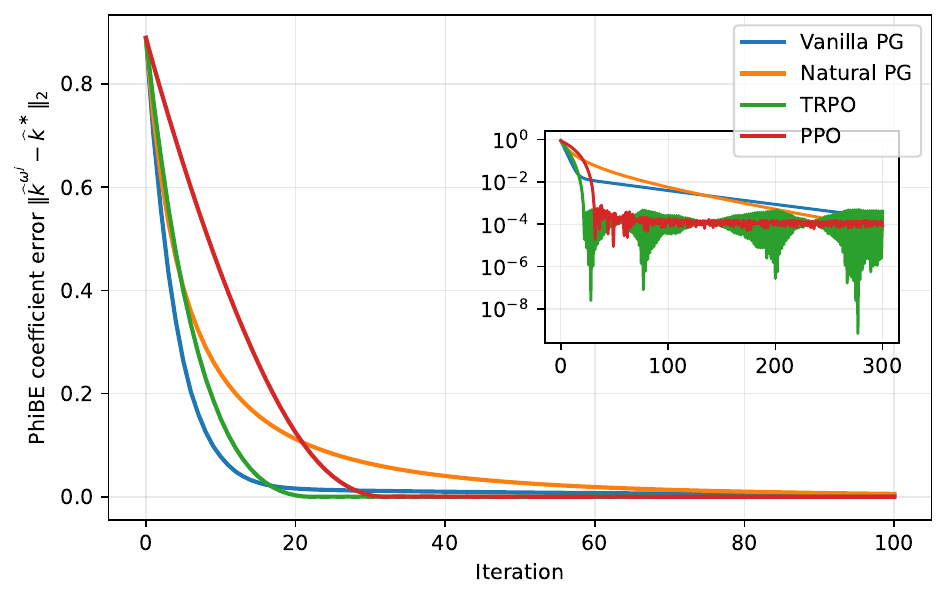}
        \caption{$A=-10$.}
        \label{PhiBE_change_A}
    \end{subfigure}
    \begin{subfigure}[t]{0.38\textwidth}
        \centering
        \includegraphics[width=\textwidth]{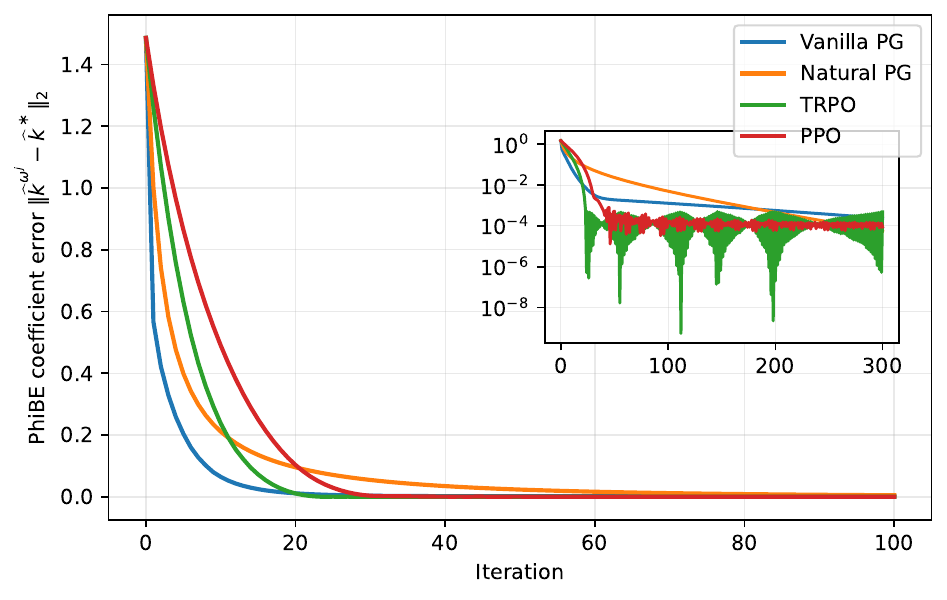}
        \caption{$B=10$, $\omega^0=(-0.5,1,1)$.}
        \label{PhiBE_change_B_2}
    \end{subfigure}
    \caption{Optimal-PhiBE with known discrete-time dynamics: changes in $A$ and $B$.}
    \label{fig:PhiBE_change_A_B_extra}
\end{figure}

\noindent\textbf{PhiBE with known discrete-time dynamics.}
We complement the results of \Cref{sec:lqr_phibe} by studying the sensitivity of Optimal-PhiBE to changes in the dynamics and in the policy-update parameters. Figure~\ref{fig:PhiBE_change_A_B_extra} considers changes in $A$ and $B$. For $A=-10$, all methods continue to converge to a low error with respect to $k^*$, although the transient behavior changes. The case $B=10$ is again more sensitive to the policy initialization and update size. With the admissible initialization $\omega^0=(-0.5,1,1)$, the four methods reach the low-error region, showing behavior consistent with the corresponding known-dynamics experiment.

\begin{figure}
    \centering
    \begin{subfigure}[t]{0.22\textwidth}
        \centering
        \includegraphics[width=\textwidth]{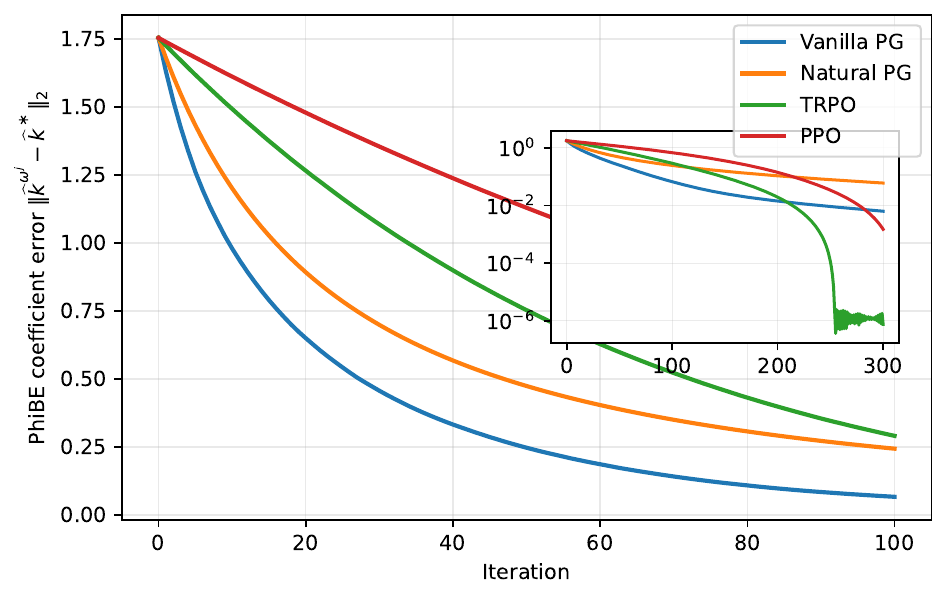}
        \caption{Effective steps divided by $10$.}
        \label{PhiBE_lr_small}
    \end{subfigure}
    \begin{subfigure}[t]{0.22\textwidth}
        \centering
        \includegraphics[width=\textwidth]{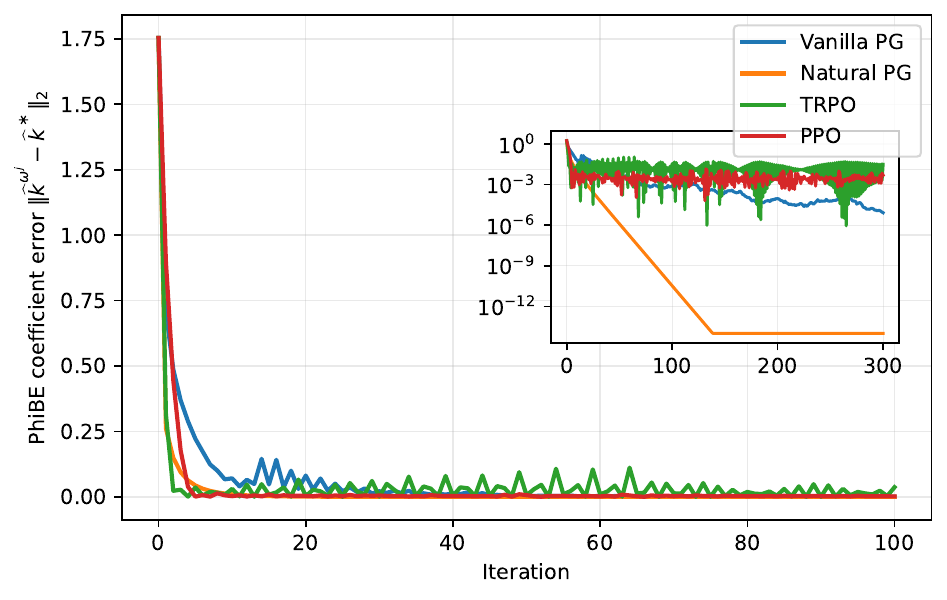}
        \caption{Effective steps multiplied by $10$.}
        \label{PhiBE_lr_large}
    \end{subfigure}
    \begin{subfigure}[t]{0.22\textwidth}
        \centering
        \includegraphics[width=\textwidth]{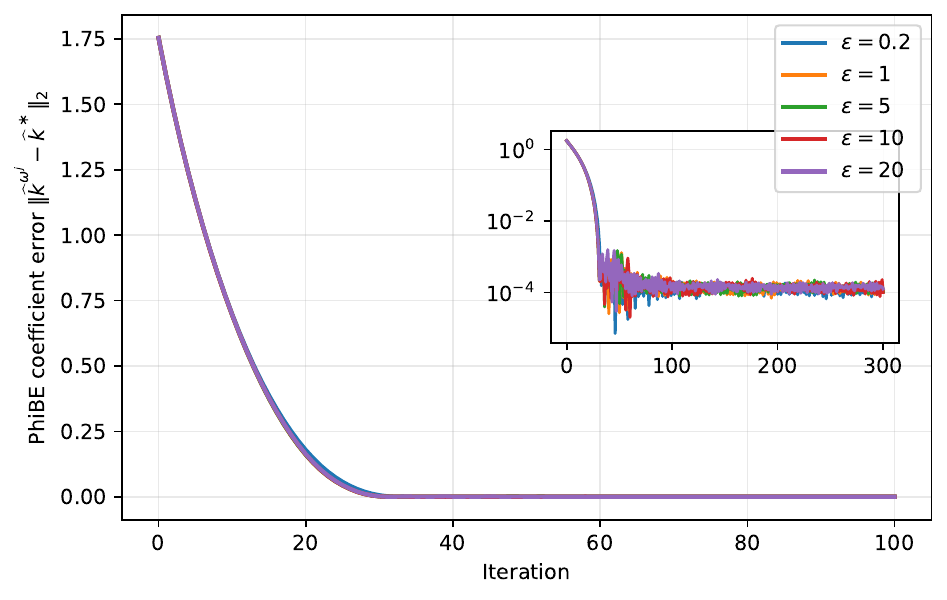}
        \caption{PPO clipping ratios.}
        \label{PhiBE_change_epsl}
    \end{subfigure}
    \begin{subfigure}[t]{0.22\textwidth}
        \centering
        \includegraphics[width=\textwidth]{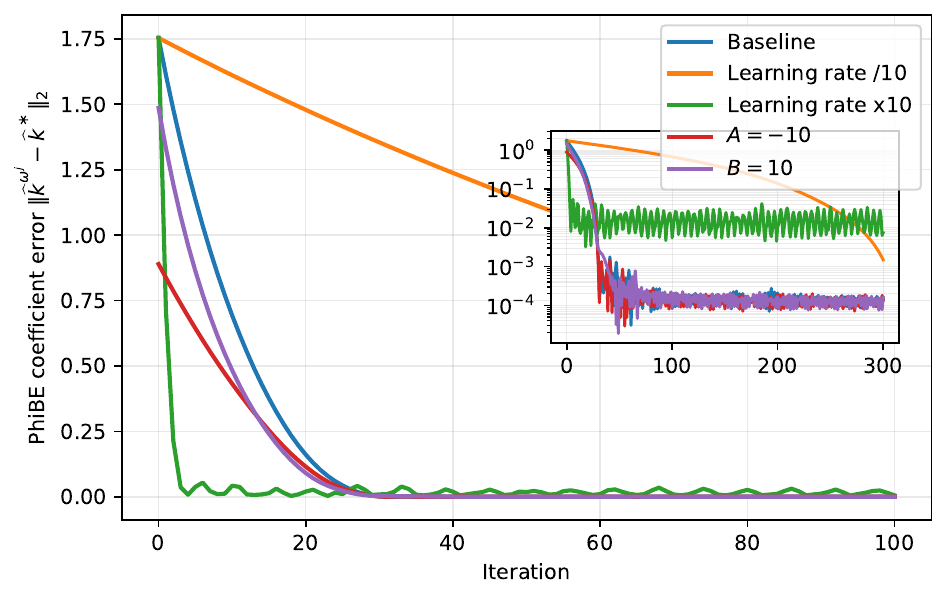}
        \caption{PPO robustness with $\epsilon=10$.}
        \label{PhiBE_ppo_robustness}
    \end{subfigure}
    \caption{Optimal-PhiBE with known discrete-time dynamics: update-parameter studies.}
    \label{fig:PhiBE_change_lr}
\end{figure}

Figure~\ref{fig:PhiBE_change_lr} studies the effect of the update parameters. Dividing the effective steps by ten gives smoother but slower convergence. Multiplying them by ten accelerates the initial decrease and reaches the low-error region quickly, but produces a more visible oscillatory floor. Thus, as in the known-dynamics case, the effective update size determines the main trade-off between convergence speed and stability.

The PPO curves corresponding to the tested clipping ratios remain close throughout the optimization, with differences appearing mainly at the long-run error level. The additional PPO experiment shows the same qualitative robustness observed in the known-dynamics setting. For these experiments, the learning rate therefore has a more visible effect on the optimization than the clipping ratio.

\medbreak

\noindent\textbf{Data-driven PhiBE.}
We provide here the implementation details and additional experiments for the model-free setting of \Cref{sec:lqr_on_policy}. At every actor iteration, a fresh batch is generated from the current policy and used both to estimate the PhiBE critic and to construct the policy update.

\begin{algorithm}
\caption{Data-driven PhiBE algorithm for LQR}
\label{alg:on_policy_lqr}
\begin{algorithmic}[1]
\State \textbf{Input:} initial policy $\omega^0$, basis $\phi(s)=(1,s,s^2)^\top$, batch size $n$, local interval $\Delta t$, and update parameters.
\For{$k=0,1,\ldots,K-1$}
    \State Sample $\tau_i\sim\operatorname{Exp}(\beta)$ and $S_i=S_{\tau_i}$ from the current policy, starting from $S_0\sim\mathcal U(-1,1)$.
    \State Draw $a_i\sim\pi_{\omega^k}(\cdot|S_i)$ and observe $(S_i,a_i,r_i,S_i')$, $i=1,\ldots,n$.
    \State Set $\Delta S_i=S_i'-S_i$ and $\hat b_i=\Delta S_i/\Delta t$.
    \State Form
    \[
        F_k=\frac1n\sum_{i=1}^n\phi(S_i)
        \left[
            \beta\phi(S_i)
            -\partial_s\phi(S_i)\hat b_i
        \right]^\top,
    \]
    \State Form
    \[
        g_k=\frac1n\sum_{i=1}^n
        \big(r_i-\lambda\log\pi_{\omega^k}(a_i|S_i)\big)\phi(S_i),
        \qquad \xi_k=F_k^{-1}g_k,
    \]
    and set $\hat V_k(s)=\xi_k^\top\phi(s)$.
    \State Evaluate
    \[
        \hat A_i=r_i-\lambda\log\pi_{\omega^k}(a_i|S_i)
        +\hat b_i\,\partial_s\hat V_k(S_i)
        -\beta\hat V_k(S_i).
    \]
    \State Use $(1/\beta)n^{-1}\sum_{i=1}^n\hat A_i\nabla_\omega\log\pi_{\omega^k}(a_i|S_i)$, or the corresponding NPG, TRPO, or PPO direction, to obtain $\omega^{k+1}$.
\EndFor
\State \textbf{Return:} $\omega^K$.
\end{algorithmic}
\end{algorithm}

\begin{figure}
    \centering
    \begin{subfigure}[t]{0.19\textwidth}
        \centering
        \includegraphics[width=\textwidth]{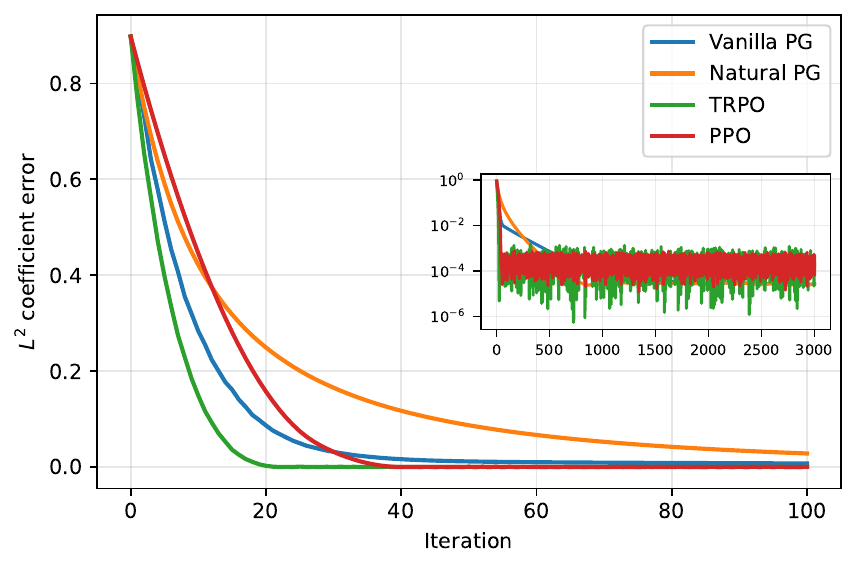}
        \caption{$A=-10$.}
        \label{data_driven_change_A}
    \end{subfigure}
    \begin{subfigure}[t]{0.19\textwidth}
        \centering
        \includegraphics[width=\textwidth]{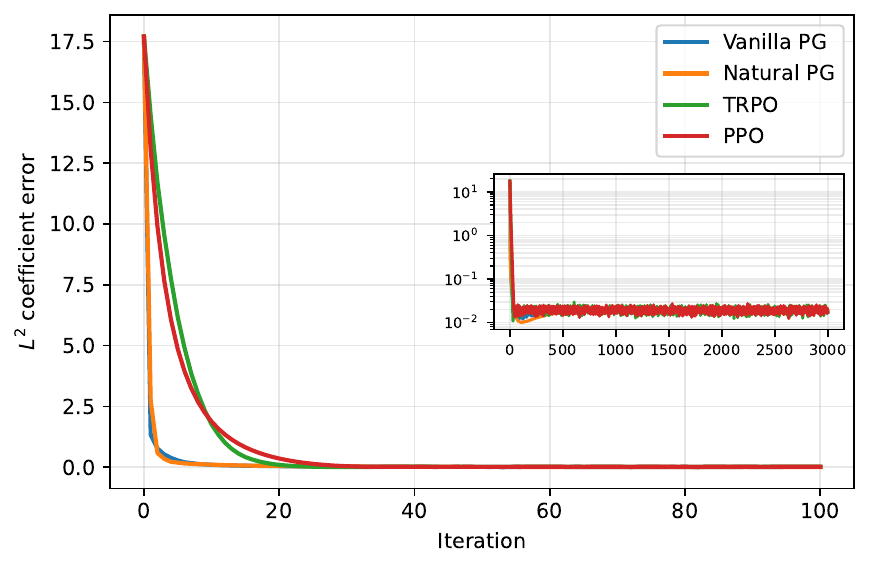}
        \caption{$B=10$, $\omega^0=(0,1,1)$.}
        \label{data_driven_change_B_1}
    \end{subfigure}
    \begin{subfigure}[t]{0.19\textwidth}
        \centering
        \includegraphics[width=\textwidth]{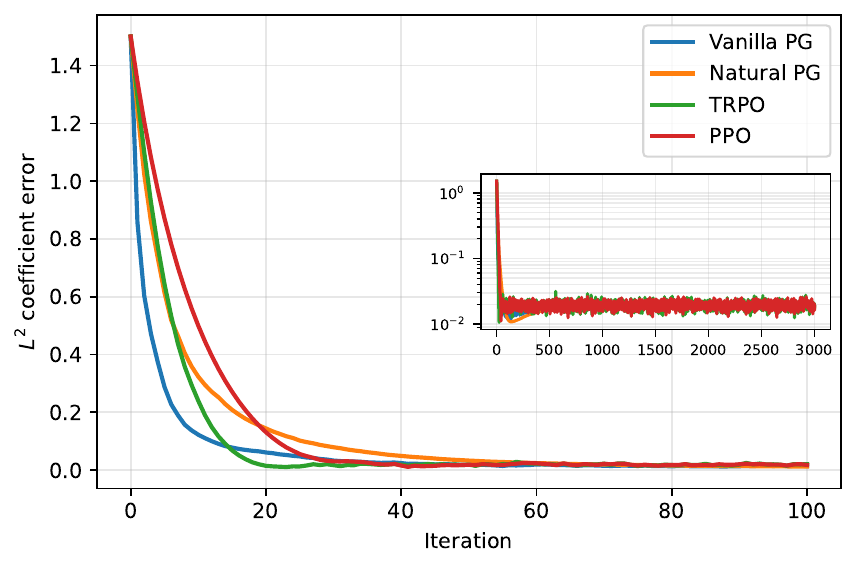}
        \caption{$B=10$, $\omega^0=(-0.5,1,1)$.}
        \label{data_driven_change_B_2}
    \end{subfigure}
    \begin{subfigure}[t]{0.19\textwidth}
        \centering
        \includegraphics[width=\textwidth]{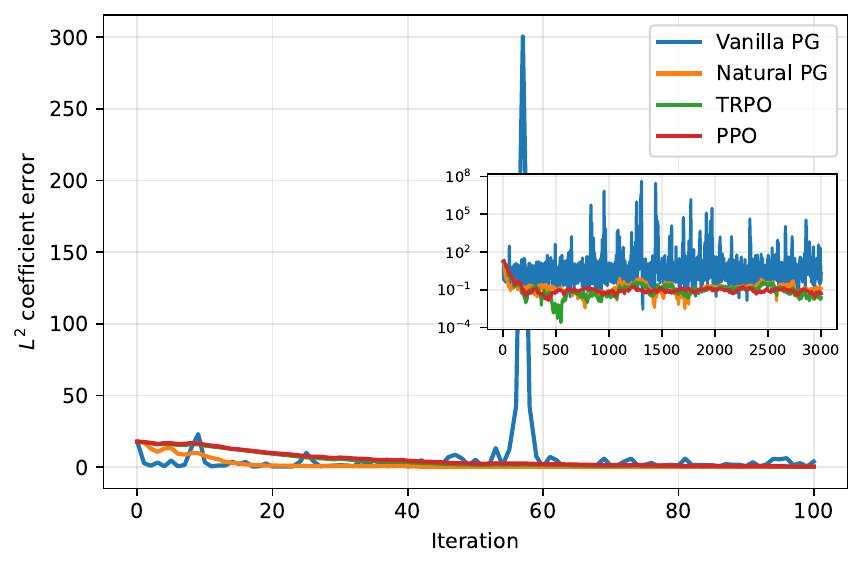}
        \caption{$\sigma=10$.}
        \label{data_driven_change_sigma}
    \end{subfigure}
    \begin{subfigure}[t]{0.19\textwidth}
        \centering
        \includegraphics[width=\textwidth]{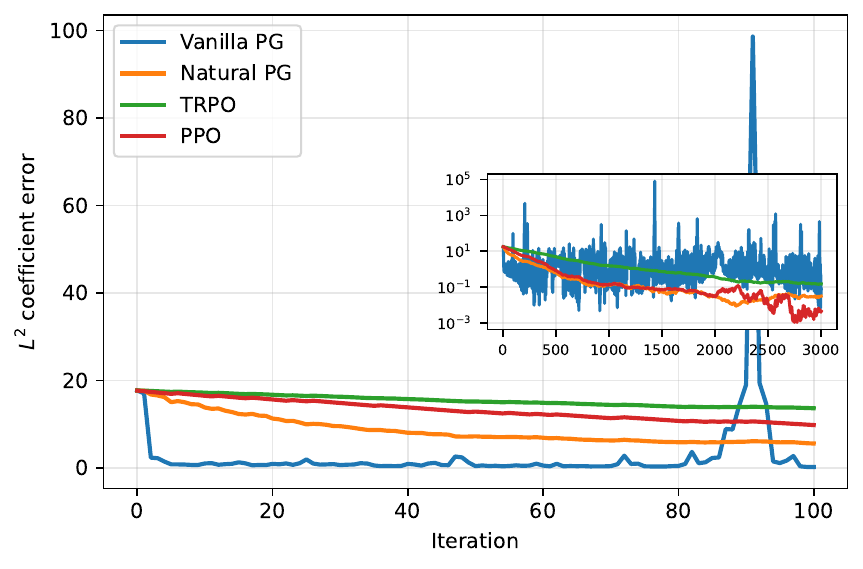}
        \caption{$\sigma=10$ and effective steps divided by $10$.}
        \label{data_driven_change_sigma_lr}
    \end{subfigure}
    \caption{Data-driven PhiBE: changes in the dynamics.}
    \label{fig:data_driven_change_dynamics_extra}
\end{figure}

We first study the effect of changing the dynamics. Figure~\ref{fig:data_driven_change_dynamics_extra} shows that the $A=-10$ case and both admissible initializations for $B=10$ still enter a low-error region. Unlike the corresponding known-dynamics experiments, the curves fluctuate around a nonzero level because the critic and advantage are reconstructed from a new finite batch at every iteration. These fluctuations therefore persist even after the main optimization transient has ended.

High diffusion provides the main stress test. For $\sigma=10$, vanilla PG exhibits occasional large excursions, while NPG, TRPO, and PPO remain in a substantially narrower range. Dividing the effective steps by ten reduces the size of the usual policy-update fluctuations, but also slows convergence. The remaining large excursions show that decreasing the actor step alone does not remove the variability introduced by the finite-sample drift estimates $\hat b_i=\Delta S_i/\Delta t$.

\begin{figure}
    \centering
    \begin{subfigure}[t]{0.22\textwidth}
        \centering
        \includegraphics[width=\textwidth]{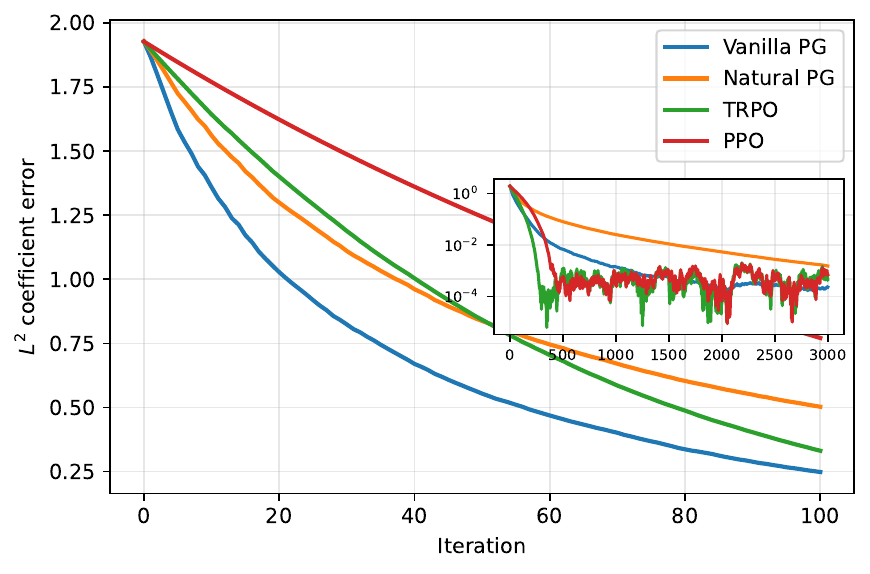}
        \caption{Effective steps divided by $10$.}
        \label{data_driven_lr_small}
    \end{subfigure}
    \begin{subfigure}[t]{0.22\textwidth}
        \centering
        \includegraphics[width=\textwidth]{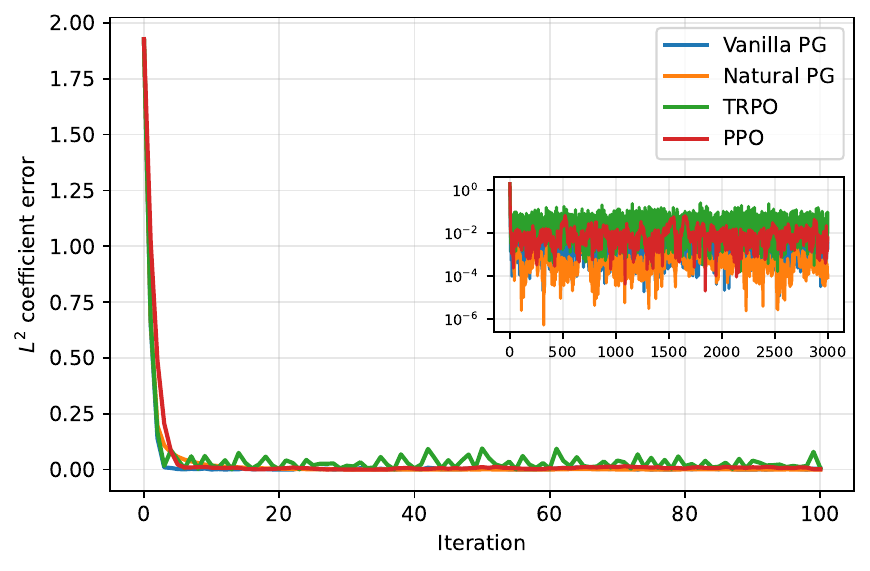}
        \caption{Effective steps multiplied by $10$.}
        \label{data_driven_lr_large}
    \end{subfigure}
    \begin{subfigure}[t]{0.22\textwidth}
        \centering
        \includegraphics[width=\textwidth]{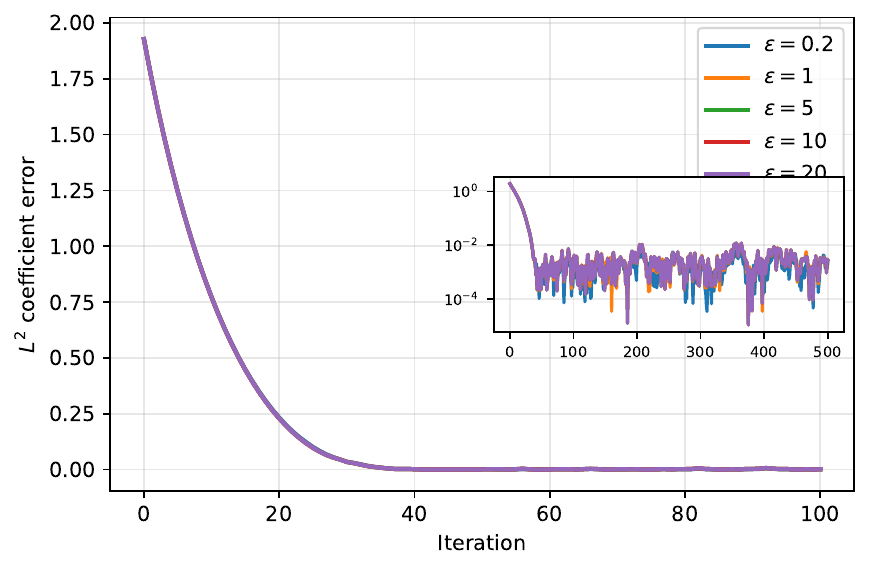}
        \caption{PPO clipping ratios.}
        \label{data_driven_change_epsl}
    \end{subfigure}
    \begin{subfigure}[t]{0.22\textwidth}
        \centering
        \includegraphics[width=\textwidth]{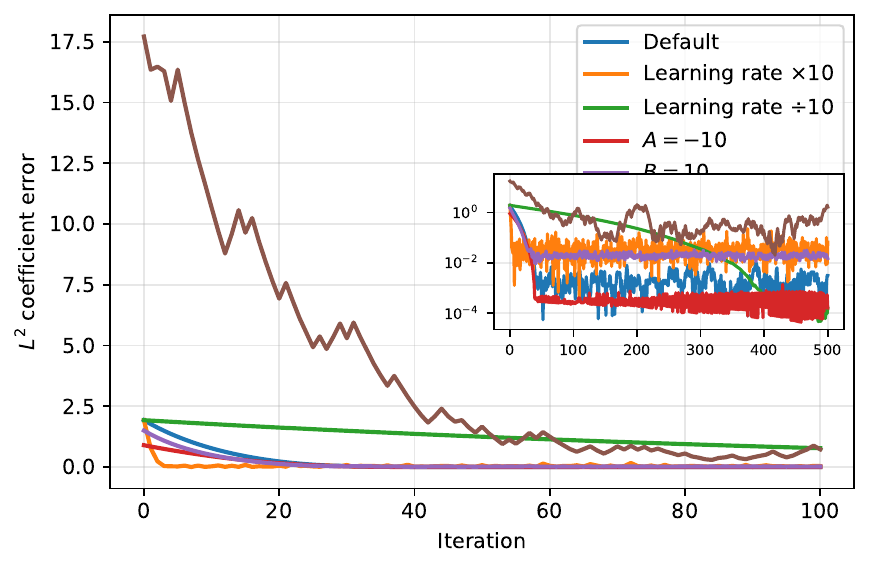}
        \caption{PPO robustness with $\epsilon=10$.}
        \label{data_driven_ppo_robustness}
    \end{subfigure}
    \caption{Data-driven PhiBE: sensitivity to the update parameters.}
    \label{fig:data_driven_change_lr}
\end{figure}

Figure~\ref{fig:data_driven_change_lr} studies the sensitivity to the policy-update parameters. Dividing the effective steps by ten produces smoother but slower convergence, whereas multiplying them by ten reaches the low-error region faster but increases the iteration-to-iteration fluctuations. Thus, the update size controls the speed--stability trade-off, while the persistent long-run fluctuations remain associated with the resampled critic and advantage.

The PPO curves remain close for the tested clipping ratios, indicating that clipping has a comparatively small effect on the long-run error in this configuration. In the robustness experiment, taking $\epsilon=10$ largely removes the effect of clipping. Combined with a larger learning rate, this leads to substantially less stable updates, illustrating the role of restricting the policy change when the critic itself is estimated from fresh finite-sample data.

\medbreak
\noindent\textbf{Online data-driven PhiBE.}
We provide implementation details and additional experiments for \Cref{sec:lqr_online}. The initial policy is $\omega^0=(0.5,1,1)$, the initial-state distribution is $\rho_0=\mathcal U(-1,1)$, the critic starts from $\xi_0=0$, and $\Delta t=0.1$. To obtain a state from the current discounted occupancy measure, the synthetic simulator draws $\tau_k\sim\operatorname{Exp}(\beta)$ and samples $S_{\tau_k}$ using the exact policy-averaged LQR transition law. It then samples an action from the current policy and generates one local transition with that action held fixed. The critic and actor updates use the resulting tuple; the simulator's occupancy-state draws are not counted in the budget of $200{,}000$ local transitions. The exact environment parameters are also used to evaluate the plotted coefficient error and by the numerical admissibility safeguards.

Vanilla PG follows \eqref{eq:online_critic_update} and \eqref{eq:online_actor_update}, with $\alpha_k=0.003(1+k/1000)^{-0.55}$ and $\eta_k=0.001(1+k/1000)^{-0.8}$. In the PPO variant, the same critic is updated after every transition, but the actor is updated after each window of $W=2$ transitions. The stored advantages are then evaluated with the updated critic and used for two epochs of the clipped surrogate with $\epsilon=0.2$. Each epoch uses $W\eta_k$ times the mean surrogate gradient, with $k$ taken at the end of the window. Thus, the two actor implementations have different update frequencies and effective step sizes. The default comparison is repeated over five seeds; the other parameter studies use one seed per configuration.

\begin{figure}
    \centering
    \begin{subfigure}[t]{0.23\textwidth}
        \centering
        \includegraphics[width=\textwidth]{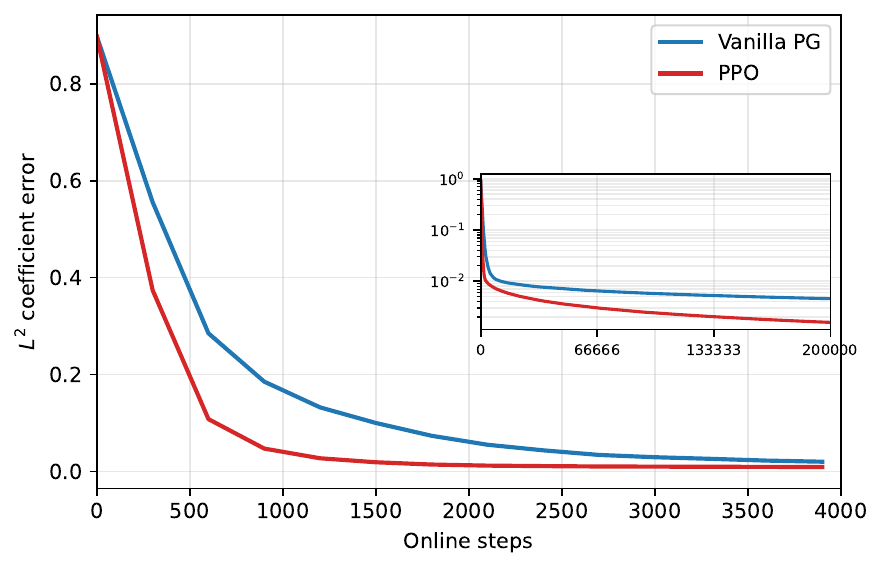}
        \caption{$A=-10$.}
        \label{fig:online_change_A}
    \end{subfigure}\hfill
    \begin{subfigure}[t]{0.23\textwidth}
        \centering
        \includegraphics[width=\textwidth]{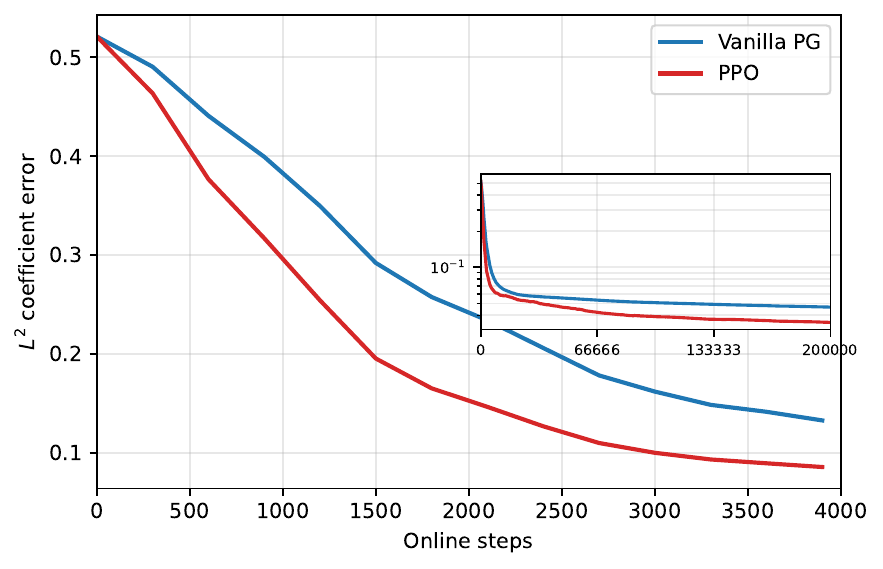}
        \caption{$B=10$, $\omega^0=(0,0,1)$.}
        \label{fig:online_change_B}
    \end{subfigure}\hfill
    \begin{subfigure}[t]{0.23\textwidth}
        \centering
        \includegraphics[width=\textwidth]{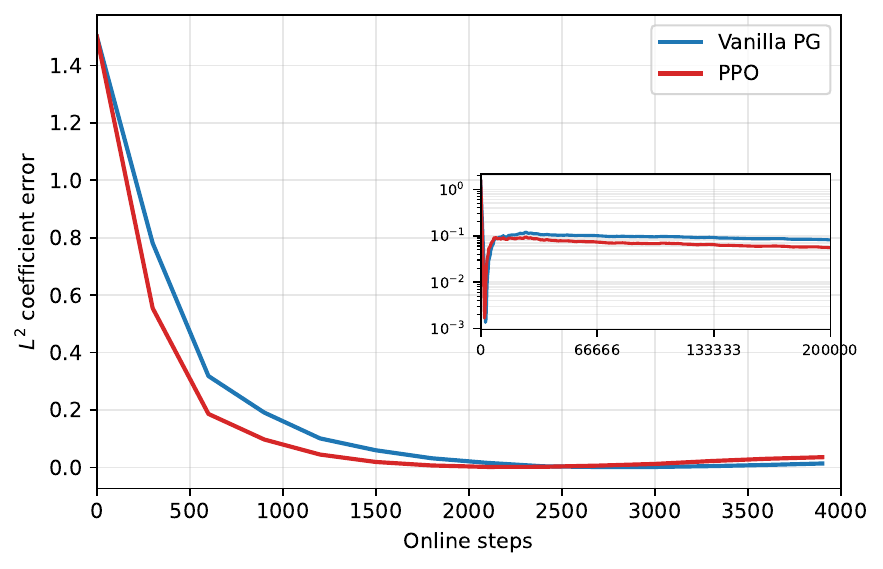}
        \caption{$B=10$, $\omega^0=(-0.5,1,1)$.}
        \label{fig:online_change_B_init2}
    \end{subfigure}\hfill
    \begin{subfigure}[t]{0.23\textwidth}
        \centering
        \includegraphics[width=\textwidth]{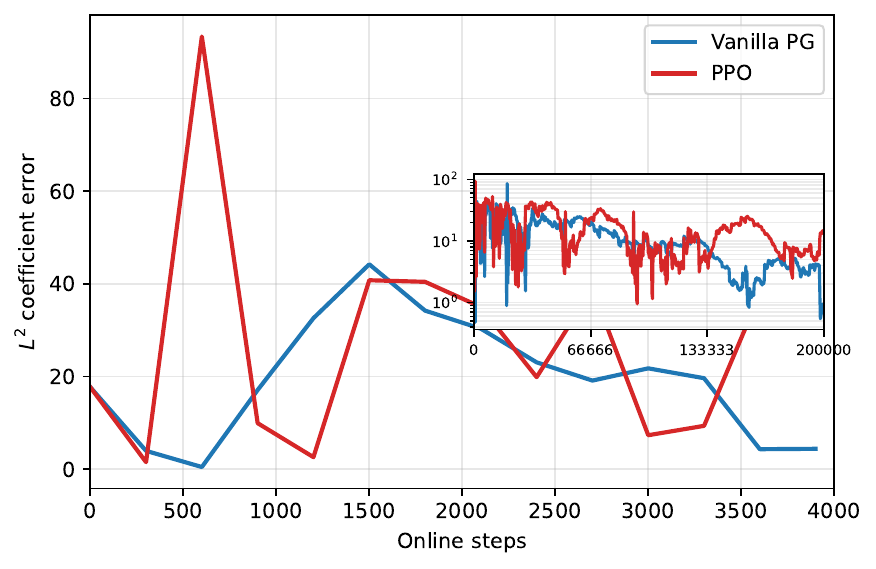}
        \caption{$\sigma=10$.}
        \label{fig:online_change_sigma}
    \end{subfigure}
    \caption{Online data-driven PhiBE: changes in the dynamics. Both learning rates are divided by ten in the $B=10$ and $\sigma=10$ experiments.}
    \label{fig:online_dynamics}
\end{figure}

Figure~\ref{fig:online_dynamics} examines changes in the dynamics. For $A=-10$, both methods continue to reduce the error, although their final errors are larger than under the default dynamics. With $B=10$, convergence is more gradual and the remaining error depends on the initial policy. The high-diffusion case $\sigma=10$ is markedly harder: both curves show large excursions and remain well above the default error level. Critic-step caps and policy projections activate frequently in this case, so these trajectories should not be interpreted as unconstrained convergence of either actor. For the $B=10$ and $\sigma=10$ stress tests, both learning rates are divided by ten.

\begin{figure}
    \centering
    \begin{subfigure}[t]{0.23\textwidth}
        \centering
        \includegraphics[width=\textwidth]{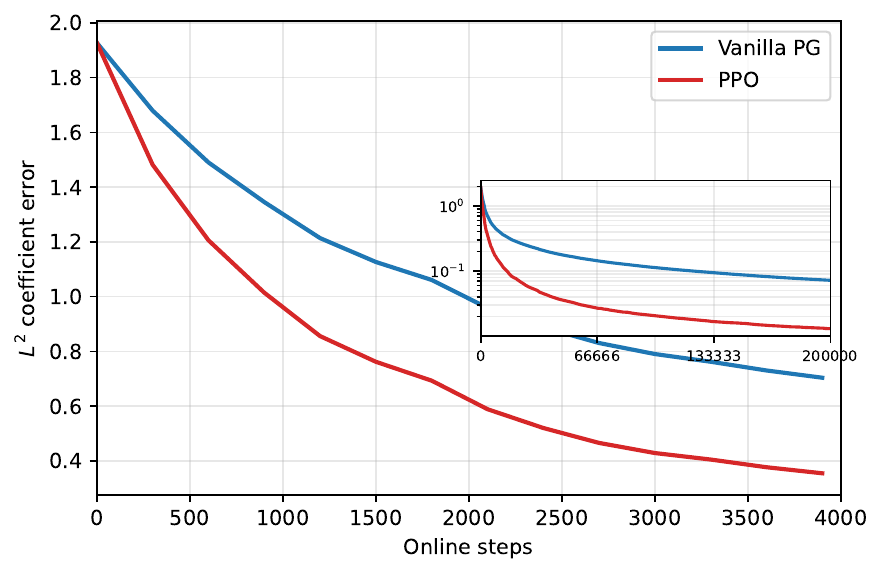}
        \caption{Learning rates divided by $10$.}
        \label{fig:online_lr_small}
    \end{subfigure}\hfill
    \begin{subfigure}[t]{0.23\textwidth}
        \centering
        \includegraphics[width=\textwidth]{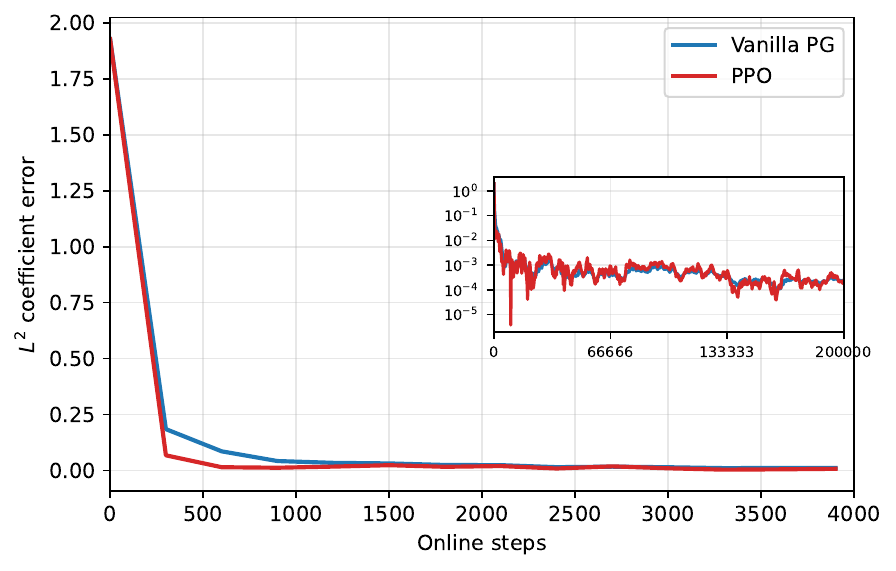}
        \caption{Learning rates multiplied by $2$.}
        \label{fig:online_lr_large}
    \end{subfigure}\hfill
    \begin{subfigure}[t]{0.23\textwidth}
        \centering
        \includegraphics[width=\textwidth]{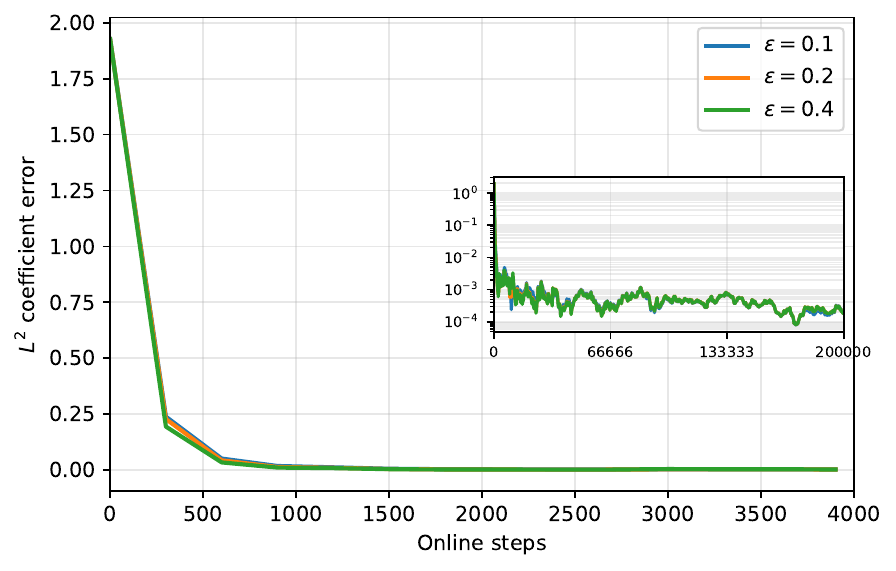}
        \caption{PPO clipping ratios.}
        \label{fig:online_ppo_clip}
    \end{subfigure}\hfill
    \begin{subfigure}[t]{0.23\textwidth}
        \centering
        \includegraphics[width=\textwidth]{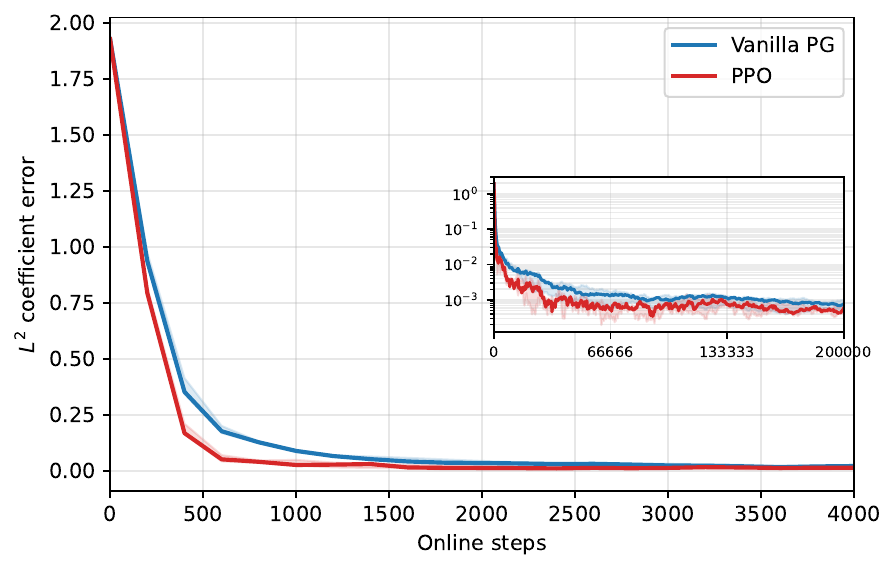}
        \caption{Five independent seeds.}
        \label{fig:online_seeds}
    \end{subfigure}
    \caption{Online data-driven PhiBE: update parameters and seed-to-seed variation. The last panel shows medians and interquartile bands.}
    \label{fig:online_updates}
\end{figure}

Figure~\ref{fig:online_updates} studies the update parameters and variation across seeds. Dividing both learning rates by ten slows convergence and leaves a larger error at the fixed transition budget. Doubling them accelerates convergence and remains stable in this tested configuration. The PPO curves for $\epsilon\in\{0.1,0.2,0.4\}$ are nearly indistinguishable; clipping is rarely active under the default parameters. Finally, the five-seed experiment gives median final errors of approximately $7.5\times10^{-4}$ for Vanilla PG and $5.8\times10^{-4}$ for PPO. These results show comparable final accuracy, without isolating an effect of clipping from the other differences between the actor updates.

\subsubsection{Differential wheeled robot}

We provide here the implementation details omitted from \Cref{subsec:example_differential_wheeled}. The value network has three hidden layers of width $128$, while the policy network has one hidden layer of width $128$. All hidden layers use SiLU activations. Collocation states are sampled on $[-0.5,1.5]^2\times[-\pi,\pi)$, with $70\%$ uniform samples and $30\%$ samples concentrated near the target, using spatial standard deviation $0.3$. Adam and gradient clipping at norm $10$ are used throughout, with PPO clipping parameter $0.2$.

For known dynamics, fixed-policy evaluation uses $10{,}000$ critic steps. Joint training records epochs $0$ through $60$. Each policy stage is preceded by $1000$ critic steps using batches of $3000$ states and followed by $40$ PPO batches of size $3000$, with five PPO updates per batch. The critic and actor learning rates are both $10^{-3}$.

For data-driven dynamics, the observation interval is $\dt=0.01$, and fixed-policy evaluation again uses $10{,}000$ critic steps. Joint training records epochs $0$ through $360$. Each critic step uses $1000$ collocation states. Every policy stage is followed by $20$ actor batches of $5000$ discounted-occupancy states, with five PPO updates per batch. The critic and actor learning rates are $10^{-3}$ and $5\times10^{-4}$, respectively.

The reference values used for validation are computed independently from the training procedure using the known simulator with $64$ Monte Carlo rollouts, horizon $6$, and time step $10^{-3}$ on a $31\times31$ spatial grid. These reference values are not used in either the critic or actor loss. Trajectory plots use the deterministic policy mean, time step $0.01$, and horizon $5$. All trajectory panels display the same window $[-0.1,1.1]^2$; this restriction is only used for visualization and does not modify the simulated trajectories.

\noindent\textbf{Known-dynamics nonlinear robot wheel.}
We provide here the value-function and trajectory comparisons corresponding to the known-dynamics experiments of \Cref{subsec:example_differential_wheeled}.

\begin{figure}
    \centering
    \begin{subfigure}[t]{0.22\textwidth}
        \centering
        \includegraphics[width=\textwidth]{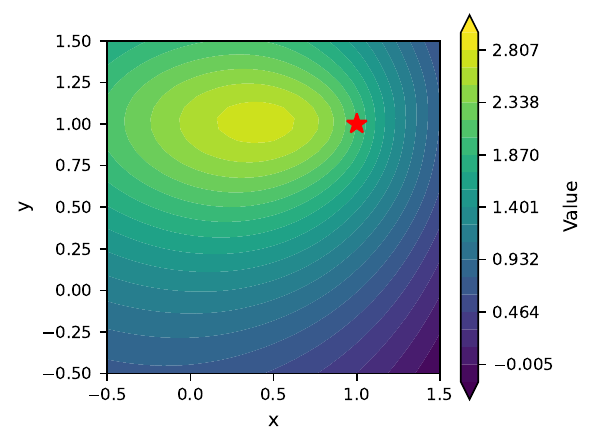}
        \caption{Fitted, $\theta=0$.}
        \label{fig:robot_pe_kd_pred_theta0}
    \end{subfigure}
    \begin{subfigure}[t]{0.22\textwidth}
        \centering
        \includegraphics[width=\textwidth]{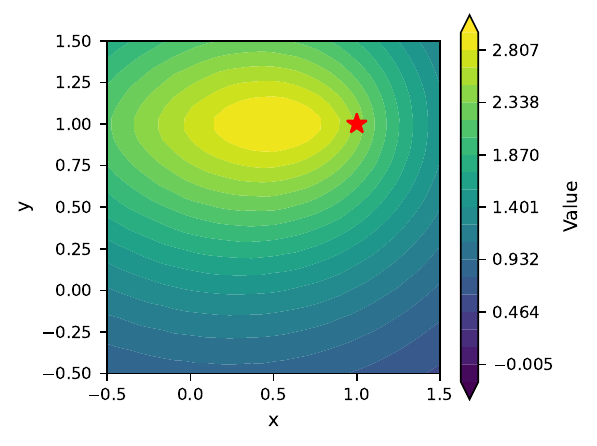}
        \caption{Reference, $\theta=0$.}
        \label{fig:robot_pe_kd_ref_theta0}
    \end{subfigure}
    \begin{subfigure}[t]{0.22\textwidth}
        \centering
        \includegraphics[width=\textwidth]{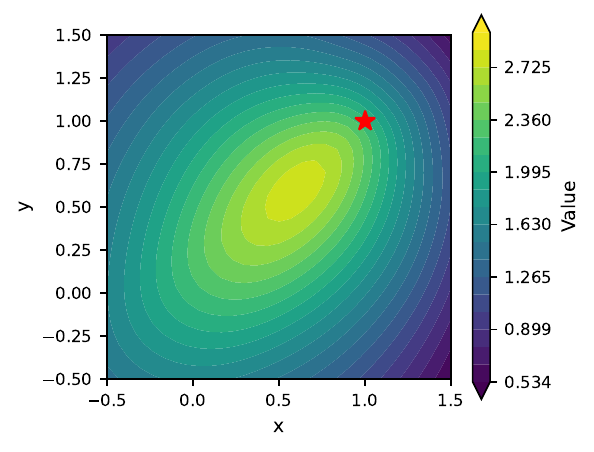}
        \caption{Fitted, $\theta=\pi/4$.}
        \label{fig:robot_pe_kd_pred_thetapi_4}
    \end{subfigure}
    \begin{subfigure}[t]{0.22\textwidth}
        \centering
        \includegraphics[width=\textwidth]{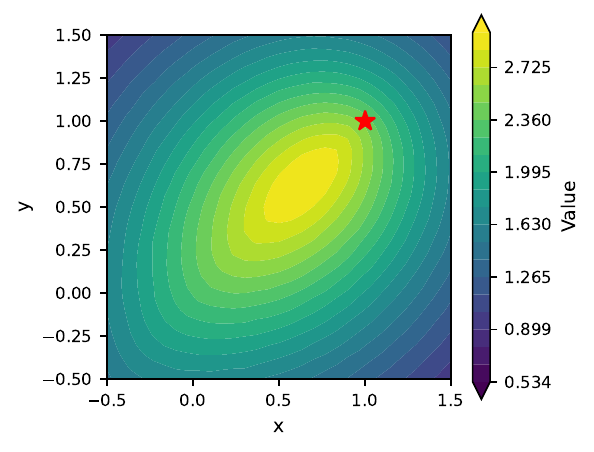}
        \caption{Reference, $\theta=\pi/4$.}
        \label{fig:robot_pe_kd_ref_thetapi_4}
    \end{subfigure}

    \medskip
    \begin{subfigure}[t]{0.22\textwidth}
        \centering
        \includegraphics[width=\textwidth]{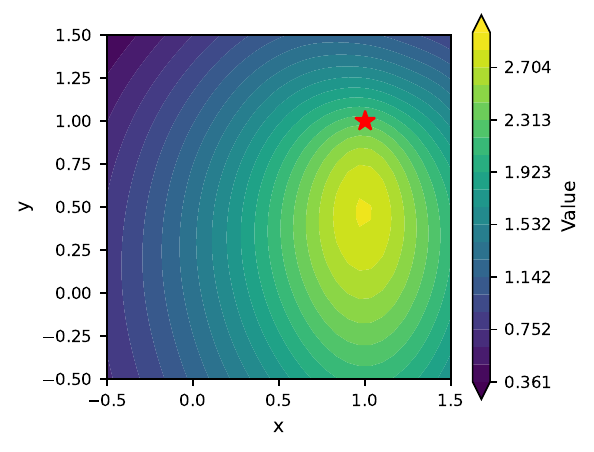}
        \caption{Fitted, $\theta=\pi/2$.}
        \label{fig:robot_pe_kd_pred_thetapi_2}
    \end{subfigure}
    \begin{subfigure}[t]{0.22\textwidth}
        \centering
        \includegraphics[width=\textwidth]{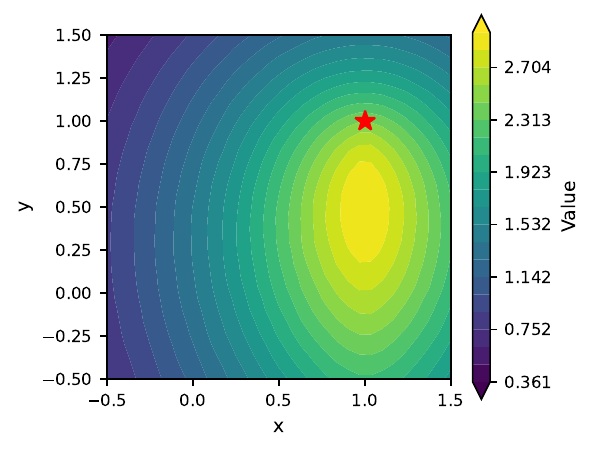}
        \caption{Reference, $\theta=\pi/2$.}
        \label{fig:robot_pe_kd_ref_thetapi_2}
    \end{subfigure}
    \begin{subfigure}[t]{0.22\textwidth}
        \centering
        \includegraphics[width=\textwidth]{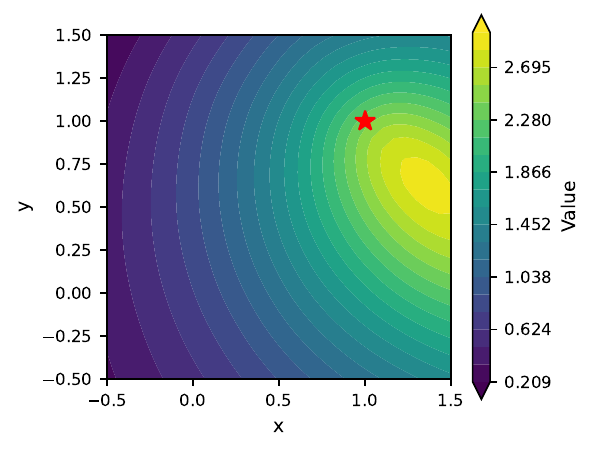}
        \caption{Fitted, $\theta=3\pi/4$.}
        \label{fig:robot_pe_kd_pred_theta3pi_4}
    \end{subfigure}
    \begin{subfigure}[t]{0.22\textwidth}
        \centering
        \includegraphics[width=\textwidth]{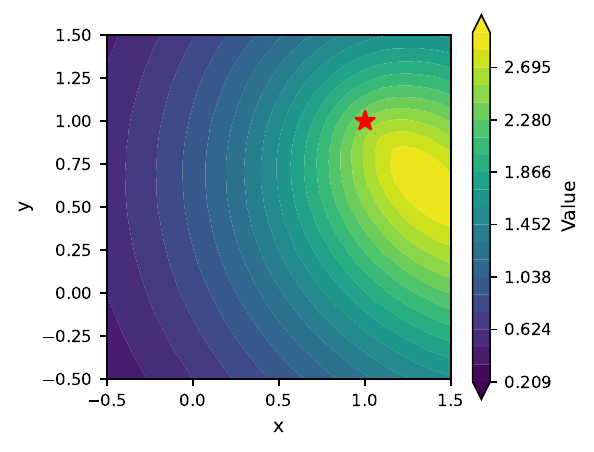}
        \caption{Reference, $\theta=3\pi/4$.}
        \label{fig:robot_pe_kd_ref_theta3pi_4}
    \end{subfigure}
    \caption{Fixed-policy evaluation with known dynamics: fitted critic and independent Monte Carlo reference for different headings.}
    \label{fig:PE_known_dynamics}
\end{figure}

For fixed-policy evaluation, Figure~\ref{fig:PE_known_dynamics} compares the fitted critic with an independent Monte Carlo estimate of the same entropy-regularized discounted return. A common color scale is used within every fitted--reference pair. Across the four tested headings, the fitted critic closely reproduces the location of the high-value region and the orientation of the reference contours. The agreement is therefore not restricted to a single heading and confirms that the neural critic captures the dependence of the value function on both position and orientation.

\begin{figure}
    \centering
    \begin{subfigure}[t]{0.35\textwidth}
        \centering
        \includegraphics[width=\textwidth]{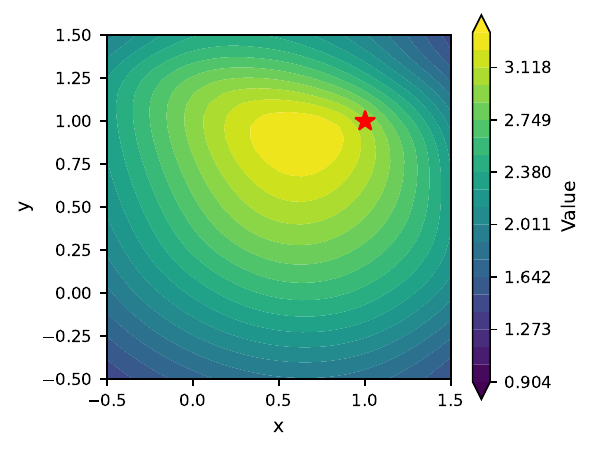}
        \caption{Fitted value, epoch $60$.}
        \label{fig:robot_kd_pred_v}
    \end{subfigure}
    \begin{subfigure}[t]{0.35\textwidth}
        \centering
        \includegraphics[width=\textwidth]{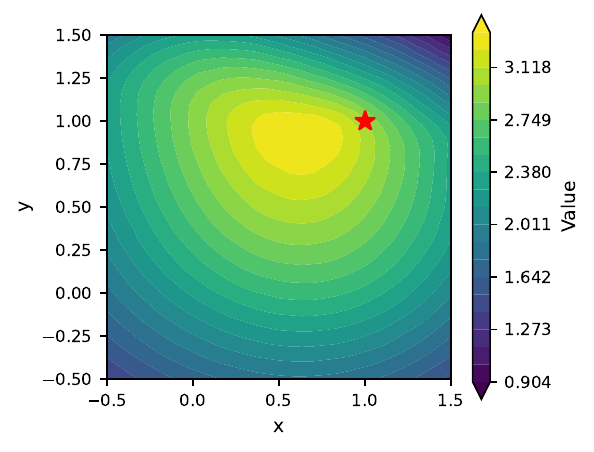}
        \caption{Reference value, epoch $60$.}
        \label{fig:robot_kd_ref_v}
    \end{subfigure}
    \caption{Known-dynamics policy optimization: final value-function comparison at $\theta=0$.}
    \label{fig:PG_known_dynamics_value}
\end{figure}

For the final epoch-$60$ policy, Figure~\ref{fig:PG_known_dynamics_value} compares the fitted value with the corresponding Monte Carlo reference at $\theta=0$. The two surfaces have nearly the same high-value region and contour geometry, showing that the critic remains accurate after the policy has been updated repeatedly.

\begin{figure}
    \centering
    \begin{subfigure}[t]{0.25\textwidth}
        \centering
        \includegraphics[width=\textwidth]{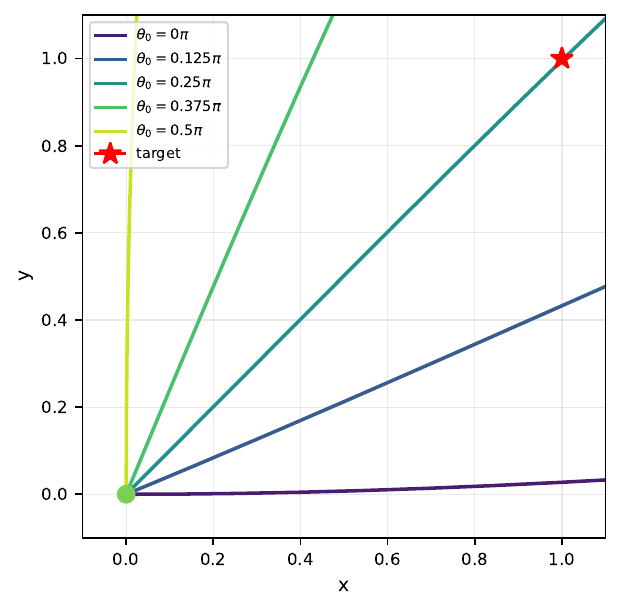}
        \caption{Epoch $0$.}
        \label{fig:robot_kd_traj_epoch0}
    \end{subfigure}
    \begin{subfigure}[t]{0.25\textwidth}
        \centering
        \includegraphics[width=\textwidth]{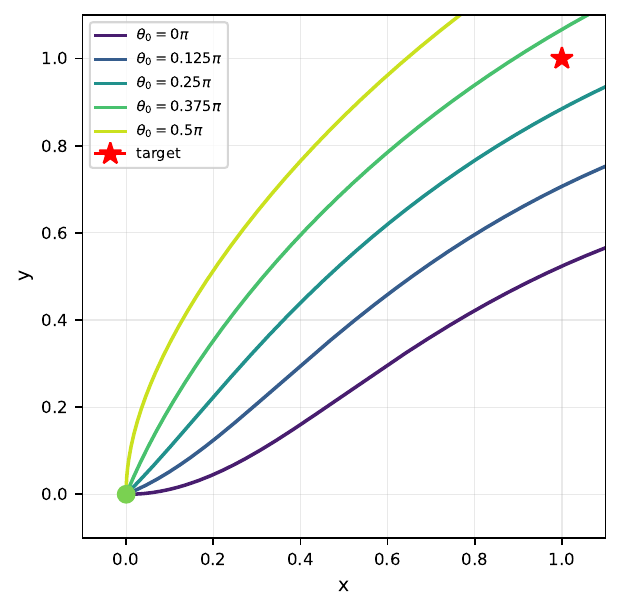}
        \caption{Epoch $10$.}
        \label{fig:robot_kd_traj_epoch10}
    \end{subfigure}
    \begin{subfigure}[t]{0.25\textwidth}
        \centering
        \includegraphics[width=\textwidth]{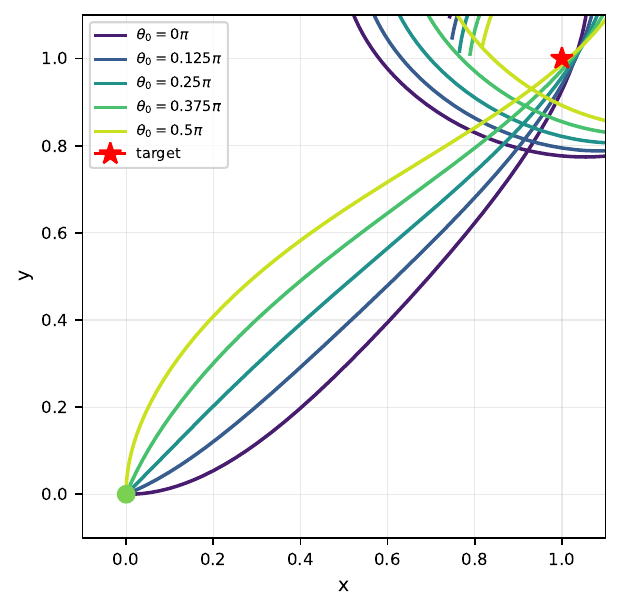}
        \caption{Epoch $60$.}
        \label{fig:robot_kd_traj_epoch60}
    \end{subfigure}
    \caption{Known-dynamics policy evolution. All panels display the common window $[-0.1,1.1]^2$.}
    \label{fig:PG_known_dynamics_traj}
\end{figure}

Figure~\ref{fig:PG_known_dynamics_traj} illustrates the evolution of the deterministic mean policy. At epoch $0$, the trajectories are largely determined by their initial headings and most do not approach the target. By epoch $10$, the learned policy already produces visible heading-dependent turns toward the target region. At epoch $60$, the trajectories corresponding to the different initial headings are redirected toward the neighborhood of $(1,1)$. After reaching this region, some trajectories continue to curve because the robot cannot stop, the target is not absorbing, and no terminal condition is imposed.

\noindent\textbf{Data-driven results.}
We provide here the value-function and trajectory comparisons corresponding to the data-driven experiments of \Cref{subsec:example_differential_wheeled}.

\begin{figure}
    \centering
    \begin{subfigure}[t]{0.22\textwidth}
        \centering
        \includegraphics[width=\textwidth]{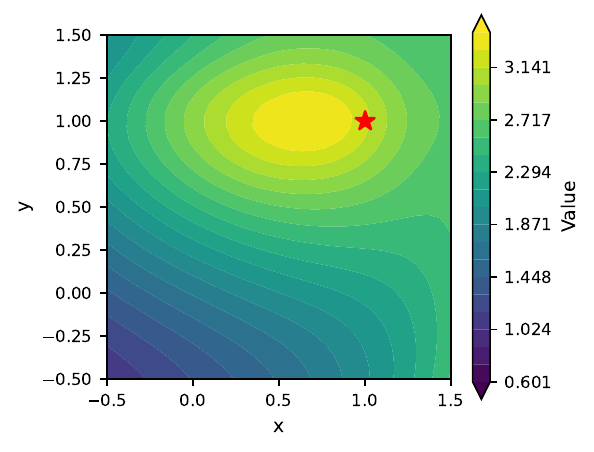}
        \caption{Fitted, $\theta=0$.}
        \label{fig:robot_pe_dd_pred_theta0}
    \end{subfigure}
    \begin{subfigure}[t]{0.22\textwidth}
        \centering
        \includegraphics[width=\textwidth]{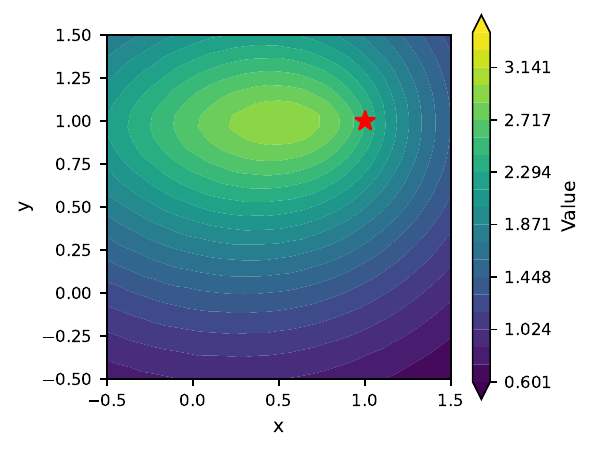}
        \caption{Reference, $\theta=0$.}
        \label{fig:robot_pe_dd_ref_theta0}
    \end{subfigure}
    \begin{subfigure}[t]{0.22\textwidth}
        \centering
        \includegraphics[width=\textwidth]{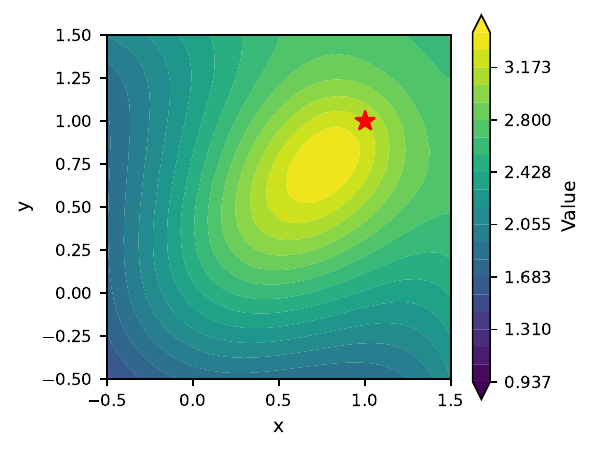}
        \caption{Fitted, $\theta=\pi/4$.}
        \label{fig:robot_pe_dd_pred_thetapi_4}
    \end{subfigure}
    \begin{subfigure}[t]{0.22\textwidth}
        \centering
        \includegraphics[width=\textwidth]{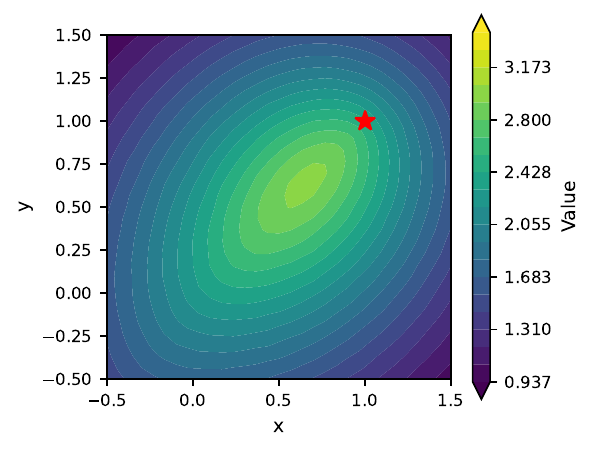}
        \caption{Reference, $\theta=\pi/4$.}
        \label{fig:robot_pe_dd_ref_thetapi_4}
    \end{subfigure}

    \medskip
    \begin{subfigure}[t]{0.22\textwidth}
        \centering
        \includegraphics[width=\textwidth]{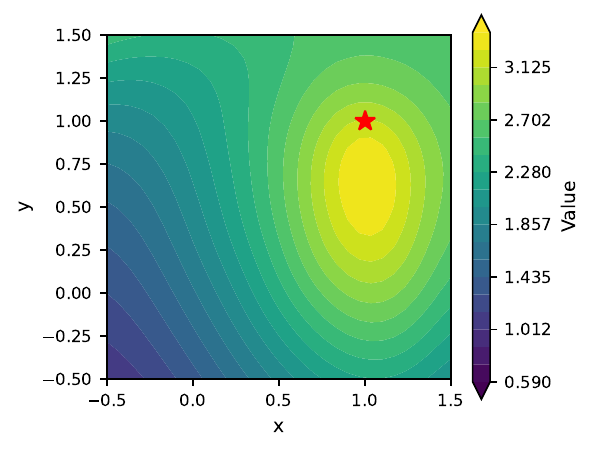}
        \caption{Fitted, $\theta=\pi/2$.}
        \label{fig:robot_pe_dd_pred_thetapi_2}
    \end{subfigure}
    \begin{subfigure}[t]{0.22\textwidth}
        \centering
        \includegraphics[width=\textwidth]{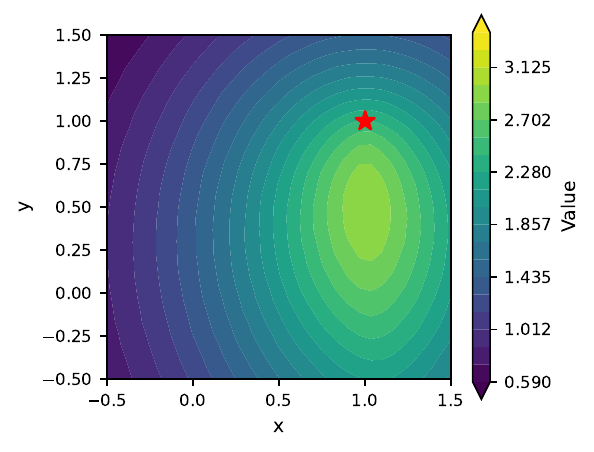}
        \caption{Reference, $\theta=\pi/2$.}
        \label{fig:robot_pe_dd_ref_thetapi_2}
    \end{subfigure}
    \begin{subfigure}[t]{0.22\textwidth}
        \centering
        \includegraphics[width=\textwidth]{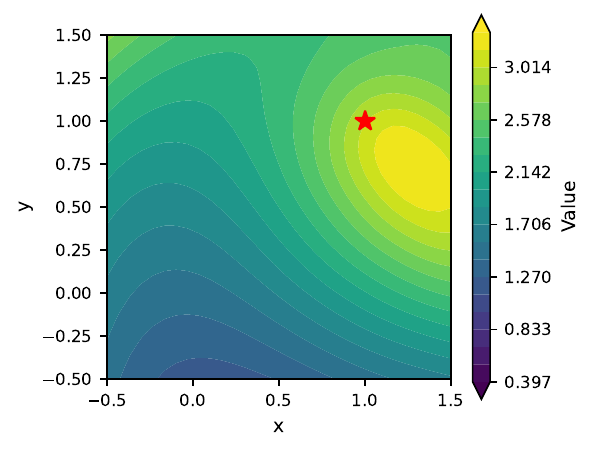}
        \caption{Fitted, $\theta=3\pi/4$.}
        \label{fig:robot_pe_dd_pred_theta3pi_4}
    \end{subfigure}
    \begin{subfigure}[t]{0.22\textwidth}
        \centering
        \includegraphics[width=\textwidth]{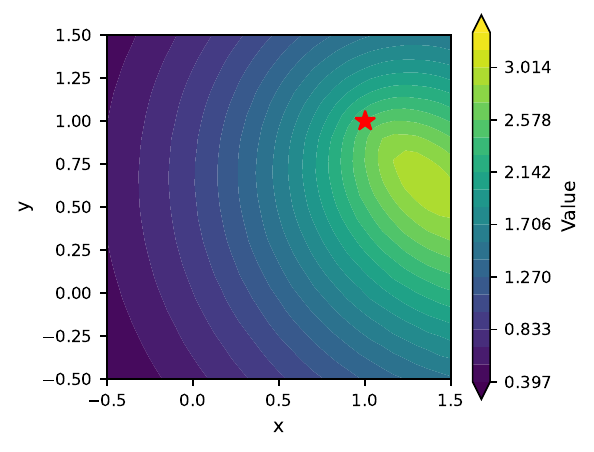}
        \caption{Reference, $\theta=3\pi/4$.}
        \label{fig:robot_pe_dd_ref_theta3pi_4}
    \end{subfigure}
    \caption{Fixed-policy evaluation with data-driven PhiBE: fitted critic and independent Monte Carlo reference for different headings.}
    \label{fig:PE_data_driven}
\end{figure}

For fixed-policy evaluation, Figure~\ref{fig:PE_data_driven} compares the fitted PhiBE critic with independent Monte Carlo estimates of the same entropy-regularized discounted return. A common color scale is used within every fitted--reference pair. Across the tested headings, the fitted critic reproduces the location of the high-value region and its systematic deformation with the robot orientation. Compared with the known-dynamics critic, visible differences remain in some contour shapes and value levels, as expected when the reward and local PhiBE quantities are estimated from finitely many transitions. Nevertheless, the agreement in the global value structure shows that the data-driven critic recovers the main position- and heading-dependent features required for policy improvement.

\begin{figure}
    \centering
    \begin{subfigure}[t]{0.35\textwidth}
        \centering
        \includegraphics[width=\textwidth]{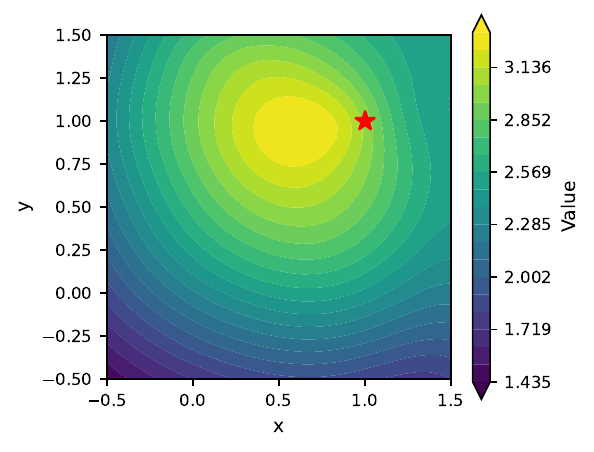}
        \caption{Fitted value, epoch $360$.}
        \label{fig:robot_dd_pred_v}
    \end{subfigure}
    \begin{subfigure}[t]{0.35\textwidth}
        \centering
        \includegraphics[width=\textwidth]{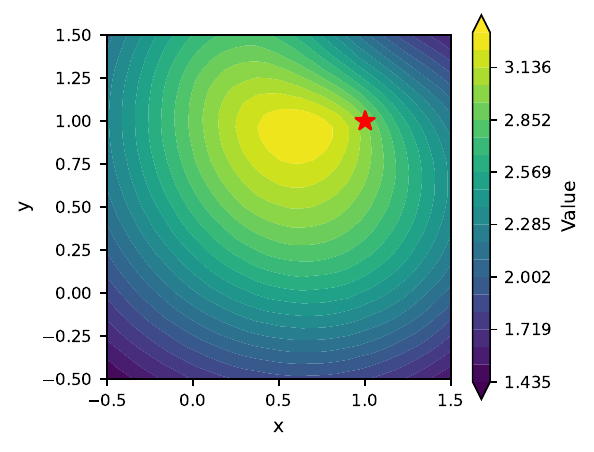}
        \caption{Reference value, epoch $360$.}
        \label{fig:robot_dd_ref_v}
    \end{subfigure}
    \caption{Data-driven policy optimization: final value-function comparison at $\theta=0$.}
    \label{fig:PG_data_driven_value}
\end{figure}

For the final epoch-$360$ policy, Figure~\ref{fig:PG_data_driven_value} compares the fitted value with the independent Monte Carlo reference at $\theta=0$. The two functions recover the same main high-value region and global spatial structure, while some differences remain in the detailed contour geometry. This is consistent with the finite-sample approximation used by the data-driven critic.

\begin{figure}
    \centering
    \begin{subfigure}[t]{0.25\textwidth}
        \centering
        \includegraphics[width=\textwidth]{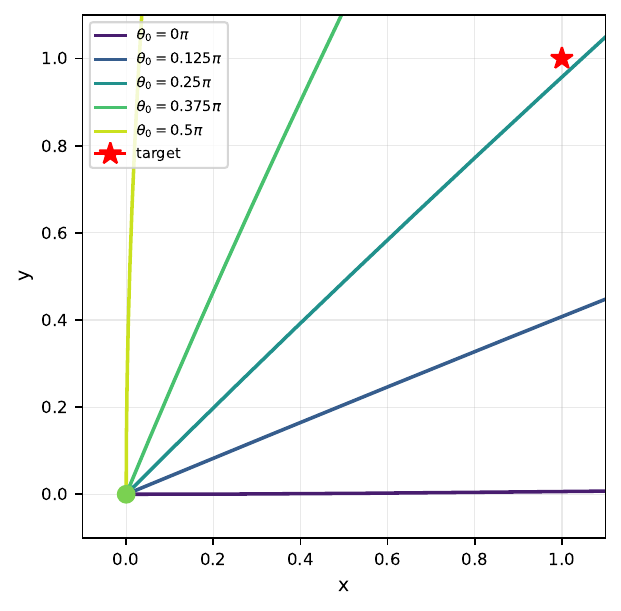}
        \caption{Epoch $0$.}
        \label{fig:robot_dd_traj_epoch0}
    \end{subfigure}
    \begin{subfigure}[t]{0.25\textwidth}
        \centering
        \includegraphics[width=\textwidth]{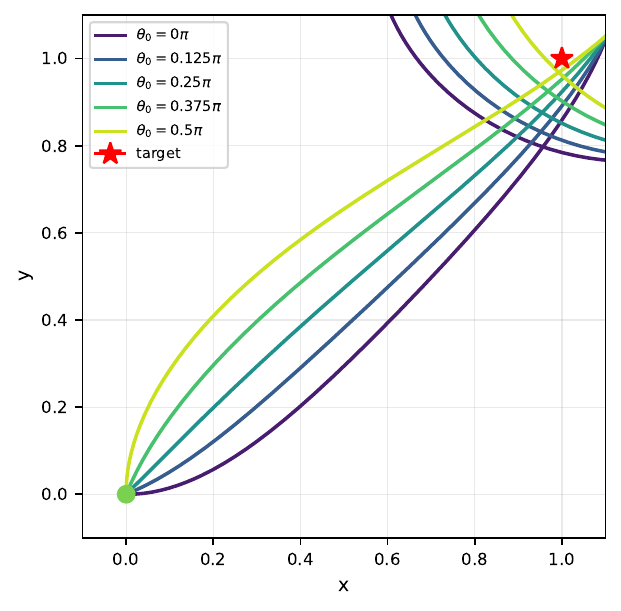}
        \caption{Epoch $40$.}
        \label{fig:robot_dd_traj_epoch40}
    \end{subfigure}
    \begin{subfigure}[t]{0.25\textwidth}
        \centering
        \includegraphics[width=\textwidth]{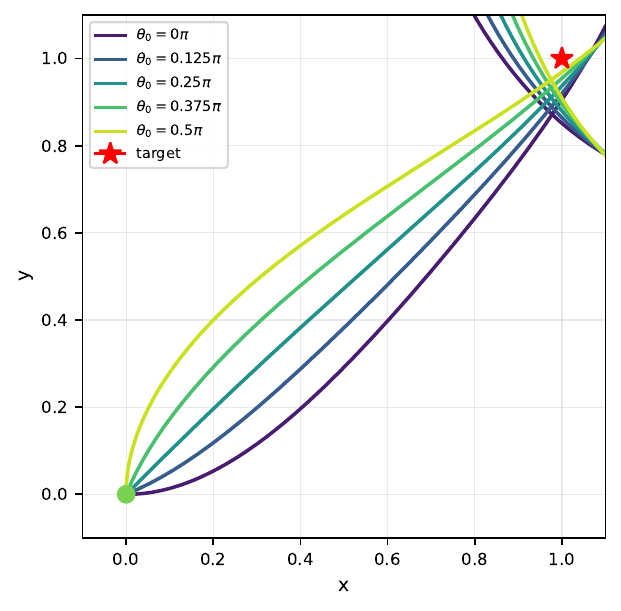}
        \caption{Epoch $360$.}
        \label{fig:robot_dd_traj_epoch360}
    \end{subfigure}
    \caption{Data-driven policy evolution. All panels display the common window $[-0.1,1.1]^2$.}
    \label{fig:PG_data_driven_traj}
\end{figure}

Figure~\ref{fig:PG_data_driven_traj} shows the evolution of the deterministic mean policy. At epoch $0$, the trajectories are largely determined by their initial headings and most do not approach the target. By epoch $40$, all five tested headings are already redirected toward the target neighborhood. The epoch-$360$ policy preserves this behavior and produces a consistent approach to the high-reward region. As in the known-dynamics experiment, trajectories may continue to curve after reaching this neighborhood because the robot has fixed forward speed, the target is not absorbing, and no terminal condition is imposed.

\end{document}